\documentclass[lettersize,journal]{IEEEtran}
\usepackage{amsmath,amsfonts}
\usepackage{algorithmic}
\usepackage{algorithm}
\usepackage{array}
\usepackage[caption=false,font=normalsize,labelfont=sf,textfont=sf]{subfig}
\usepackage{caption}
\usepackage{textcomp}
\usepackage{stfloats}
\usepackage{url}
\usepackage{verbatim}
\usepackage{graphicx}
\usepackage{color}
\usepackage{multicol}
\usepackage{multirow}
\usepackage{pifont}
\usepackage{mathtools}
\usepackage[textsize=tiny]{todonotes}
\usepackage{booktabs}
\usepackage{cite}
\usepackage{cuted}
\usepackage{dsfont}
\usepackage[table]{xcolor}
\DeclareMathOperator*{\argmax}{arg\,max}

\usepackage[numbers,sort&compress]{natbib}
\begin{document}

\title{OODBench: Out-of-Distribution Benchmark for Large Vision-Language Models}

\author{Ling~Lin, Yang Bai, Heng Su, Yang Long,~\IEEEmembership{Senior Member~IEEE}, Congcong Zhu, ~\IEEEmembership{Member,~IEEE}, Yaoxing Wang, Yang Zhou, Ling Shao,~\IEEEmembership{Fellow~IEEE}, Huazhu Fu, ~\IEEEmembership{Senior Member,~IEEE}, Jingrun Chen
        
\IEEEcompsocitemizethanks{
		
\IEEEcompsocthanksitem

        Ling Lin, Heng Su, Congcong Zhu, and Jingrun Chen are with the School of Artificial Intelligence and Data Science, University of Science and Technology of China, Hefei 230000, China, and Suzhou Institute for Advanced Research, USTC, Suzhou 215123, China. E-mail: linglin00d@gmail.com; hengsu@mail.ustc.edu.cn; cczly@ustc.edu.cn; jingrunchen@ustc.edu.cn.

        Yang Bai, Yang Zhou and Huazhu Fu are with the Institute of High Performance Computing (IHPC), A*STAR, Singapore (E-mail: bai\_yang@a-star.edu.sg; yang\_zhou@a-star.edu.sg; hzfu@ieee.org).

        Yang Long is with the Department of Computer Science, Durham University, United Kingdom (E-mail: yang.long@durham.ac.uk).

        Ling Shao is with the UCAS-Terminus AI Lab, University of Chinese Academy of Sciences, Beijing, China  (E-mail:ling.shao@ieee.org).
        
        Yaoxing Wang is with the Unmanned System Research Institute, Northwestern Polytechnical University, Xi’an 71002, China. E-mail: wangyx24@mail.nwpu.edu.cn;

Ling Lin and Yang Bai contributed equally to this work. 
		
		(Corresponding author: Congcong Zhu, Jingrun Chen)
		
}}



\maketitle

\begin{abstract}
Existing Visual-Language Models (VLMs) have achieved significant progress by being trained on large-scale datasets, and are typically evaluated under IID-style settings, where training and test data are assumed to follow similar distributions. However, in real-world scenarios, it is often impractical to expect that all data processed by an AI system satisfy this assumption. Furthermore, failure to appropriately handle out-of-distribution (OOD) objects may introduce safety risks in real-world applications (\textit{e.g.}, autonomous driving or medical assistance). Unfortunately, existing benchmarks remain limited in systematically assessing the performance of VLMs on OOD data. Therefore, we propose \textbf{OODBench}, a predominantly automated method with minimal human verification, for constructing new benchmarks and evaluating the ability of VLMs to process OOD data. OODBench contains 40K instance-level OOD instance–category pairs, and we show that current VLMs still exhibit notable performance degradation on OODBench, even when the underlying image categories are common. In addition, we propose a reliable automated evaluation protocol that employs a Basic-to-Advanced Progression of prompted questions to assess the impact of OOD data on questions of varying difficulty more fully. Lastly, we summarize substantial findings and insights to facilitate future research in the acquisition and evaluation of OOD data \footnote{The dataset is open to the public for research. \url{https://huggingface.co/datasets/LingLin00/OODBench}}.
\end{abstract}

\begin{IEEEkeywords}
Out-of-Distribution, MultiModal Large Language Model, Benchmark, Dataset.
\end{IEEEkeywords}


\section{Introduction}
\IEEEPARstart{R}{ecently}, Visual-Language Models (VLMs) have demonstrated strong performance across multiple domains. Trained on large-scale multimodal data, these models show strong potential across a range of application scenarios.
These results have led to a common perception that large-scale VLMs exhibit strong generalization capabilities. However, comparatively little attention has been paid to the performance of these VLMs under OOD conditions. This limitation cannot be solely attributed to the absence of evaluation benchmarks, but rather reflects the fundamental difficulty of formulating OOD in the context of VLMs. In contrast to conventional recognition tasks defined over a closed and discrete label space, VLMs operate in an open-vocabulary and open-world regime, where semantic concepts are continuous, compositional, and weakly structured. Under such conditions, the boundary between in-distribution and out-of-distribution data becomes ill-defined, rendering category-based definitions of OOD inadequate.
Therefore, we construct a benchmark to evaluate the performance of VLMs on OOD data, thereby advancing the progress of VLMs in OOD research and taking a critical step towards realizing intelligent systems that can operate safely in the real world.


In terms of classification, the OOD data can be classified into two categories according to the different forms of distributional changes: a data distribution shift where labels are altered, known as \textit{semantic shift}, and a shift where labels remain the same but the data distribution changes, also known as \textit{covariate shift}. The OOD data are essentially associated with the model, as it relies on the models' assumptions and learned distributions made during the training process. However, most of the mainstream VLMs are closed-source or only open to model parameters, which prevents us from directly accessing the training distributions of these models. Therefore, how to collect OOD data outside of these model training data distributions presents a fundamental challenge.


In the traditional out-of-distribution literature~\cite{yang2024generalized, hendrycks2019scaling, wang2022vim}, a common paradigm assumes that models are trained on a closed and predefined category space, and OOD benchmarks are constructed by introducing samples from novel categories that are absent from the training set. Such approaches fundamentally rely on the disjointness of label supports, \textit{i.e.}, semantic shift induced by unseen categories. However, this formulation implicitly assumes that the category space is enumerable and well-defined. For modern vision–language models, this assumption no longer holds. Due to large-scale weakly supervised training on open-world data, semantic categories are inherently open-ended, continuous, and loosely structured, rendering category-based OOD construction both theoretically ill-posed and empirically unverifiable. Moreover, the practice of collecting extremely rare or unseen categories to approximate OOD does not guarantee a rigorous semantic shift, and often lacks real-world relevance. Such data collection strategies may introduce uncontrolled biases, thereby undermining the statistical validity of the resulting benchmarks.

To address these limitations, we depart from category-based formulations and instead revisit OOD from the perspective of semantic learning mechanisms in VLMs. Specifically, rather than focusing on whether a category has been observed, we characterize OOD in terms of whether its semantics are sufficiently modeled during training. Based on this principle, we propose a perception-aligned definition of semantic OOD for VLMs, categorizing OOD instances into two types:\newline
(1) Contextually non-salient or weakly aligned objects: objects that are present in the image but exhibit weak semantic coupling with the primary scene semantics. These objects receive limited or ambiguous visual evidence and are rarely emphasized during training, resulting in insufficient semantic alignment with their textual descriptions.\newline
(2) Morphologically shifted variants of salient objects: instances of semantically relevant objects whose visual realizations deviate from their canonical forms, leading to degradation, distortion, or absence of key discriminative evidence required for reliable semantic grounding.

The core of this definition lies in the fact that VLM training predominantly relies on primary semantic objects to establish semantic alignment, while non-primary objects and their morphological variations receive significantly less modeling emphasis. As a result, these instances may be weakly grounded in the learned vision-language semantic space, which naturally correspond to OOD.

We further provide rigorous statistical validation in Section~\ref{Statistic_evidence_ood} to demonstrate that the proposed construction indeed constitutes OOD for existing VLMs. From the perspective of training distribution, we show that the semantic overlap between the constructed data and the training corpora of open-source VLMs is negligible at the population level. From the distributional perspective, we establish statistical equivalence between training-side and benchmark-side OOD subsets in the joint image–category space. From the behavioral perspective, we observe consistent performance degradation patterns from ID to OOD across both open-source and closed-source models. Together, these complementary analyses form a coherent chain of evidence supporting the validity of our OOD formulation.

This definition further requires an operational criterion for identifying such weakly modeled semantics at scale. Building upon this definition, we further characterize the distinction between ID and OOD samples from an information-theoretic perspective. We show that, under the optimal InfoNCE objective, the contrastive scoring function is monotonically related to pointwise mutual information (PMI), and can thus serve as an observable proxy for semantic information strength. In practice, we employ parameterized contrastive models to partition the data and introduce a multi-model cross-validation mechanism to mitigate estimation errors arising from individual models. Specifically, we define the intersection of OOD predictions across multiple models as candidate hard OOD samples (OOD-H), and the symmetric difference as candidate simple OOD samples (OOD-S). A subsequent human verification step is incorporated to ensure consistency with the proposed semantic definition. 

\begin{figure}
    \centering
  \includegraphics[width=0.3\textwidth]{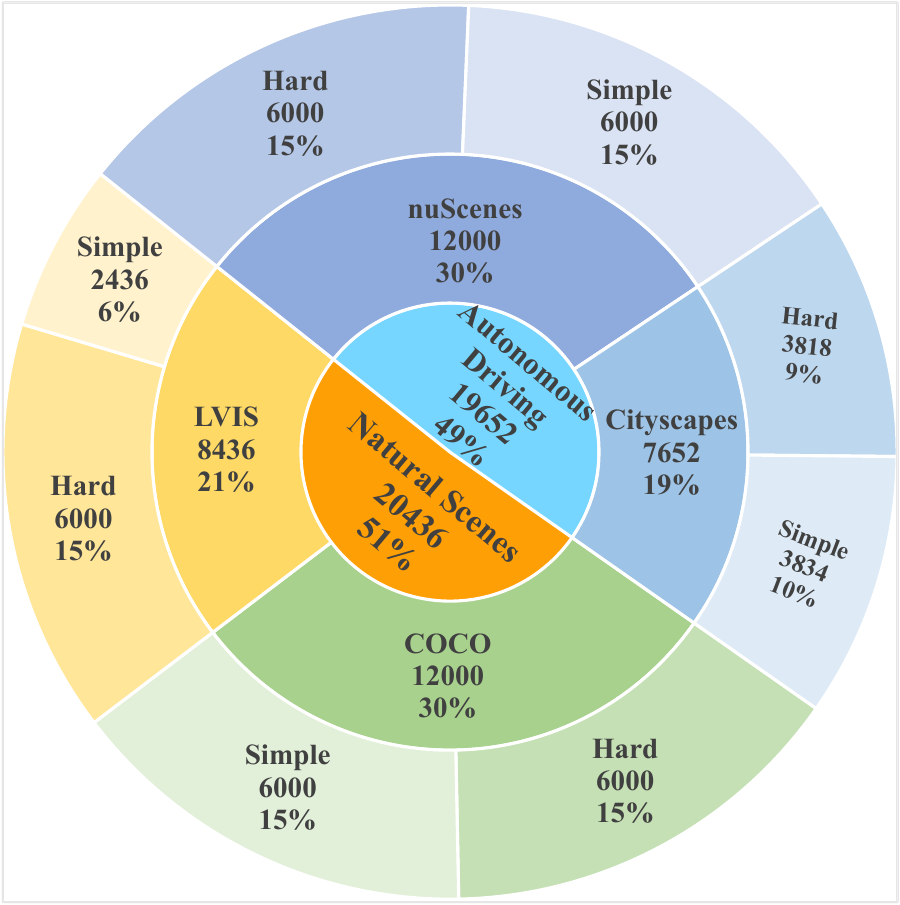}
   \caption{Distribution of categories and fields in \textbf{OODBench}.}
  \label{fig:DatasetDistributionPieChart}
\end{figure}

Based on this pipeline, we construct an initial dataset from a publicly available source\footnote{Details of data distribution and model performance on the initial dataset are provided in Appendix~\ref{appendix: collection details}.}, comprising approximately 77k binary (yes-or-no) samples. To balance evaluation efficiency and statistical reliability, we further perform random downsampling to obtain the final benchmark, termed \textbf{OODBench}. Fig.~\ref{fig:DatasetDistributionPieChart} illustrates the sample distribution of OODBench, which contains about 40k yes-or-no samples, where OOD-S is 22k and OOD-H is 18k. To evaluate VLM performance on OOD data from multiple dimensions, we propose the 'Basic-to-Advanced Procession Metric', which evaluates their ability in image understanding, quantity perception, and logical reasoning, respectively.

Our main contributions can be summarized as the following three: 
\begin{itemize}
    \item We propose a principled OOD data division framework from an information-theoretic perspective, where contrastive scores are interpreted as proxies for semantic mutual information. This leads to a semantic coupling criterion for identifying OOD samples, and enables an efficient, largely automated pipeline with lightweight human validation.
    \item We propose OODBench, an OOD benchmark for VLMs grounded in semantic OOD definitions. It includes a systematically constructed OOD dataset, an OOD data division pipeline, a Basic-to-Advanced Procession Metric, and a comprehensive evaluation protocol.
    \item We evaluate 10 SOTA VLMs on OODBench and demonstrate through extensive experiments that current VLMs still perform poorly under semantic OOD settings.
\end{itemize}

\section{Related Works}
\label{related_work}

\textbf{Out-of-Distribution.}
Out-of-distribution (OOD) detection aims to recognize inputs sampled from a distribution different from the model’s training data~\cite{averly2023unified, yang2023imagenet,ye2025oodbench+,wu2026revisiting}. Traditional OOD research mainly focuses on semantic shift (S-OOD), where novel classes arise, yet covariate shift (C-OOD)—where labels remain the same but inputs deviate—can be equally disruptive~\cite{yang2024generalized, salehi2021unified}. In safety-critical fields like autonomous driving, even minor shifts may cause severe misclassifications and catastrophic results. Recent OOD research~\cite{averly2023unified, yang2023imagenet, jaeger2022call, guerin2023out, zhu2024rethinking} has shifted the focus from new category detection to failure detection scenarios, in which the OOD detector is expected to recognize misclassified samples. In this approach, candidate samples that fail to be classified are labeled as OOD data, and samples that are correctly classified are labeled as ID data to enhance the safety and reliability of deep learning models in real-world applications.

While robust models may correctly handle such shifts, weaker models often benefit from rejecting uncertain samples to mitigate high-risk errors~\cite{liu2021towards}. Reflecting this, OOD research has gradually shifted from purely detecting unseen classes to failure detection, encompassing both semantic and covariate shifts that can cause misclassification~\cite{wiles2021fine, liu2021towards}. Nevertheless, there remains limited exploration of how modern Vision-Language Models (VLMs) contend with these conditions. Our work addresses this gap by constructing an OOD benchmark for VLMs, emphasizing common categories that remain challenging under real-world variation.

\textbf{Multimodal Large Language Models \& Benchmark.} 
Recent Vision-Language Models (VLMs) trained on massive image-text corpora have achieved impressive gains in tasks like visual reasoning, captioning, and Q\&A~\cite{li2023blip,team2024gemini,achiam2023gpt,zou2025uncertainty,miao2025laser,wang2024say}. Notable examples include GPT-4~\cite{achiam2023gpt, hurst2024gpt}, Gemini~\cite{team2024gemini}, LLaVA~\cite{li2024llava}, and InternVL~\cite{chen2024expanding,chen2024far}, often leveraging large-scale datasets such as LAION~\cite{schuhmann2022laion} and MS COCO~\cite{lin2014microsoft}.

Although benchmarks like MMBench~\cite{liu2025mmbench}, SeedBench~\cite{li2023seed}, and GQA~\cite{hudson2019gqa} assess general vision-language performance, they mainly cover in-distribution cases and rarely address out-of-distribution (OOD) data. For example, SeedBench offers 19K multiple-choice questions across images and videos, highlighting model limits but rarely exploring out-of-distribution scenarios. To fill the gap, we introduce OODBench, testing VLM robustness under covariate shift. By incorporating simple and hard OOD categories and employing a “Basic-to-Advanced” protocol, OODBench reveals key weaknesses in current models and encourages future work on safer, more reliable multimodal systems.

While OOD data can trigger hallucinations, not all hallucinations arise from distribution shifts. Benchmarks like HallusionBench~\cite{guan2024hallusionbench} study language and visual hallucinations, whereas OODBench systematically applies covariate shifts to measure genuine OOD resilience.

\section{The OODBench}
\label{oodbench}

\subsection{Existing Definitions and Formulations}
\label{preliminary}
\textbf{Distributional Characterization. }Let $\mathcal{Z} = \mathcal{X} \times \mathcal{Y}$ be the joint space of images and their corresponding labels, where $\mathcal{X}$ denotes the input image manifold and $\mathcal{Y}$ represents the categorical label space. A Vision-Language Model (VLM) $f: \mathcal{X} \rightarrow \mathcal{Y}$ is typically optimized on an empirical distribution $P_{\text{tr}}(x, y)$ where $(x, y) \in \mathcal{Z}$. In the context of out-of-distribution (OOD) generalization, we characterize the discrepancy between the training (ID) distribution $P_{\text{tr}}$ and the test distribution $P_{\text{te}}$ through two fundamental shifts:
\begin{itemize}
    \item \textit{Semantic Shift}~\citep{lipton2018detecting}: Formally defined as the emergence of disjoint label supports. Let $\mathcal{Y}_{\text{tr}}$ and $\mathcal{Y}_{\text{te}}$ be the label sets for training and testing, respectively. A semantic shift occurs when $\mathcal{Y}_{\text{tr}} \cap \mathcal{Y}_{\text{te}} = \emptyset$, such that for any $y \in \mathcal{Y}_{\text{te}}$, the marginal probability in the training set satisfies $P_{\text{tr}}(y) = 0$.
    \item \textit{Covariate Shift}~\citep{shimodaira2000improving, lipton2018detecting}: This involves a transformation in the marginal image distribution $P(x)$ while the conditional label distribution remains invariant. Formally, $P_{\text{tr}}(x) \neq P_{\text{te}}(x)$ but $P_{\text{tr}}(y|x) = P_{\text{te}}(y|x)$. For VLMs, this manifests as a distribution shift where $y \in \mathcal{Y}_{\text{tr}}$, yet the visual realization of $x$ deviates significantly from the training manifold.
\end{itemize}

\begin{figure*}[t]
    \centering
    \includegraphics[width=\linewidth]{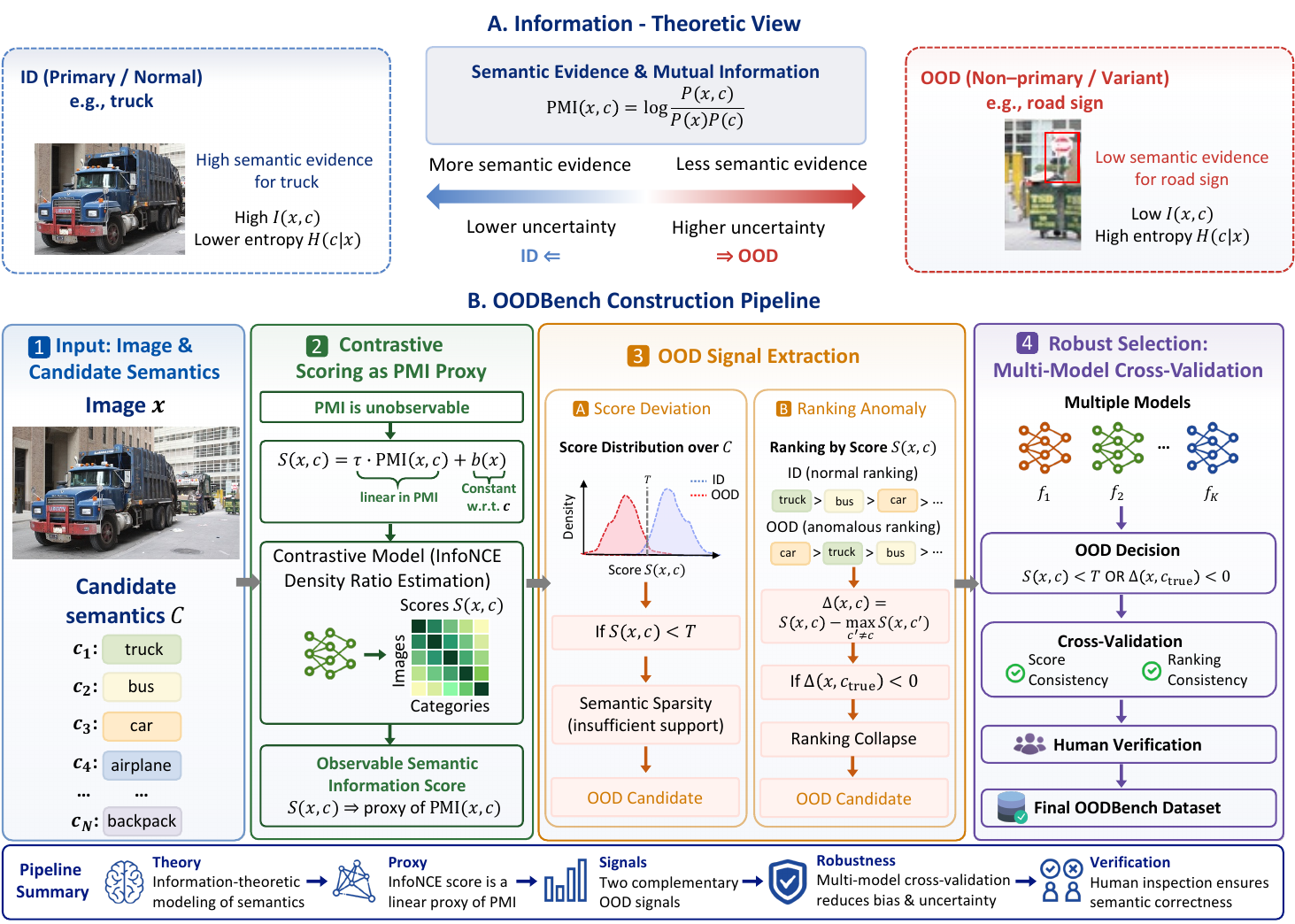}
    \caption{Overview of OODBench. Given an image and a set of candidate semantics, we approximate the unobservable semantic information (PMI) with contrastive scores via InfoNCE density ratio estimation. The resulting score is a linear transformation of PMI (up to an image-dependent bias), providing a valid, monotonic proxy. Based on this proxy, we extract two complementary OOD signal -- score deviation and ranking anomaly, and apply multi-model cross-validation and human verification to construct a high-quality semantic OOD dataset.}
    \label{fig: OOD_Split_Pipeline}
\end{figure*}

\subsection{Problem Setting and Theoretical Preliminaries}
\label{data_collect}
While vision–language models have demonstrated strong zero-shot capabilities, their behavior under out-of-distribution conditions remains insufficiently characterized. We formalize OOD instances in OODBench from a semantic learning perspective, defining them as samples that violate the semantic regularities captured during training. The overall framework is illustrated in Fig.~\ref{fig: OOD_Split_Pipeline}.\\

\noindent \textbf{Definition 1 (Semantic OOD Instance).}\newline
An instance $(x, y)$ is considered an OOD sample if its visual content violates the learned semantic regularities. Specifically, an instance is categorized as OOD if it satisfies at least one of the following conditions:
\begin{itemize}
    \item \textit{Contextual Irrelevance:} The object $y$ is spatially present but is neither a primary semantic entity nor semantically related to them;
    \item \textit{Morphological Variance:} The instance $y$ appears in a variant or anomalous form that deviates from its typical semantic representation.
\end{itemize}

\subsubsection{Information-Theoretic Formulation}


Let the random variable $X$ represent an image, and let $C_t$ represent the semantic description of the object $t$. The Pointwise Mutual Information (PMI) between a specific image instance $x$ and a semantic concept $c$ is defined as the log-ratio of their joint probability to the product of their marginals:

\begin{equation}
\text{PMI}(x,c) = \log \frac{p(x,c)}{p(x)p(c)}.
\end{equation}

The Mutual Information $I(C_t; X)$ can then be expressed as the expected value of the PMI over all possible pairs $(x,c)$, which also quantifies the reduction in uncertainty of $C_t$ given the observation of $X$:

\begin{equation}
\begin{split}
I(C_t; X) &= H(C_t) - H(C_t \mid X) \\
&= \mathbb{E}_{(x,c) \sim p(x,c)} \left[ \log \frac{p(x,c)}{p(x)p(c)} \right] \\
&= \mathbb{E}_{(x,c) \sim p(x,c)} \left[ \text{PMI}(x,c) \right].
\end{split}
\label{eq:mutual_info}
\end{equation}

To characterize the extent to which an image supports a object $t$, we introduce the concept of local evidence. For any semantic object $t$, suppose there exists a set of local evidence in the image $X$ that is relevant to that object:

\begin{equation}
    \begin{split}
        Z_t = \{Z_{t,1}, Z_{t,2},\cdots, Z_{t,n_t}\},
    \end{split}
\end{equation}

\noindent where each $Z_{t,j}$ represents an information source in the image that is relevant to object $t$; \textit{e.g.}, the visible region, shape attributes, texture features, key components, or contextual relationships, \textit{etc}. Under this definition, given a set of evidence $Z_t$, the posterior uncertainty of the semantic description is $H(C_t | Z_{t,1:n_t})$. Thus, the \textit{cumulative evidence strength} of the image for object $t$ can be defined as:

\begin{equation}
    \begin{split}
        \varepsilon_t(X):=I(C_t; Z_{t,1:n_t})=\sum_{j=1}^{n_t}I(C_t;Z_{t,j} \mid Z_{t,1:j-1}),
    \end{split}
\end{equation}

\noindent where $Z_t$ is assumed to capture the sufficient evidence in $X$ relevant to $t$. This definition shows that the cumulative evidence strength equals the sum of the marginal information gains from each piece of local evidence about the object's description.

From the relationship between conditional entropy and conditional mutual information, it follows that:

\begin{equation}
    \begin{split}
        H(C_t \mid Z_{t,1:n+1}) = H(C_t \mid Z_{t,1:n}) & -I(C_t;Z_{t,n+1} \mid Z_{t,1:n}) \\
        & \leq H(C_t \mid Z_{t,1:n}),
    \end{split}
\end{equation}

\noindent where the inequality holds strictly when the new evidence $Z_{t,n+1}$ contains non-zero information. This indicates that as valid evidence accumulates, the uncertainty of the semantic description decreases monotonically. Therefore, the certainty of the caption is determined by the cumulative amount of valid evidence provided by the image regarding that semantic objective.

\subsubsection{Statistical Characterization of Semantic Dominance} Based on the above analysis, the importance of a semantic object can be characterized by its average cumulative evidence strength across the data distribution. Let $t$ denote any candidate semantic object in an image; its cumulative evidence strength is defined as $\varepsilon_t (X):= I(C_t; Z_{t,1:n_t})$. On the data distribution $\mathcal{D}$, the primary semantic object can be distinguished from non-primary semantic objects by their expected evidence strength:

\begin{equation}
    \begin{split}
        \mathbb{E}_{X\sim \mathcal{D}}[\varepsilon_m(X)] > \mathbb{E}_{X\sim \mathcal{D}}[\varepsilon_s(X)],
    \end{split}
\end{equation}

\noindent \textit{i.e.}, the primary semantic objects, are statistically more observable and distinguishable.

Combining the definition of cumulative evidence strength, the above equation is equivalent to:

\begin{equation}
    \begin{split}
        \mathbb{E}_{X\sim \mathcal{D}}[I(C_m;Z_{m,1:n_m})]>\mathbb{E}_{X\sim \mathcal{D}}[I(C_s;Z_{s,1:n_s})],
    \end{split}
    \label{eq:main_second_object_mutual_info_inequ}
\end{equation}

\noindent this inequality indicates a stronger statistical dependence between the primary semantic object and the image evidence, thereby enabling greater information gain from the observations.

In the absence of an observed image $X$, the generation of the caption is determined solely by linguistic prior knowledge and is independent of specific semantic objects. Therefore, for any semantic object $t$, its marginal distribution satisfies:

\begin{equation}
    \begin{split}
        p(C_t)\approx p_{\text{lang}}(C),
    \end{split}
\end{equation}

\noindent therefore, its marginal entropy is approximately constant across different objects, \textit{i.e.},

\begin{equation}
    \begin{split}
        H(C_t)\approx \text{const}.
    \end{split}
\label{eq:marginal_entropy_approx_const}
\end{equation}

Furthermore, by the definition of mutual information in Eq.~\ref{eq:mutual_info}, in conjunction with Eq.~\ref{eq:marginal_entropy_approx_const}, and assuming that the variation in the marginal entropy term $H(C_t)$ is limited, the inequality in Eq.~\ref{eq:main_second_object_mutual_info_inequ} implies that:

\begin{equation}
    \begin{split}
        H(C_s \mid X)>H(C_m \mid X),
    \end{split}
    \label{eq:entrpy_m_s}
\end{equation}

\noindent these results indicate that the difference in mutual information stems primarily from the conditional entropy term; that is, after observing the image $X$, the semantic uncertainty of non-primary semantic objects remains significantly higher than that of primary semantic objects.

\subsubsection{Information Degradation under Morphological Variance} Within the framework of local evidence, consider the differences in evidence structures for the same semantic object $t$ across different morphological states. Let:

\begin{itemize}
    \item $Z_t^\text{norm}$ denotes the set of evidence when the object is in its normal state.
    \item $Z_t^\text{pert}$ denotes the set of evidence when the object undergoes a state change (such as a variant or anomaly).
\end{itemize}

Due to morphological variations, part of the key evidence may be missing, distorted, or exhibit reduced discriminative power. As a result, $Z_t^{\text{pert}}$ exhibits degraded information contribution compared to $Z_t^{\text{norm}}$. According to the definition of cumulative evidence strength, it follows that:

\begin{equation}
    \begin{split}
        \varepsilon_t^{\text{norm}}(X):=I(C_t;Z_t^{\text{norm}}),\varepsilon_t^{\text{pert}}(X):=I(C_t;Z_t^{\text{pert}}).
    \end{split}
\end{equation}

Therefore, under conditions of evidence degradation, we have:

\begin{equation}
    \begin{split}
        \varepsilon_t^{\text{norm}}(X)>\varepsilon_t^{\text{pert}}(X).
    \end{split}
\end{equation}

Combining this with the definition of cumulative evidence strength, the above formula is equivalent to:

\begin{equation}
    \begin{split}
        \mathbb{E}_{X\sim \mathcal{D}}[I(C_t;Z_t^{\text{pert}})]<\mathbb{E}_{X\sim \mathcal{D}}[I(C_t;Z_t^{\text{norm}})].
    \end{split}
    \label{eq:norm_pert_object_mutual_info_inequ}
\end{equation}

Furthermore, from Eqs.~\ref{eq:mutual_info} and~\ref{eq:marginal_entropy_approx_const}, we obtain:

\begin{equation}
    \begin{split}
        H(C_t \mid X)^{\text{pert}}>H(C_t \mid X)^{\text{norm}}.
    \end{split}
    \label{eq:entrpy_n_p}
\end{equation}

These results indicate that, compared to normal morphology, when the object undergoes morphological changes, the statistical dependence between it and the image is weakened due to the degradation of valid evidence, leading to a monotonic increase in the conditional uncertainty of the semantic description. In other words, morphological changes weaken the information coupling between the original semantics and the observation, rendering the semantic state unstable.

\subsection{Connection Between Contrastive Scores and Mutual Information}
\label{Contrastive_Scores_and_Mutual_Information}
Let $x$ denote an image and $c$ a semantic caption. Within the InfoNCE framework, given a set of candidates $\mathcal{C} = \{c_1, \dots, c_N\}$, one positive sample is drawn from the conditional distribution $p(c|x)$, while the remaining negative samples are drawn independently from the marginal distribution $p(c)$. The model employs a scoring function $f(x, c)$ and defines the posterior probability as follows:

\begin{equation}
    \begin{split}
        q_\theta(\mathcal{I}=i \mid x,\mathcal{C})=\frac{\exp (f(x,c_i)/\tau)}{\sum_{k=1}^N \exp(f(x,c_k)/\tau)},
    \end{split}
    \label{eq: model_posterior}
\end{equation}

\noindent where $\tau > 0$ is a temperature parameter, and $\mathcal{I}$ denotes the index of the positive sample in the candidate set.

Under this sampling mechanism, Bayes' theorem implies that the true posterior satisfies:

\begin{equation}
    \begin{split}
        p(\mathcal{I}=i \mid x, \mathcal{C})=\frac{\frac{p(c_i|x)}{p(c_i)}}{\sum_{k=1}^N \frac{p(c_k|x)}{p(c_k)}}.
    \end{split}
    \label{eq: ture_posterior}
\end{equation}

Given that the InfoNCE loss is inherently a cross-entropy loss for an $N$-way classification task, the optimal solution under the assumption of sufficient model capacity satisfies:

\begin{equation}
    \begin{split}
        q_\theta ^* (\mathcal{I}=i \mid x,\mathcal{C})=p(\mathcal{I}=i \mid x,\mathcal{C}).
    \end{split}
\end{equation}

Comparing Eq.~\ref{eq: model_posterior} with Eq.~\ref{eq: ture_posterior}, it follows that the unnormalized term of the optimal scoring function must be proportional to the density ratio $\frac{p(c|x)}{p(c)}$, \textit{i.e.},

\begin{equation}
    \begin{split}
        \exp (f^*(x,c)/\tau)\propto \frac{p(c\mid x)}{p(c)}.
    \end{split}
\end{equation}

Taking the natural logarithm of the above equation yields:

\begin{equation}
    \begin{split}
        f^*(x,c)=\tau \log \frac{p(c \mid x)}{p(c)} + b(x),
    \end{split}
\end{equation}

\noindent where $b(x)$ is an additive term depending solely on $x$. Furthermore, since

\begin{equation}
    \begin{split}
        \log \frac{p(c \mid x)}{p(c)} = \log \frac{p(x,c)}{p(x)p(c)}=\text{PMI}(x, c),
    \end{split}
\end{equation}

\noindent it follows that:

\begin{equation}
    \begin{split}
        f^*(x,c)=\tau \text{PMI}(x,c)+b(x).
    \end{split}
    \label{eq:CS_PMI_relative}
\end{equation}

Consequently, under the optimal InfoNCE objective, the contrastive scoring function recovers a linear transformation of the PMI. Specifically, when comparing different candidate semantics for the same image $x$, $b(x)$ does not affect the ranking; thus, $f(x,c)$ serves as a monotonic proxy for $\text{PMI}(x,c)$. Furthermore, given that:

\begin{equation}
    \begin{split}
        I(X; C)=\mathbb{E}_{(x,c)\sim p(x,c)}[\text{PMI}(x,c)],
    \end{split}
\end{equation}

\noindent it can be inferred that the contrastive scores correspond monotonically to the mutual information $I(X;C)$ in an expectation sense. The complete derivation is provided in the Appendix~\ref{appendix: MI_CS_proof}.

\subsection{Score- and Ranking-based OOD Detection}

In the preceding sections, we have characterized the fundamental distinctions between ID and OOD samples from an information-theoretic perspective. As established in Eq.~\ref{eq:CS_PMI_relative}, when the InfoNCE objective reaches its optimum, the contrastive scoring function can be expressed as a linear transformation of the PMI. Consequently, given a specific image $x$, the contrastive score can be used as a monotonic proxy for pointwise mutual information, and thus reflects the semantic grounding strength between an image and a candidate concept. Leveraging this property, the contrastive score $S(x,c)$ serves as an observable proxy for semantic information intensity, thereby providing an operational criterion for OOD discrimination. Under this theoretical framework, the subsequent sections will propose two concrete methods for partitioning candidate OOD data. Furthermore, in Section~\ref{Statistic_evidence_ood}, we will conduct systematic statistical analyses and empirical studies to validate the effectiveness and robustness of the proposed approaches.

\subsubsection{Score Deviation}
\label{Score_Deviation}
For any arbitrary image-text pair $(x, c)$, let $S(x, c)$ denote its contrastive score. Based on Eq.~\ref{eq:main_second_object_mutual_info_inequ} and Eq.~\ref{eq:norm_pert_object_mutual_info_inequ}, it can be observed that primary semantic objects yield higher mutual information than non-primary ones. Similarly, the normal states of an object exhibit higher mutual information than their variants or anomalous forms. Given the monotonic correspondence between mutual information and contrastive scores, it can be inferred that, compared to ID samples, OOD samples typically possess lower semantic information content, which manifests as lower values in the contrastive scoring space.

In practical scenarios, the approximation of the parameterized model to the ideal optimal solution can be formulated as:

\begin{equation}
S_\theta(x,c) = f^*(x,c) + \epsilon_\theta(x,c), 
\label{eq:PMI_parameterized}
\end{equation} 

\noindent where $f^*(x,c)$ denotes the ideal linear transformation of the PMI, and $\epsilon_\theta(x,c)$ characterizes the estimation bias introduced by factors such as limited model capacity, training strategies, and sampling mechanisms. Notably, in low-score regions, the relative impact of the noise term is significantly amplified due to the inherent weakness of the true semantic signal, leading to high instability in ranking results based on extreme values. Consequently, relying solely on the tail of the ranking as the basis for OOD discrimination tends to introduce substantial noise, thereby undermining the reliability of the partitioning.

Furthermore, we formalize the aforementioned observations from an information-theoretic perspective. In an ideal representation space, the image and text semantics of ID samples are highly aligned, manifested as elevated semantic mutual information $I(x;c)$. Conversely, OOD samples exhibit a significant collapse in mutual information due to semantic drift or structural impairment. To this end, we propose a classification criterion based on a semantic coupling threshold. Assuming the existence of a critical mutual information capacity $T$, the semantic density of the sample space exhibits a bimodal or stratified trend:
\begin{itemize}
    \item Semantic Support Region ($S(x,c) \geq T$): Samples within this region maintain robust semantic coupling, where the information entropy is primarily contributed by core features within the ID manifold, consistent with ID distribution characteristics.
    \item Semantic Sparsity Region ($S(x,c) < T$): When the contrastive score $S(x,c)$ falls below this threshold, it implies that the mutual information between the image-text pair is insufficient to sustain the semantic integrity required for ID samples, causing them to fall into the semantic sparsity zone.
\end{itemize}
The logical foundation of this partitioning method lies in the fact that $T$ constitutes the boundary between "structured semantics" and "unstructured noise or variants." Statistically, as $S(x,c)$ decreases, the statistical support of the sample for the predefined semantics $c$ monotonically diminishes, while its probability of being an OOD candidate increases accordingly. Consequently, score deviations naturally constitute the first type of statistical signal in OOD detection.

\subsubsection{Ranking Anomaly} Under ideal conditions, the contrastive score and mutual information maintain consistency in terms of ranking; thus, the relative strength of semantic information can be characterized through the ranking structure. For any semantic candidate $c$, we define its relative dominance as:

\begin{equation}
    \Delta(x,c) = S(x,c) - \max_{c' \neq c} S(x,c').
\end{equation}

This metric measures the relative informational advantage of semantic $c$ over other candidates. In ID distributions, since the ground-truth semantics correspond to higher mutual information, it typically holds that $\Delta(x, c_{\text{true}}) > 0$, indicating that the true semantics dominate the ranking. However, when semantic perturbations occur in the image, the mutual information of relevant semantics decreases due to the degradation of effective evidence, leading to a reduction in $\Delta(x, c_{\text{true}})$. Furthermore, if stable alternative semantics suppress the true category—resulting in the emergence of an anomalous surrogate class such that $\Delta(x, c_{\text{true}}) < 0$—it signifies that the original semantic ranking structure has been disrupted, and the true semantics no longer occupy the dominant position. Based on this, we define the aforementioned phenomenon as a Ranking Anomaly. This anomaly refers to the loss of relative ranking advantage by a semantic item that should originally possess it. It reflects that the organization of semantic information in the image has deviated from the stable ranking structure observed under ID distributions, serving as a direct manifestation of OOD semantic perturbations at the ranking level. Consequently, ranking anomalies constitute the second category of statistical signals for OOD detection.

\subsubsection{Multi-model cross-validation mechanism} In practical applications, the estimation errors of contrastive scores originate from the synergistic effects of both model-level and data-level factors. These superimposed errors directly compromise the stability and reliability of OOD discrimination.

(1) Model Level: Mutual Information Estimation Bias. Different parameterized models exhibit varying capabilities in approximating the InfoNCE optimal solution. Consequently, their corresponding error terms, $\epsilon_\theta(x,c)$, differ significantly, leading to instability in contrastive score estimation and ranking structures within a single-model setting.

(2) Data Level: High Uncertainty of OOD Samples. As established in Eq.~\ref{eq:entrpy_m_s} and Eq.~\ref{eq:entrpy_n_p}, OOD samples satisfy $H(\mathcal{C} \mid X)_{\text{OOD}} > H(\mathcal{C} \mid X)_{\text{ID}}$, indicating higher conditional entropy and stronger semantic uncertainty. Consequently, OOD samples not only yield lower mutual information but are also more sensitive to model perturbations, making their contrastive scores and ranking structures more prone to fluctuations across different models.

In summary, a single model in OOD scenarios is simultaneously plagued by estimation bias and inherent uncertainty, resulting in a lack of discriminative stability. To address this, we introduce a multi-model cross-validation mechanism. Specifically, given a set of models $\{f_k\}_{k=1}^K$, we compute the contrastive scores and ranking structures for each model and filter candidates based on two consistency criteria: consistency in score deviation and consistency in ranking anomalies. This strategy enhances robustness by averaging out independent model errors and leveraging the stable anomalous patterns exhibited by OOD samples across multiple models. Furthermore, by integrating the two categories of statistical signals—score deviation and ranking anomaly—under multi-model consistency constraints, we identify a sample as an OOD candidate if it satisfies either of the following conditions:

\begin{equation}
    \mathbb{I}_{\text{OOD}}(x,c) =
    \begin{cases}
        1, & \text{if } S(x,c) < T, \\ 
        1, & \text{if } \Delta(x,c) < 0, \\
        0, & \text{otherwise.}
    \end{cases}
\end{equation}

The aforementioned criteria provide a unified characterization of significant semantic information deviation and systematic disruption of ranking structures within an information-theoretic framework. Following the OOD partitioning, we further validate the results through manual sampling inspections to ensure that the identified candidates strictly adhere to our definition of OOD.

\section{Statistical Evidence for the OOD Nature of OODBench in Modern VLMs}
\label{Statistic_evidence_ood}

To formally verify that the constructed benchmark constitutes OOD data for modern MLLMs, we establish a statistical inference framework from both the training-distribution perspective and the behavioral-response perspective.

\paragraph{OOD Evidence from Open-Source Models}
Let $\mathcal{D}_{\text{train-OOD}}$ denote the train set constructed under the same semantic partition rules as OODBench. For each model $\textbf{m}$, we define its semantic overlap rate with respect to $\mathcal{D}_{\text{train-OOD}}$ as
\begin{equation}
r_m = \frac{\mathcal{H}_m}{|C|},
\end{equation}
where $\mathcal{H}_m$ denotes the number of semantically matched image-text pairs observed in its training corpus, and $|C|$ is the total number of pairs in $\mathcal{D}_{\text{train-OOD}}$.

We introduce a pre-registered OOD decision rule based on the upper confidence bound:
\begin{equation}
Z_m = \mathds{1}\big[\mathrm{UCI}_{0.95}(r_m) \leq \varepsilon \big],
\end{equation}
where $\varepsilon$ is a small tolerance. A dataset is considered training-distribution OOD with respect to model $\mathbf{m}$ if its estimated semantic overlap with the model's training corpus is statistically negligible.


Let $Z_m \sim \mathrm{Bernoulli}(\theta)$ denote whether model $\textbf{m}$ regards the dataset as OOD. Given $k$ successes out of $n$ evaluated models, we estimate the population-level OOD proportion $\theta$.

Using exact binomial inference and Bayesian posterior estimation, we derive a $95\%$ lower confidence bound of the form
\begin{equation}
\theta \geq \theta_{\min}(k,n),
\end{equation}
where $\theta_{\min}(k,n)$ is determined by the observed success count and sample size.

In our empirical setting (\textit{e.g.}, $k=49$, $n=50$ under $\varepsilon=5\%$), this yields $\theta_{\min} \approx 0.89$, indicating that a large majority of comparable models are expected to classify the data as OOD.

This result implies that, at the population level, the constructed split is overwhelmingly outside the training semantic distribution of existing open-source MLLMs.

\paragraph{Distributional transfer from train to benchmark split.}
Let $(X,C) \in \mathcal{X} \times \mathcal{C}$ denote the joint image-text variable. We consider two distributions:
\[
P_{\text{train}}(X,C), \quad P_{\text{val}}(X,C),
\]
corresponding to train-OOD and val-OOD constructed under identical rules.

To verify their equivalence, we measure the distributional discrepancy using kernel Maximum Mean Discrepancy (MMD):
\begin{equation}
\mathrm{MMD}^2(P_{\text{train}}, P_{\text{val}}) = \|\mu_{\text{train}} - \mu_{\text{val}}\|_{\mathcal{H}}^2.
\end{equation}

We define a tolerance threshold $\xi$ as the $95$-th percentile of intra-distribution fluctuations:
\begin{equation}
\xi = \mathrm{Quantile}_{0.95}\left(\mathrm{MMD}^2(D^{(1)}, D^{(2)})\right),
\end{equation}
where $D^{(1)}, D^{(2)}$ are random splits from $P_{\text{train}}$.

The equivalence hypothesis is formulated as:
\[
H_0: \mathrm{MMD}^2 \geq \xi, \quad
H_1: \mathrm{MMD}^2 < \xi.
\]

Empirically, the upper confidence bound of $\mathrm{MMD}^2(P_{\text{train}}, P_{\text{val}})$ satisfies
\begin{equation}
\text{UCI}_{0.95} < \xi,
\end{equation}
thus rejecting $H_0$ and establishing distributional equivalence:
\begin{equation}
P_{\text{train}}^{\text{OOD}} \approx P_{\text{val}}^{\text{OOD}}.
\end{equation}

This result enables the transfer of training-level OOD conclusions to the final benchmark split.

\paragraph{Behavioral OOD evidence for closed-source models.}
For models with inaccessible training data, we instead characterize OOD via performance degradation. Let
\begin{equation}
\Delta_m^{(k)} = T_{m,\text{ID}}^{(k)} - T_{m,\text{OOD}}^{(k)},
\end{equation}
where $T^{(k)}$ denotes evaluation metrics (\textit{e.g.}, ACC, F1).

OOD behavior is verified through the hypothesis test:
\[
H_0: \Delta_m^{(k)} = 0, \quad
H_1: \Delta_m^{(k)} > 0.
\]

Across all evaluated closed-source models, we observe statistically significant $\Delta_m^{(k)} > 0$, and their magnitudes fall within the degradation band induced by open-source OOD responses:
\begin{equation}
\Delta_c^{(k)} \in [\Delta_{\min}^{(k)}, \Delta_{\max}^{(k)}].
\end{equation}

This consistency indicates that closed-source models exhibit the same OOD behavioral signature as open-source models.

\paragraph{Summary}
Combining (i) negligible training-semantic overlap, (ii) distributional equivalence between train and benchmark splits, and (iii) consistent ID$\rightarrow$OOD degradation patterns, we conclude that OODBench constitutes statistically verifiable OOD data for modern MLLMs. Detailed derivations and full statistical procedures are provided in Appendix~\ref{appendix: ood nature evidence}.

\section{Experiment}
\label{experiment}
\subsection{Selected Datasets}
\textbf{COCO.~\cite{lin2014microsoft}.} The COCO dataset is a widely used natural image dataset in computer vision. It contains over 330,000 images across 80 categories (\textit{e.g.}, people, animals, vehicles, \textit{etc.}). We chose its 2017 version validation set as the source of OODBench metadata.

\textbf{LVIS~\cite{gupta2019lvis}.} The LVIS dataset is a large-scale natural scene dataset that focuses on instance segmentation. Especially, it focuses on the recognition of fine-grained categories and provides rich instance-level annotations.

\textbf{nuScenes~\cite{caesar2020nuscenes}.} nuScenes is a large-scale autonomous driving dataset that provides 360-degree views of city streets, covering a wide range of urban environments and road conditions. The dataset provides camera, LiDAR, and radar sensor data. To control the benchmark's sample number, we selected 6,019 keyframes from the front camera in the validation set as metadata for nuScenes.

\textbf{Cityscapes~\cite{Cordts2016Cityscapes}.} Cityscapes is an autonomous driving dataset focused on urban street scenes for scene understanding. We chose images from the camera on the left side of the validation set as metadata for Cityscapes. In addition, we adjusted labels to ensure that their expressions conform to natural language conventions and excluded potentially ambiguous labels. For more details, please refer to Appendix~\ref{appendix: collection details}.

\begin{figure*}[t]
  \centering
   \includegraphics[width=0.8\linewidth]{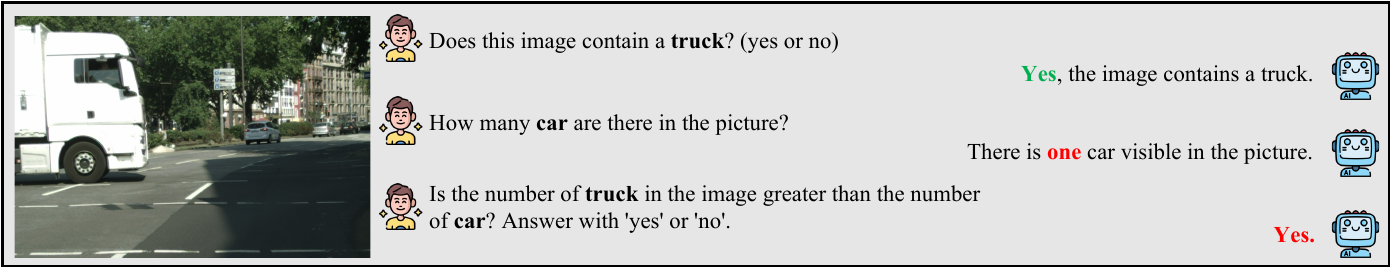}
   \caption{Basic-to-Advanced Progression Metric Example. Basic-to-Advanced Progression Metric covers three problems: existential, counting, and logical reasoning problems from top to bottom.}
   \label{fig: Conservation}
\end{figure*}

\begin{figure}[t]
    \centering
    \includegraphics[width=0.8\linewidth]{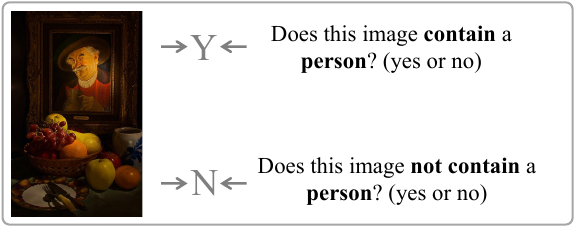}
    \caption{The \textbf{OODBench} example has two questions and one instance, with the answers alternated to avoid the model being over-scored due to biased output distribution.}
    \label{fig: contain question}
\end{figure}

\subsection{Evaluation Setup}
In our main results, in order to comprehensively evaluate the performance of the model, we selected several commonly used evaluation metrics, including Accuracy, F1 Score, Precision, Recall, and Matthews Correlation Coefficient (MCC). In addition, in the experiments of BAP Metric, we introduced three additional metrics to refine the evaluation further. Specifically, we define the `Existential Accuracy` (E-Acc) metric, where a point is awarded only if the model correctly answers the existential question. The `Counting Accuracy` (C-Acc) metric, on the other hand, requires the model to obtain a point only when it correctly answers the number of specified categories in the image. Finally, we award a point for `Logical Accuracy` (L-Acc) when the model correctly answers both counting and logical questions in the test sample. Regarding the scoring models and the hyperparameter $T$, we conducted systematic ablation studies to determine the optimal value of $T$ as well as the type and number of scoring models. We employ contrastive learning models CLIP~\cite{radford2021learning} and BLIP-2~\cite{li2023blip} as scoring models.

\textbf{Basic-to-Advanced Progression (BAP) Metric.}
To comprehensively evaluate the impact of OOD data on VLMs from different dimensions, we meticulously devised the Basic-to-Advanced Progression Metric. Specifically, as shown in Fig.~\ref{fig: Conservation}, we examine the recognition ability, counting perception ability, and reasoning ability of VLMs by constructing existential, counting, and logical reasoning problems. The existential question is the essential reasoning task. It is usually the precursor to other questions, which require the model to make a judgment about whether a particular category of objects is present in the image. The counting problem is a further judgment about the number of objects in the image after the model has correctly answered the existential question. This stage requires the model to return a specific numerical value rather than a simple `yes` or `no`. In the third stage, we evaluate the reasoning ability of the model by asking questions about different objects in the same image (\textit{e.g.}, \textit{Is the number of [class 1] in the image greater than the number of [class 2]? Answer with `yes` or `no`}). Logical reasoning problems build on counting problems, which require the model to be aware not only of the number of objects in each category but also of making comparisons between the number of categories. Progressive questioning helps to improve the interpretability of the model's reasoning process. Each step of the answer is based on clear logic, which helps the user understand the thinking process of the model. Finally, we calculated $C/N \times 100\%$ as the BAP Metric, where $C$ is the number of questions answered correctly, and $N$ is the total number of questions.

\begin{table*}[!t]
\centering
\caption{\textbf{Performance on OODBench.} We report the performance of the \textbf{10} leading VLMs on OODBench. All models perform significantly lower on OOD-H than on ID. The performance of the latest models, such as LLaVA-NeXT, Llama3.2-Vision, InternVL2.5, and Qwen2-VL, lags behind the ID data by about $20\%$ to $30\%$ on OOD-H data. Even the best closed-source model, GPT-4o, still performs $26\%$ lower on OOD-H than ID.}
\resizebox{15cm}{!}{
\begin{tabular}{cccp{2.8cm}ccccc}
\toprule 
\hline
\multirow{2}{*}{\centering Model} & \centering \multirow{2}{*}{Image Encoder} & \centering \multirow{2}{*}{ Language Model} & \centering \multirow{2}{*}{Data Type} & \multicolumn{5}{c}{\textbf{OODBench} Performance} \\
\cline{5-9}
  & & & & Accuracy($\%$) & F1($\%$) & Precision($\%$) & Recall($\%$) & MCC($\%$) \\
 \hline
 \centering Random Chance & - & - & ID/OOD-S/OOD-H & 50.00 & 50.00 & 50.00 & 50.00 & 0.00\\
 \hline

 & & \centering \textbf{Open-source Models} & & & & & &\\

\hline

& & & \multicolumn{1}{c}{ID} & 84.79 & 84.72 & 85.15 & 84.28 & 69.59\\
 & & & \multicolumn{1}{c}{OOD-S} & 64.54 & 56.16 & 73.52 & 45.44 & 31.46\\
 & & & \multicolumn{1}{c}{OOD-H} & 63.22 & 52.99 & 73.40 & 41.46 & 29.36\\
 & & & \multicolumn{1}{c}{$\text{ID}^{\text{CoT}}$} & 64.96 & 70.74 & 60.72 & 84.72 & 32.57\\
 & & & \multicolumn{1}{c}{$\text{OOD-S}^{\text{CoT}}$} & 59.66 & 55.49 & 61.88 & 50.29 & 19.66\\
\multirow{-6}{*}{ LLaVA-NeXT-8B} & \multirow{-6}{*}{ CLIP-L-14} & \multirow{-6}{*}{Llama-3-8B-Instruct} & \multicolumn{1}{c}{$\text{OOD-H}^{\text{CoT}}$} & 64.40 & 57.41 & 71.44 & 47.99 & 30.49\\

\rowcolor{gray!20}
 & & & \multicolumn{1}{c}{ID} & 89.14 & 89.02 & 90.07 & 87.98 & 78.31\\
\rowcolor{gray!20}
 & & & \multicolumn{1}{c}{OOD-S} & 64.36 & 58.19 & 70.38 & 49.60 & 30.07\\
 \rowcolor{gray!20}
& & & \multicolumn{1}{c}{OOD-H} & 61.14 & 52.82 & 67.21 & 43.50 & 23.81\\
\rowcolor{gray!20}
 & & & \multicolumn{1}{c}{$\text{ID}^{\text{CoT}}$} & 85.14 & 84.59 & 87.82 & 81.60 & 70.46\\
\rowcolor{gray!20}
 & & & \multicolumn{1}{c}{$\text{OOD-S}^{\text{CoT}}$} & 63.15 & 56.40 & 69.05 & 47.67 & 27.66\\
 \rowcolor{gray!20}
\multirow{-6}{*}{\centering DeepSeek-VL-7B-Chat} & \multirow{-6}{*}{\centering SigLIP-L $\&$ SAM-B} & \multirow{-6}{*}{\centering DeepSeek-LLM-7B} & \multicolumn{1}{c}{$\text{OOD-H}^{\text{CoT}}$} & 57.44 & 48.27 & 61.53 & 39.72 & 15.92\\

 & & & \multicolumn{1}{c}{ID} & 84.81 & 83.07 & 74.51 & 93.84 & 71.15\\
 & & & \multicolumn{1}{c}{OOD-S} & 63.03 & 49.96 & 36.91 & 77.28 & 30.56\\
& & & \multicolumn{1}{c}{OOD-H} & 60.68 & 43.58 & 30.38 & 77.10 & 26.85\\
 & & & \multicolumn{1}{c}{$\text{ID}^{\text{CoT}}$} & 82.95 & 79.50 & 66.13 & 99.64 & 69.96\\
 & & & \multicolumn{1}{c}{$\text{OOD-S}^{\text{CoT}}$} & 59.81 & 40.17 & 26.98 & 78.58 & 26.02\\
\multirow{-6}{*}{\centering DeepSeek-VL2-Small} & \multirow{-6}{*}{\centering SigLIP-SO400M} & \multirow{-6}{*}{\centering DeepSeekMoE LLM} & \multicolumn{1}{c}{$\text{OOD-H}^{\text{CoT}}$} & 56.04 & 29.82 & 18.68 & 73.92 & 18.19\\

\rowcolor{gray!20}
 & & & \multicolumn{1}{c}{ID} & 90.93 & 90.75 & 92.51 & 89.06 & 81.91\\
 \rowcolor{gray!20}
 & & & \multicolumn{1}{c}{OOD-S} & 69.56 & 67.58 & 72.27 & 63.46 & 39.40\\
 \rowcolor{gray!20}
 & & & \multicolumn{1}{c}{OOD-H} & 59.05 & 55.87 & 60.58 & 51.85 & 18.30\\
 \rowcolor{gray!20}
 & & & \multicolumn{1}{c}{$\text{ID}^{\text{CoT}}$} & 93.18 & 92.75 & 98.95 & 87.29 & 86.97\\
 \rowcolor{gray!20}
 & & & \multicolumn{1}{c}{$\text{OOD-S}^{\text{CoT}}$} & 66.90 & 59.08 & 77.36 & 47.79 & 36.58\\
 \rowcolor{gray!20}
 \multirow{-6}{*}{\centering InternVL2-8B} & \multirow{-6}{*}{\centering InternViT-300M-448px} & \multirow{-6}{*}{\centering InternLM2.5-7b-chat} & \multicolumn{1}{c}{$\text{OOD-H}^{\text{CoT}}$} & 69.26 & 61.11 & 83.17 & 48.30 & 42.44\\

 & & & \multicolumn{1}{c}{ID} & 91.66 & 91.79 & 90.37 & 93.26 & 83.37\\
 & & & \multicolumn{1}{c}{OOD-S} & 72.37 & 71.51 & 73.82 & 69.34 & 44.83\\
& & & \multicolumn{1}{c}{OOD-H} & 58.49 & 57.91 & 58.73 & 57.11 & 16.98\\
 & & & \multicolumn{1}{c}{$\text{ID}^{\text{CoT}}$} & 90.49 & 90.56 & 89.89 & 91.24 & 80.99\\
 & & & \multicolumn{1}{c}{$\text{OOD-S}^{\text{CoT}}$} & 69.87 & 65.93 & 75.84 & 58.30 & 40.84\\
\multirow{-6}{*}{\centering InternVL2.5-8B} & \multirow{-6}{*}{\centering InternViT-300M-448px-V2.5} & \multirow{-6}{*}{\centering InternLM2.5-7b-chat} & \multicolumn{1}{c}{$\text{OOD-H}^{\text{CoT}}$} & 67.09 & 63.57 & 71.19 & 57.42 & 34.84\\

\rowcolor{gray!20}
 & & & \multicolumn{1}{c}{ID} & 92.73 & 92.87 & 91.16 & 94.65 & 85.53\\
 \rowcolor{gray!20}
 & & & \multicolumn{1}{c}{OOD-S} & 68.65 & 68.42 & 68.94 & 67.91 & 37.31\\
 \rowcolor{gray!20}
 & & & \multicolumn{1}{c}{OOD-H} & 66.26 & 67.55 & 65.07 & 70.22 & 32.63\\
 \rowcolor{gray!20}
 & & & \multicolumn{1}{c}{$\text{ID}^{\text{CoT}}$} & 74.49 & 76.31 & 71.23 & 82.18 & 49.57\\
 \rowcolor{gray!20}
 & & & \multicolumn{1}{c}{$\text{OOD-S}^{\text{CoT}}$} & 59.80 & 58.26 & 60.59 & 56.10 & 19.66\\
 \rowcolor{gray!20}
 \multirow{-6}{*}{\centering Llama-3.2-11B-Vision} & \multirow{-6}{*}{\centering ViT-H-14 } & \multirow{-6}{*}{\centering Llama3.1-8B} & \multicolumn{1}{c}{$\text{OOD-H}^{\text{CoT}}$} & 61.14 & 62.20 & 60.55 & 63.94 & 22.31\\

 & & & \multicolumn{1}{c}{ID} & 90.67 & 90.92 & 88.52 & 93.46 & 81.47\\
 & & & \multicolumn{1}{c}{OOD-S} & 73.01 & 70.79 & 77.14 & 65.40 & 46.56\\
& & & \multicolumn{1}{c}{OOD-H} & 66.77 & 63.19 & 70.82 & 57.05 & 34.20\\
 & & & \multicolumn{1}{c}{$\text{ID}^{\text{CoT}}$} & 90.65 & 90.92 & 88.32 & 93.69 & 81.45\\
 & & & \multicolumn{1}{c}{$\text{OOD-S}^{\text{CoT}}$} & 71.61 & 69.27 & 75.50 & 63.99 & 43.74\\
\multirow{-6}{*}{\centering Qwen2-VL-7B-Instruct} & \multirow{-6}{*}{\centering CLIP-L-14} & \multirow{-6}{*}{\centering Qwen2-7B} & \multicolumn{1}{c}{$\text{OOD-H}^{\text{CoT}}$} & 65.23 & 62.10 & 68.24 & 56.97 & 30.87\\

\rowcolor{gray!20}
 & & & \multicolumn{1}{c}{ID} & 90.35 & 90.14 & 88.31 & 92.06 & 80.76\\
\rowcolor{gray!20}
 & & & \multicolumn{1}{c}{OOD-S} & 64.65 & 56.67 & 46.25 & 73.17 & 31.50\\
 \rowcolor{gray!20}
& & & \multicolumn{1}{c}{OOD-H} & 68.70 & 61.31 & 49.61 & 80.24 & 40.46\\
\rowcolor{gray!20}
 & & & \multicolumn{1}{c}{$\text{ID}^{\text{CoT}}$} & 93.29 & 92.81 & 86.60 & 99.97 & 87.36\\
\rowcolor{gray!20}
 & & & \multicolumn{1}{c}{$\text{OOD-S}^{\text{CoT}}$} & 68.27 & 58.75 & 45.18 & 83.94 & 41.19\\
 \rowcolor{gray!20}
\multirow{-6}{*}{\centering Qwen2.5-VL-7B-Instruct} & \multirow{-6}{*}{\centering ViT(SwiGLU, RMSNorm)} & \multirow{-6}{*}{\centering Qwen2.5 LLM} & \multicolumn{1}{c}{$\text{OOD-H}^{\text{CoT}}$} & 68.93 & 59.31 & 45.29 & 85.92 & 42.97\\
 \hline
  & & \centering \textbf{Closed-source Models} & & & & & &\\

\hline

 & & & \multicolumn{1}{c}{ID} & 91.12 & 91.29 & 89.52 & 93.14 & 82.30\\
 & & & \multicolumn{1}{c}{OOD-S} & 71.78 & 71.87 & 71.64 & 72.10 & 43.56\\
 & & & \multicolumn{1}{c}{OOD-H} & 63.54 & 64.38 & 62.93 & 65.89 & 27.11\\
 & & & \multicolumn{1}{c}{$\text{ID}^{\text{CoT}}$} & 94.37 & 94.30 & 95.56 & 93.07 & 88.77\\
 & & & \multicolumn{1}{c}{$\text{OOD-S}^{\text{CoT}}$} & 69.36 & 65.34 & 75.22 & 57.74 & 39.82\\
\multirow{-6}{*}{\centering Gemini} & \multirow{-6}{*}{\centering -} & \multirow{-6}{*}{\centering -} & \multicolumn{1}{c}{$\text{OOD-H}^{\text{CoT}}$} & 73.45 & 71.02 & 78.16 & 65.07 & 47.56\\

 \rowcolor{gray!20}
 & & & \multicolumn{1}{c}{ID} & 91.95 & 92.24 & 89.04 & 95.68 & 84.14\\
 \rowcolor{gray!20}
 & & & \multicolumn{1}{c}{OOD-S} & 73.37 & 74.13 & 72.05 & 76.34 & 46.81\\
 \rowcolor{gray!20}
 & & & \multicolumn{1}{c}{OOD-H} & 65.13 & 67.06 & 63.54 & 71.00 & 30.47\\
 \rowcolor{gray!20}
 & & & \multicolumn{1}{c}{$\text{ID}^{\text{CoT}}$} & 78.13 & 80.94 & 71.72 & 92.87 & 58.87\\
 \rowcolor{gray!20}
 & & & \multicolumn{1}{c}{$\text{OOD-S}^{\text{CoT}}$} & 62.44 & 61.49 & 63.08 & 59.98 & 24.91\\
 \rowcolor{gray!20}
 \multirow{-6}{*}{\centering GPT-4o} & \multirow{-6}{*}{\centering - } & \multirow{-6}{*}{\centering GPT-4 } & \multicolumn{1}{c}{$\text{OOD-H}^{\text{CoT}}$} & 63.62 & 64.65 & 62.88 & 66.52 & 27.29\\
 \hline
\bottomrule
\end{tabular}
}

\label{tab:Main_Results}
\end{table*}

\subsection{Main Results Analysis}
Through our OOD data collection process, we divide instance-level OOD data into distinct subsets. Based on this process, ID data are obtained by randomly pairing images and categories and excluding pairs identified as OOD. As shown in Fig.~\ref{fig: contain question}, we present two questions for each instance respectively (\textit{Does this image \textbf{contain} a [class]? (yes or no)} and \textit{Does this image \textbf{not contain} a [class]? (yes or no)}) and assign the opposite label to each question. This strategy effectively balances the label distributions of different instance data. It mitigates the tendency of VLMs to favor either `yes` or `no` responses due to training data bias. As the task format superficially resembles prompt templates used in some hallucination detection work, we systematically compare hallucination data with OODBench in Appendix~\ref{appendix: hallucination} across three dimensions—definition, generation mechanism, and experimental results—to clarify their fundamental differences.

Tab. \ref{tab:Main_Results} shows the performance of the current state-of-the-art LLaVA-NeXT~\cite{li2024llava}, DeepSeek-VL~\cite{lu2024deepseekvl}, InternVL2~\cite{chen2024far}, InternVL2.5~\cite{chen2024expanding}, Llama-3.2-Vision~\cite{dubey2024llama}, Qwen2-VL~\cite{Qwen2VL}, Gemini~\cite{team2024gemini} and GPT-4o~\cite{achiam2023gpt, hurst2024gpt} models on \textbf{OODBench}. The results indicate that most VLMs exhibit a $20\%$ to $30\%$ accuracy drop on OOD-H data relative to ID data. In addition, all models show a decreasing trend in accuracy on ID, OOD-S, and OOD-H data. Essentially, OOD-S data is data that is only identified as OOD data by one scoring model and not identified as OOD data by another scoring model. In contrast, OOD-H data is identified as OOD data by both scoring models. These results show decreasing accuracy on ID, OOD-S, and OOD-H data, suggesting that multiple well-pretrained VLMs are more effective scoring models for collecting data aligned with real-world scenarios than a single scoring model. Moreover, this process requires no extra human effort, minimizing labor costs. Notably, VLMs (\textit{e.g.}, LLaVA-NeXT, DeepSeek-VL, InternVL2/2.5, Qwen2-VL) show poor recall on OOD-H data, with LLaVA-NeXT and DeepSeek-VL even falling below $50\%$ random chance. By contrast, recall on ID data remains around $90\%$, yielding a $26$–$43\%$ drop on OOD-H.

These results suggest that instances in OOD data are more likely to be missed than those in ID data. In safety-critical settings, low recall can be particularly problematic. For example, an autonomous driving system must detect all hazards—pedestrians, vehicles, or traffic signs, since missing any can greatly increase accident risk. Appendix~\ref{appendix: sub-dataset performance} details VLM performance on natural and driving datasets, illustrating how OOD data affect model behavior across scenarios.

\begin{table}
\centering
\caption{Performance of \textbf{10} leading VLMs on \textbf{OODBench} for Basic-to-Advanced Progress.}
\resizebox{8.0cm}{!}{
\begin{tabular}{ccccc} 
\toprule 
\hline
\centering \multirow{2}{*}{Model}  & \centering \multirow{2}{*}{Data Type} & \multicolumn{3}{c}{\textbf{OODBench BAP} Performance} \\
\cline{3-5}
  & & E-Acc($\%$) & C-Acc($\%$) & L-Acc($\%$) \\
 \hline
\midrule 
 & ID & 81.39 & 38.25 & 21.70\\
 & OOD-S & 66.61 & 33.75 & 21.48\\
\multirow{-3}{*}{\centering LLaVA-NeXT-8B} & OOD-H & 44.22 & 21.71 & 13.45 \\
  \rowcolor{gray!20}
 & ID & 84.66 & 38.67 & 24.40\\
  \rowcolor{gray!20}
 & OOD-S & 69.73 & 26.72 & 17.50\\
\rowcolor{gray!20}
\multirow{-3}{*}{\centering DeepSeek-VL-7B-Chat} & OOD-H & 46.25 & 22.34 & 14.46 \\
 & ID & 70.77 & 24.32 & 15.04\\
 & OOD-S & 63.54 & 30.75 & 17.96\\
\multirow{-3}{*}{\centering DeepSeek-VL2-Small} & OOD-H & 37.74 & 20.32 & 12.09 \\
 \rowcolor{gray!20}
 & ID & 87.57 & 51.07 & 33.35\\
 \rowcolor{gray!20}
 & OOD-S & 73.20 & 40.43 & 26.62\\
 \rowcolor{gray!20}
\multirow{-3}{*}{\centering InternVL2-8B} & OOD-H & 56.72 & 40.20 & 28.43 \\
 & ID & 91.34 & 45.63 & 32.42\\
 & OOD-S & 76.89 & 38.68 & 29.03\\
\multirow{-3}{*}{\centering InternVL2.5-8B} & OOD-H & 61.62 & 35.40 & 26.33 \\
 
 \rowcolor{gray!20}
 & ID & 94.57 & 47.27 & 28.44\\
 \rowcolor{gray!20}
 & OOD-S & 77.99 & 38.95 & 22.49\\
 \rowcolor{gray!20}
\multirow{-3}{*}{\centering Llama-3.2-11B-Vision-Instruct} & OOD-H & 74.33 & 31.58 & 18.56 \\
 & ID & 91.73 & 46.13 & 30.01\\
 & OOD-S & 75.31 & 42.36 & 30.72\\
\multirow{-3}{*}{\centering Qwen2-VL-7B-Instruct} & OOD-H & 59.77 & 33.44 & 23.88 \\
  \rowcolor{gray!20}
 & ID & 86.34 & 37.38 & 32.75\\
  \rowcolor{gray!20}
 & OOD-S & 73.39 & 39.25 & 28.13\\
  \rowcolor{gray!20}
\multirow{-3}{*}{\centering Qwen2.5-VL-7B-Instruct} & OOD-H & 56.52 & 33.67 & 24.39 \\
\hline
 & ID & 91.43 & 58.36 & 35.56\\
 & OOD-S & 77.84 & 52.20 & 34.12\\
\multirow{-3}{*}{\centering Gemini} & OOD-H & 72.28 & 42.82 & 27.86 \\
  \rowcolor{gray!20}
 & ID & 94.84 & 48.26 & 30.83\\
  \rowcolor{gray!20}
 & OOD-S & 81.26 & 41.43 & 27.93\\
  \rowcolor{gray!20}
\multirow{-3}{*}{\centering GPT-4o} & OOD-H & 76.38 & 36.00 & 24.36 \\
\bottomrule
\end{tabular}}

\label{tab: BAP}
\end{table}

\subsection{Does Chain-of-Thought Reasoning Work on OOD Data?}

Chain-of-Thought (CoT) prompting encourages models to generate intermediate steps before producing an answer, and has been reported to improve performance on reasoning-style tasks. However, when confronted with OOD data, models often fail to reason accurately using their existing knowledge, as such data fall outside the training distribution. A natural question is whether CoT prompting still helps in these cases. Specifically, does explicitly expanding the inference process help models handle unfamiliar inputs? To investigate this, we add a CoT prompt requiring step-by-step analysis and evaluate whether it improves VLM performance on OOD data. Our prompt setup mirrors the main experiment, with an added requirement for stepwise analysis: \textit{e.g.}, \textit{Does this image contain [class]? Yes or No? Let's break down the information step by step.} and \textit{Does this image not contain [class]? Yes or No? Let's break down the information step by step.}

Table~\ref{tab:Main_Results} reports the performance of different VLMs with CoT prompting. Gemini~\cite{team2024gemini}, InternVL2~\cite{chen2024far}, and InternVL2.5~\cite{chen2024expanding} exhibit similar improvements: on challenging OOD-H samples, accuracy increases by about 10\% over baseline; ID accuracy remains stable or rises slightly; OOD-S accuracy declines by 2--3\% on average. Precision gains 15.23\%, 22.59\%, and 12.46\% for Gemini, InternVL2, and InternVL2.5, respectively, while recall is unchanged or marginally lower. In contrast, LLaVA-NeXT~\cite{li2024llava}, Qwen2-VL~\cite{Qwen2VL}, and Qwen2.5-VL~\cite{bai2025qwen2} show negligible OOD-H gains: Qwen2-VL remains stable on ID and OOD-S, whereas LLaVA-NeXT’s ID accuracy drops from 84.79\% to 64.96\%. Moreover, GPT-4o~\cite{achiam2023gpt, hurst2024gpt}, DeepSeek-VL~\cite{lu2024deepseekvl}, DeepSeek-VL2~\cite{wu2024deepseek}, and Llama-3.2-Vision~\cite{dubey2024llama} suffer overall performance degradation after CoT prompting: GPT-4o and Llama-3.2-Vision ID accuracy falls by 13.82\% and 18.27\%, respectively, and OOD-H accuracy declines by 1.5--5\%. Hence, CoT yields negative gains for these models.

\subsection{BAP Evaluation of ID and OOD Data Differences}
To assess VLM performance on OOD data across multiple dimensions, we introduce the Basic-to-Advanced Progression (BAP) metric, which evaluates model performance in recognition, counting, and reasoning tasks. Specifically, we define three metrics—Existential Accuracy (E-Acc), Counting Accuracy (C-Acc), and Logical Accuracy (L-Acc)—using a progressive questioning framework to assess existential, counting, and logical challenges, respectively.

For BAP evaluation, we select images containing at least two distinct labels from the ID, OOD-S, and OOD-H datasets. Each image is paired with two distinct category labels (without duplication), forming test samples with paired labels from different categories. The quantity labels for counting tasks are obtained by counting the number of target object instances in the source annotations.

As task difficulty increases from E-Acc to C-Acc to L-Acc, all models exhibit a consistent decline in performance. This trend becomes more pronounced as the data distribution shifts from ID to OOD-S and OOD-H, corresponding to increasing divergence from the training distribution and resulting in varying degrees of performance degradation across metrics. Most models achieve robust results on ID data, particularly on E-Acc and C-Acc. However, performance declines as the data transitions to OOD-S and further to OOD-H, with the largest drop observed on L-Acc. On more challenging OOD data (\textit{e.g.}, OOD-H), models exhibit a substantial decrease in logical accuracy.

E-Acc mainly tests recognition without complex reasoning. On OOD-H data, GPT-4o~\cite{achiam2023gpt, hurst2024gpt} shows the best recognition, correctly identifying category existence in images. In contrast, C-Acc evaluates the ability to count object instances, requiring both recognition and accurate counting. Gemini~\cite{team2024gemini} leads with $42.82\%$, underscoring its relative strength in counting.
L-Acc evaluates logical reasoning by comparing category counts (\textit{e.g.}, \emph{Is A greater than B?}). This task couples counting with higher-level reasoning, and all models show a sharp drop in performance, highlighting the challenge of logical reasoning and quantitative relations in OOD settings.

\subsection{Ablation study of different types of generalized scoring models}
\label{appendix: ood detector type ablation}
In order to explore the impact of different types of scoring models, we designed and conducted a series of systematic ablation experiments. In the specific setup, we selected GroupViT~\cite{xu2022groupvit} and LiT~\cite{zhai2022lit} to gradually replace the original CLIP~\cite{radford2021learning} with BLIP2~\cite{li2023blip} to re-divide the samples in the COCO dataset into OOD, and we constructed two new sets of division data. Subsequently, we input the three sets of delineation results (\textit{i.e.}, original division + division of the two alternative scoring models) into two open-source multimodal models (InternVL2, Qwen2-VL) and two closed-source models (Gemini, GPT-4o), respectively, for consistency testing to assess the impact of different scoring models on the downstream performance performance.

It should be noted that in order to control the testing overhead, the original OODBench benchmark introduces a random downsampling strategy when the number of samples exceeds 6,000. In order to avoid the possible data distribution bias introduced by downsampling, we retain all the samples judged as OOD, \textit{i.e.}, we do not perform the downsampling operation again when we use GroupViT with LiT for reclassification. At the same time, in order to ensure the fairness of the experiment, we continue to use the sample sets corresponding to the CLIP and BLIP2 classification results in the initial benchmark to ensure the consistency of the processing flow among the three groups.

For ease of presentation, we define the three datasets as follows:
\begin{itemize}
    \item The data divided by CLIP and BLIP2 are recorded as Group \textbf{I}.
    \item The data divided by BLIP2 and GroupViT are recorded as Group \textbf{II}.
    \item The data divided by GroupViT and LiT are recorded as Group \textbf{III}.
\end{itemize}

\begin{table*}[ht]
\centering
\caption{Impact of replacing scoring models on the downstream performance of different multimodal models. We compare the accuracy differences ($\Delta$) of data divided by different combinations of scoring models—Group \textbf{I}, Group \textbf{II}, and Group \textbf{III}—across four representative VLMs (Qwen2-VL-7B-Instruct, InternVL2-8B, Gemini, and GPT-4o). Group \textbf{I} is used as the baseline, and the performance drop or gain of Group \textbf{II} and Group \textbf{III} is reported accordingly. The results show that although different scoring models induce slight variance in OOD data division, their impact on downstream model performance is minimal, supporting the replaceability of scoring models in the OODBench framework.}
\resizebox{13.5cm}{!}{
\begin{tabular}{ccccccccc}
\toprule 
\hline
\centering \multirow{2}{*}{Model} & \centering \multirow{2}{*}{Scoring Models} & \multicolumn{6}{c}{\textbf{COCO-OOD} Performance} \\
\cline{3-9}
  & & Num & Accuracy($\%$) & F1($\%$) & Precision($\%$) & Recall($\%$) & MCC($\%$) & Time\\
 \hline
 \centering Random Chance & - & - & $\text{50.00}\%$ & $\text{50.00}\%$ & $\text{50.00}\%$ & $\text{50.00}\%$ & $\text{0.00}\%$ & -\\
 
 \hline
\textbf{Open-source Models} & & & & & & \\
 \hline

 & & 10160 & 78.74 & 76.78 & 84.58 & 70.30 & 58.32 & 00:11:44$^\text{\textbf{\textit{2}}}$ \\
 & \multirow{-2}{*}{CLIP \& BLIP2} & $\Delta$ & + 0.00 & + 0.00 & + 0.00 & + 0.00 & + 0.00 & -  \\
& & 9988 & 79.85 & 78.06 & 85.65 & 71.71 & 60.50 & 00:21:40$^\text{\textbf{\textit{1}}}$  \\
 & \multirow{-2}{*}{BLIP2 \& GroupViT} & $\Delta$ & -1.11 & -1.28 & -1.07 & -1.41 & -2.18 & -\\
 & & 12090 & 80.17 & 78.46 & 85.88 & 72.22 & 61.12 & 00:26:15$^\text{\textbf{\textit{1}}}$  \\
\multirow{-6}{*}{Qwen2-VL-7B-Instruct} & \multirow{-2}{*}{GroupViT \& LiT} & $\Delta$ & -1.43 & -1.68 & -1.30 & -1.92 & -2.80 & -\\

\rowcolor{gray!20}
 & & 10160 & 66.22 & 65.99 & 66.44 & 65.55 & 32.44 & 01:03:50$^\text{\textbf{\textit{2}}}$ \\
\rowcolor{gray!20}
 & \multirow{-2}{*}{CLIP \& BLIP2} & $\Delta$ & +0.00 & +0.00 & +0.00 & +0.00 & +0.00 & -  \\
 \rowcolor{gray!20}
& & 9988 & 67.08 & 66.91 & 67.25 & 66.58 & 34.16 & 01:49:45$^\text{\textbf{\textit{1}}}$  \\
\rowcolor{gray!20}
 & \multirow{-2}{*}{BLIP2 \& GroupViT} & $\Delta$ & -0.86 & -0.92 & -0.81 & -1.03 & -1.72& -\\
\rowcolor{gray!20}
 & & 12090 & 68.17 & 67.91 & 68.48 & 67.34 & 36.35 & 02:13:10$^\text{\textbf{\textit{1}}}$  \\
 \rowcolor{gray!20}
\multirow{-6}{*}{\centering InternVL2-8B} & \multirow{-2}{*}{GroupViT \& LiT} & $\Delta$ & -1.95 & -1.92 & -2.04 & -1.79 & -3.91 & -\\

 \hline

 & & 10150 & 67.02 & 68.71 & 65.37 & 72.41 & 34.25 & 02:59:57 \\
 & \multirow{-2}{*}{CLIP \& BLIP2} & $\Delta$ & +0.00 & +0.00 & +0.00 & +0.00 & +0.00 & -  \\
& & 9972 & 66.65 & 68.36 & 65.02 & 72.06 & 33.49 & 03:28:29  \\
 & \multirow{-2}{*}{BLIP2 \& GroupViT} & $\Delta$ & +0.37 & +0.35 & +0.35 & +0.35 & +0.76 & -\\
 & & 12072 & 66.96 & 68.76 & 65.22 & 72.71 & 34.16 & 04:04:54\\
\multirow{-6}{*}{Gemini} & \multirow{-2}{*}{GroupViT \& LiT} & $\Delta$ & +0.06 & -0.05 & +0.15 & -0.30 & +0.09 & -\\

\rowcolor{gray!20}
 & & 10158 & 73.66 & 74.35 & 72.44 & 76.37 & 47.38 & 09:58:34 \\
\rowcolor{gray!20}
 & \multirow{-2}{*}{CLIP \& BLIP2} & $\Delta$ & +0.00 & +0.00 & +0.00 & +0.00 & +0.00 & -  \\
 \rowcolor{gray!20}
& & 9908 & 73.74 & 74.57 & 72.28 & 77.01 & 47.58 & 12:32:43  \\
\rowcolor{gray!20}
 & \multirow{-2}{*}{BLIP2 \& GroupViT} & $\Delta$ & -0.08 & -0.22 & +0.16 & -0.64 & -0.20 & -\\
\rowcolor{gray!20}
 & & 12090 & 75.86 & 76.56 & 74.39 & 78.86 & 51.81 & 08:41:31  \\
 \rowcolor{gray!20}
\multirow{-6}{*}{GPT-4o \footnotemark[3]} & \multirow{-2}{*}{GroupViT \& LiT} & $\Delta$ & -2.20 & -2.21 & -1.95 & -2.49 & -4.43 & -\\

 \hline

\bottomrule
\end{tabular}
}

\label{tab: replace ood detector}
\end{table*}

\footnotetext[3]{There may be some differences between the performance results of GPT-4o in this experiment and those in Tab.~\ref{tab: coco}, which mainly stems from the differences in the model versions used. To ensure the fairness and consistency of the results of the scoring model type ablation experiments, we uniformly used the same version of GPT-4o (20240806) for evaluation in this experiment.}

We first evaluated Qwen2-VL by taking the performance of group \textbf{I} data as the baseline and calculating the accuracy difference ($\Delta$) between group \textbf{II} and group \textbf{III} data under the model relative to the baseline. The experimental results show that on Qwen2-VL, the accuracy of group II and group \textbf{III} data decreases by 1.11\% and 1.43\%, respectively, compared with that of group \textbf{I}. On Gemini, it improves by 0.86\% and 0.06\%, respectively, and the performance difference between the three groups of data on InternVL and GPT-4o is also minimal, and the details of the values are shown in Tab.~\ref{tab: replace ood detector}. In the Time column, the time superscript indicates the number of graphics cards used in the test, and we used NVIDIA L40s graphics cards in the testing phase.

Consider a class of vision-language models that learn a compatibility score $S_\theta(x,c)$ via cross-modal alignment (\textit{e.g.}, contrastive learning or its variants). If the learned score satisfies Eq.~\ref{eq:PMI_parameterized},
where $f^*(x,c)$ is a monotone transform of $\text{PMI}(x,c)$, then for sufficiently small approximation error $\epsilon_\theta$, the ranking induced by $S_\theta(x,c)$ preserves that of mutual information. Models such as LiT and GroupViT fall into this class, as they: Optimize log-linear compatibility between image and text representations;
Approximate $\log p(c\mid x)$ up to a normalization constant; Differ only in parameterization (frozen encoder, latent grouping), not in the underlying statistical objective.

Therefore, they inherit the monotonic relationship between contrastive score and semantic mutual information. The specific training paradigm is not essential; what matters is whether the model learns a log-likelihood–like compatibility function between image and text. As long as this condition holds, the resulting score inherently preserves the ranking induced by semantic mutual information.



\begin{figure*}[ht]
    \centering
        \includegraphics[width=0.8\linewidth]{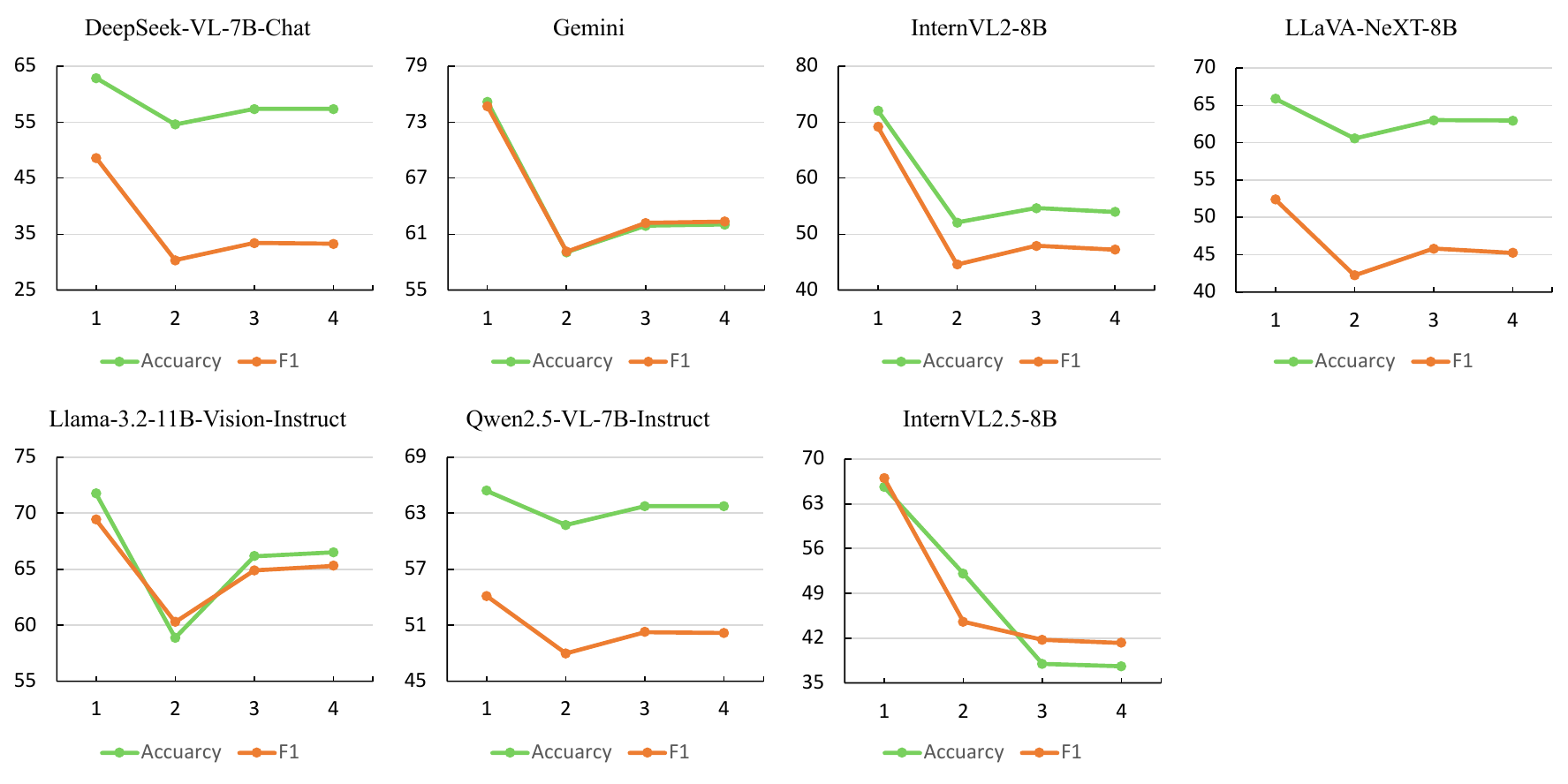}
    \caption{Impact of the number of scoring models on the quality of OOD data division under the cross-validation mechanism. The horizontal axis indicates the number of scoring models involved in the division process (ranging from 1 to 4), while the vertical axis shows the corresponding Accuracy and F1 score (\%) obtained by evaluating the divided OOD data on multiple mainstream VLMs. The results demonstrate that using a single scoring model yields the highest performance but suffers from potential bias. Incorporating two scoring models significantly reduces this bias, producing OOD data more aligned with the VLMs’ potential OOD distribution. However, increasing the number of scoring models beyond two offers limited benefits and may introduce instability, as evidenced by performance rebound and eventual convergence.}
    \label{fig: ood detector_num_ablation}
\end{figure*}

\subsection{Ablation studies with different numbers of scoring models}
\label{appendix: ood detector num ablation}
To further investigate the effect of different numbers of scoring models on the division of OOD data under the cross-validation mechanism, we conducted ablation experiments on the number of scoring models. We selected four models, CLIP~\cite{radford2021learning}, BLIP2~\cite{li2023blip}, GroupViT~\cite{xu2022groupvit}, and LiT~\cite{zhai2022lit}, as scoring models, and sequentially increased the number of scoring models involved in the division in a fixed order. Consistent with the scoring models type ablation experiment~\ref{appendix: ood detector type ablation}, to avoid introducing additional bias, we did not perform downsampling operations on the OOD data divided by the above models to maintain the consistency of the division process. To assess the impact of the number of scoring models on the OOD data division results, we tested the OOD data divided by different numbers of scoring models on several mainstream VLMs. We visualised the results in the form of a line graph (shown in Figure~\ref{fig: ood detector_num_ablation}). The horizontal axis of the graph represents the number of scoring models involved in the division, and the vertical axis represents the accuracy and F1 score obtained from testing on each VLM.

The experimental results show that when only a single scoring model (number 1) is used, each VLM reaches the highest value in both Accuracy and F1, which indicates that a single scoring model may bring about model bias, which in turn leads to a mismatch between the divided OOD data and the potential OOD distribution of the large VLMs. When the number of scoring models is increased to 2, the performance of each large VLM decreases significantly, indicating that the cross-validation mechanism can mitigate the error brought about by single scoring model bias to some extent, thus obtaining data that more closely fits the OOD distribution of the large VLM. However, when the number of scoring models is further increased to 3, the performance of all six VLMs except InternVL2.5-8B shows a performance rebound, \textit{i.e.}, the data divided by three scoring models generally performs higher than that of 2 scoring models in terms of Accuracy and F1. Nonetheless, this performance rebound is still significantly lower than that seen with the single scoring model division. When the number of scoring models reaches 4, we observe that the test performance is almost the same as with three scoring models, suggesting that further increases in the number of scoring models do not lead to substantial improvement in division, and that the division effect tends to stabilise.

Combining the results of the experiments with two to four scoring models, \textbf{we found that a higher number of scoring models is not better. With two scoring models, the division results are closest to the potential OOD distribution of the major VLMs.} Continuing to increase the number of scoring models may instead introduce additional uncertainty, which leads to the division of data deviating from the potential OOD distribution of the major large VLMs and gradually converging. Based on performance and division stability considerations, we ultimately chose to use \textbf{2} scoring models to construct the default configuration of the OOD data division scheme in OODBench.

\begin{table}[h]
\caption{An ablation study of hyperparameter $T$ using the LLama-3.2-Vision~\cite{dubey2024llama}.}
\label{tab: Ablation T}
\begin{center}

\resizebox{8.5cm}{!}{
\begin{tabular}{cccccccc} 
\toprule 
\hline
\centering \multirow{2}{*}{\textbf{Threshold T}} & \centering \multirow{2}{*}{Data Type} & \multicolumn{6}{c}{nuScenes Performance} \\
\cline{3-8}
 & & Num & Accuracy & F1 & Precision & Recall & MCC \\
 \hline

 & OOD-S & 25798 & 66.30 & 63.34 & 69.43 & 58.24 & 33.02 \\
 \multirow{-2}{*}{\centering 0.05} & OOD-H & 9882 & 58.86 & 60.28 & 58.28 & 62.42 & 17.77 \\

   \rowcolor{gray!20}
 & OOD-S & 23546 & 62.65 & 55.69 & 68.44 & 46.95 & 26.64 \\
  \rowcolor{gray!20}
 \multirow{-2}{*}{\centering 0.01} & OOD-H & 5934 & 60.79 & 59.08 & 61.76 & 56.62 & 21.65 \\

 & OOD-S & 23644 & 60.93 & 53.31 & 66.24 & 44.60 & 23.13 \\
 \multirow{-2}{*}{\centering 0.005} & OOD-H & 4274 & 59.20 & 57.42 & 60.03 & 55.03 & 18.45 \\

\bottomrule
\end{tabular}}
\end{center}
\end{table}

\subsection{The selection of the threshold T}
In the process of OOD data construction, we adopt the semantic coupling criterion introduced in Sec.~\ref{Score_Deviation}, where the contrastive score $S(x,c)$ serves as a proxy for the underlying semantic mutual information $I(x;c)$. Specifically, an instance is regarded as an OOD candidate when its semantic compatibility score falls below a predefined threshold $T$, \textit{i.e.}, $S(x,c)<T$.

This criterion corresponds to selecting samples located in the semantic sparse region, where the image–text mutual information is insufficient to support coherent in-distribution semantics. To investigate the effect of the semantic coupling threshold $T$, we conduct an ablation study by varying $T$ over $\{0.05,0.01,0.005\}$. We adopt Llama-3.2-Vision as the evaluation model due to its strong multimodal reasoning capability. The results are reported in Tab.~\ref{tab: Ablation T}.

From a semantic perspective, decreasing $T$ corresponds to relaxing the minimum required semantic coupling strength, thereby allowing samples with increasingly weaker image–text alignment to be included in the OOD set. As a consequence, the distinction between different OOD regimes becomes less pronounced. Concretely, when $T=0.05$, the performance gap between OOD-S and OOD-H is 7.44, indicating a clear semantic stratification. However, as $T$ decreases to 0.01 and 0.005, the gap reduces to 1.86 and 1.73, respectively, suggesting that the boundary between moderately shifted (OOD-S) and heavily degraded (OOD-H) samples becomes increasingly blurred.

This trend is further reflected in the data distribution. As $T$ decreases, the number of OOD samples shrinks, particularly for OOD-H, which drops significantly from 9,882 to 4,274. This indicates that overly strict thresholds tend to filter out samples with extremely weak semantic signals, leading to a loss of hard OOD cases and reducing the diversity of the OOD set. Therefore, from the perspective of semantic separability and distributional coverage, a moderate threshold is preferred. In our experiments, we set $T=0.05$, which achieves a favorable balance between maintaining clear semantic stratification and preserving sufficient OOD diversity.

\section{Conclusion}
In this paper, we introduce \textbf{OODBench} to evaluate the performance of Vision Language Models in the face of out-of-distribution data. OODBench is able to more accurately reflect the coping ability of VLMs when they encounter data distribution shifts in real-world scenarios. In addition, we present the Basic-to-Advanced Progression Metric for evaluating the recognition, count-awareness, and reasoning abilities of VLMs. Our automated OOD division benchmarking process adapts efficiently to new data sources and does not require significant human effort, enabling future evaluations of the ability of vision language models to be out-of-distribution of different domains.




\bibliographystyle{IEEEtran}
\bibliography{IEEEtrans}

\begin{IEEEbiographynophoto}{Ling Lin} is currently pursuing the Ph.D. degree with the School of Artificial Intelligence and Data Science, University of Science and Technology of China (USTC). His research interests include multimodal large language models, out-of-distribution research, multimodal reasoning, and 3D scene understanding. He has published papers in several leading journals and conferences such as IEEE Transactions on Image Processing, IEEE Transactions on Neural Networks and learning systems, IEEE Transactions on Circuits and Systems for Video Technology, ICASSP, and ICME.
\end{IEEEbiographynophoto}

\begin{IEEEbiographynophoto}{Yang Bai} received the M.S. degree in computer science from Newcastle University, U.K., in 2018, and the Ph.D. degree in computer science from Durham University, U.K., in 2023. He is currently a Scientist at the Institute of High Performance Computing (IHPC), A*STAR, Singapore. His research interests include multimodal foundation models, vision--language learning, and medical artificial intelligence, with a focus on medical image understanding and clinical multimodal reasoning.
\end{IEEEbiographynophoto}

\begin{IEEEbiographynophoto}{Heng Su}
is currently pursuing a Master's degree at the School of Artificial Intelligence and Data Science, University of Science and Technology of China (USTC). He received his Bachelor's degree from China University of Petroleum (East China) (UPC). His research interests include Multimodal Large Language Models (MLLMs), visual faithfulness, and spatial hallucination.
\end{IEEEbiographynophoto}

\begin{IEEEbiographynophoto}{Yang Long\,(Senior Member, IEEE)} is an Associate Professor in the Department of Computer Science, Durham University, U.K. He is also an MRC Innovation Fellow aiming to design scalable AI solutions for large-scale healthcare applications. His research background is in the interdisciplinary field of computer vision and machine learning. He is passionate about unveiling the black-box of AI and transferring the knowledge to seek scalable, interactable, interpretable, and sustainable solutions for other disciplinary researches, e.g., physical activity, mental health, design, education, security, and geoengineering. He has authored/coauthored 30+ top-tier papers in refereed journals/conferences such as IEEE TPAMI, TIP, CVPR, AAAI, and ACM MM, and holds a patent and a Chinese National Grant.
\end{IEEEbiographynophoto}

\begin{IEEEbiographynophoto}{Congcong Zhu\,(Member, IEEE)} received the Ph.D. degree from the School of Computer Engineering and Science, Shanghai University, Shanghai, China, in 2022. He is currently an Associate Researcher at the Suzhou Institute for Advanced Research, University of Science and Technology of China, Suzhou, China. His research interests include embodied artificial intelligence, multimodal content generation, and multimodal understanding. He has published papers in several journals, including IEEE Transactions on Image Processing, IEEE Transactions on Multimedia, IEEE Transactions on Circuits and Systems for Video Technology, Pattern Recognition, and IEEE Transactions on Instrumentation and Measurement, as well as in leading conferences, including ICML, CVPR, ICCV, AAAI, and ACM MM. He was nominated for the Shanghai Computer Society’s Outstanding Doctoral Dissertation Award, received the Outstanding Graduate Student Award of Shanghai, and was recognized as an Outstanding Postdoctoral Fellow in Jiangsu Province, China.
\end{IEEEbiographynophoto}

\begin{IEEEbiographynophoto}{Yaoxing Wang} received the master's degree in computer technology from Ningxia University, Yinchuan, China, in 2022, where he is currently pursuing the Ph.D. degree in unmanned systems science and technology with the Unmanned System Research Institute, Northwestern Polytechnical University. His research interests include computer vision, 3D reconstruction, and thermography.
\end{IEEEbiographynophoto}

\begin{IEEEbiographynophoto}{Yang Zhou} is a Senior Scientist at the Institute of High Performance Computing (IHPC), A*STAR, Singapore. He earned his Ph.D. in Computer Science from Hong Kong Baptist University in 2019. From 2020 to 2022, he served as a Research Fellow at the School of Computing, National University of Singapore. His research interests span medical foundation models, vision-language modeling, probabilistic modeling, tensor analysis, natural language processing, and machine learning.
\end{IEEEbiographynophoto}

\begin{IEEEbiographynophoto}{Ling Shao\,(Fellow, IEEE)} is a Distinguished Professor with the UCAS-Terminus AI Lab, University of Chinese Academy of Sciences, Beijing, China. He was the founder of the Inception Institute of Artificial Intelligence (IIAI) and the Mohamed bin Zayed University of Artificial Intelligence (MBZUAI), Abu Dhabi, UAE. His research interests include generative AI, vision and language, and AI for healthcare. He is a fellow of the IEEE, the IAPR, the BCS and the IET.
\end{IEEEbiographynophoto}

\begin{IEEEbiographynophoto}{Huazhu Fu\,(Senior Member, IEEE)} is a Principal Scientist at the Institute of High Performance Computing (IHPC), A*STAR, Singapore. He earned his Ph.D. from Tianjin University in 2013. His research focuses on medical image analysis, AI for healthcare, and trustworthy AI. He is an Associate Editor for several distinguished journals, including IEEE Transactions on Medical Imaging, IEEE Transactions on Neural Networks and Learning Systems, and IEEE Journal of Biomedical and Health Informatics.
\end{IEEEbiographynophoto}

\begin{IEEEbiographynophoto}{Jingrun Chen} received the B.S. degree from Nanjing University, China, in 2005, and the PhD degree from the Chinese Academy of Sciences, China, in 2010. From 2010 to 2015, he served as a Visiting Assistant Professor at the University of California, Santa Barbara, and from 2015 to 2021, he was a Distinguished Professor at Soochow University. He is currently a Professor at the Suzhou Institute for Advanced Research/School of Mathematical Sciences at the University of Science and Technology of China. His main work has been published in academic journals on applied and computational mathematics such as the SIAM series, Mathematics of Computation, and Journal of Computational Physics, as well as in interdisciplinary academic journals like Nature Communications and Materials Horizons.    
\end{IEEEbiographynophoto}

\clearpage

\twocolumn[
\begin{center}
{\Huge OODBench: Out-of-Distribution Benchmark for Large Vision-Language Models\par}
{\Huge Supplementary Material\par}
\end{center}
\vspace{1em}
]


\appendices

\setcounter{section}{0}
\setcounter{subsection}{0}
\setcounter{subsubsection}{0}

\renewcommand{\thesection}{\Alph{section}}
\renewcommand{\thesubsection}{\thesection.\arabic{subsection}}
\renewcommand{\thesubsubsection}{\thesubsection.\arabic{subsubsection}}

\makeatletter
\def\@seccntformat#1{\csname the#1\endcsname\quad}
\makeatother

\renewcommand{\theequation}{A.\arabic{equation}}
\setcounter{equation}{0}

\begin{strip}
\begin{center}
\textbf{\textit{Outline}}
\end{center}

This document supplements the main paper with detailed results. Below is the outline.
\begin{itemize}
    \item [$\bullet$] Section~\ref{appendix: collection details} details the collection process of OODBench.
    \item [$\bullet$] Section~\ref{appendix: MI_CS_proof} provides the theoretical basis for using contrast scores as a proxy for mutual information.
    \item [$\bullet$] Section~\ref{appendix: ood nature evidence} demonstrates through a series of statistical analyses that OODBench exhibits distinct OOD properties relative to contemporary MLLMs.
    \item [$\bullet$] Section~\ref{appendix: hard sample} describes the differences between the OODBench and the hard samples.
    \item [$\bullet$] Section~\ref{appendix: hallucination} describes the differences between the OODBench and the hallucinations.
    \item [$\bullet$] Section~\ref{appendix: sub-dataset performance} reports the performance of classification metrics and BAP metrics for the OODBench sub-dataset.
    \item [$\bullet$] Section~\ref{appendix: model scale effect} describes the performance various of models of different scales on OODBench.
    \item [$\bullet$] Section~\ref{appendix: CLIP_BLIP2 logits visualization} provides a visualisation of the distribution of category logits for CLIP and BLIP2.
    \item [$\bullet$] Section~\ref{appendix: Case Demostration} provides partial data and case demostration of OODBench.
    \item [$\bullet$] Section~\ref{appendix: limitations} discussed the limitations and analysis of error cases of OODBench.
\end{itemize}
\end{strip}

\section{Collection Details}
\label{appendix: collection details}

\begin{figure}[ht]
    \centering
        \includegraphics[width=0.6\linewidth]{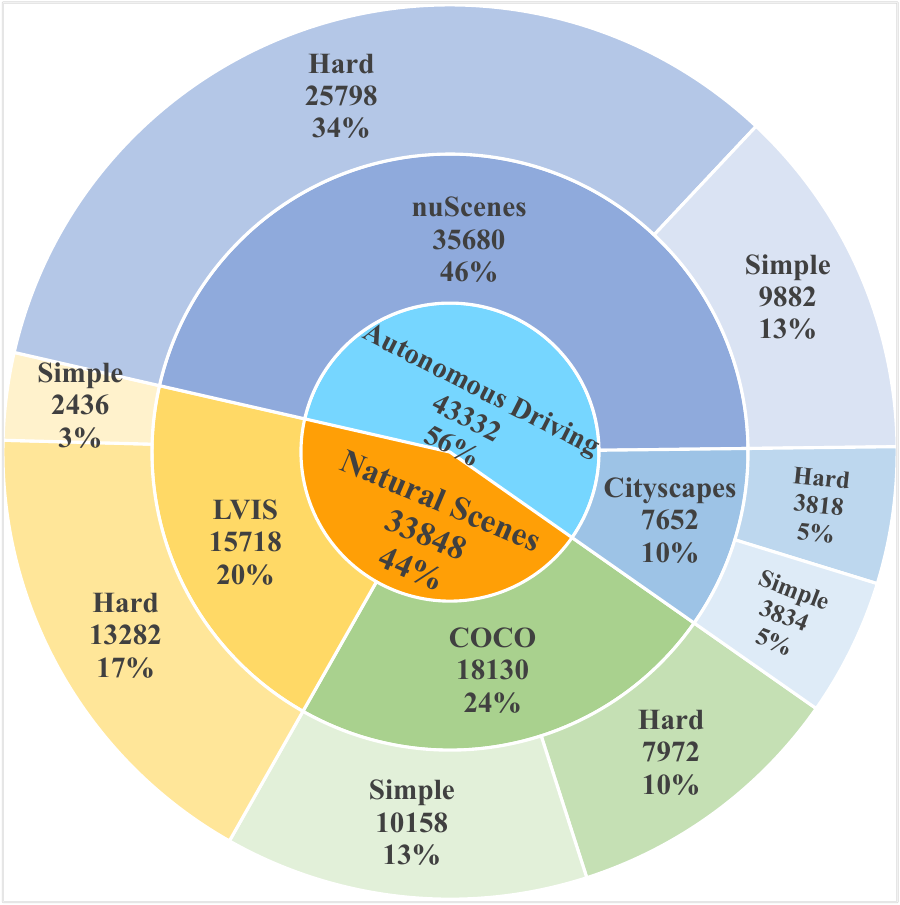}
    \caption{Distribution of categories and fields in initial benchmark.}
    \label{fig: total data pie}
\end{figure}

\textbf{Initial Benchmark.} Fig.~\ref{fig: total data pie} illustrates that we collected the initial benchmark from a publicly available dataset, which contains about 77k yes-or-no samples, where OOD-S is 51k and OOD-H is 26k. Due to the large sample size of the benchmarks, to save testing time, we randomly downsampled the OOD-Simple and OOD-Hard category data in the benchmarks with a sample size of more than 6,000 to the \textbf{OODBench} to ensure that the sample size of each category did not exceed 6,000.

\textbf{VLMs for Classification.} In terms of implementation details, we input image-question pairs corresponding to OOD samples into multimodal large models to generate textual responses. Subsequently, we convert the model output into binary labels by identifying the keywords `yes` or `no` in the generated text. This method has also been applied in the literature~\cite{zhangvisually}. In Tab.~\ref{tab:Main_Results}, an average accuracy of 90\% is achieved on ID data, which further validates the effectiveness of the method in classification tasks. In the BAP task, we also used this method to obtain the model's counting prediction results and compare them with the real labels to determine whether the model's answers to counting questions were accurate.

\textbf{nuScenes and Cityscapes dataset label modifications.} Since the labels of nuScenes do not conform to natural languages, such as `human.pedestrian.adult`, using the original labels as text directly will cause the scoring model to fail to understand their meaning, which will fail OOD data division. Therefore, as shown in Table~\ref{tab: nuscenes labels}, we manually convert these labels into a form that conforms to natural language conventions. Meanwhile, since the original label `movable object.debris` of nuScenes is not clear, we exclude it to avoid ambiguity. Similarly, we also eliminated the background, ambiguous, and semantic duplicates from the Cityscapes labels, and Table~\ref{tab: cityscapes labels} shows the final labels we retained.


\begin{table*}[ht]
\caption{Original and modified labels for nuScenes.}
\label{tab: nuscenes labels}
\begin{center}
\resizebox{14.5cm}{!}{

\begin{tabular}{ccc} 
\toprule
\textbf{Original Labels} & \textbf{Modified Labels} & \textbf{Remark} \\ 
\hline
\texttt{human.pedestrian.adult} & \texttt{Adult Pedestrian} &   \\
\rowcolor{gray!20}
\texttt{human.pedestrian.child} & \texttt{Child Pedestrian} &   \\
\texttt{human.pedestrian.wheelchair} & \texttt{Pedestrian in Wheelchair} &   \\
\rowcolor{gray!20}
\texttt{human.pedestrian.stroller} & \texttt{Pedestrian with Stroller} &   \\
\texttt{human.pedestrian.personal\_mobility} & \texttt{Pedestrian using Personal Mobility Device} &   \\
\rowcolor{gray!20}
\texttt{human.pedestrian.police\_officer} & \texttt{Police Officer} &   \\
\texttt{human.pedestrian.construction\_worker} & \texttt{Construction Worker} &   \\
\rowcolor{gray!20}
\texttt{animal} & \texttt{Animal} &   \\

\texttt{vehicle.car} & \texttt{Car} &   \\
\rowcolor{gray!20}
\texttt{vehicle.motorcycle} & \texttt{Motorcycle} &   \\

\texttt{vehicle.bicycle} & \texttt{Bicycle} &   \\
\rowcolor{gray!20}
\texttt{vehicle.bus.bendy} & \texttt{Bendy Bus} &   \\

\texttt{vehicle.bus.rigid} & \texttt{Rigid Bus} &   \\
\rowcolor{gray!20}
\texttt{vehicle.truck} & \texttt{Truck} &   \\

\texttt{vehicle.construction} & \texttt{Construction Vehicle} &   \\
\rowcolor{gray!20}
\texttt{vehicle.emergency.ambulance} & \texttt{Ambulance} &   \\

\texttt{vehicle.emergency.police} & \texttt{Police Vehicle} &   \\
\rowcolor{gray!20}
\texttt{vehicle.trailer} & \texttt{Trailer} &   \\

\texttt{movable\_object.barrier} & \texttt{Barrier} &   \\
\rowcolor{gray!20}
\texttt{movable\_object.trafficcone} & \texttt{Traffic Cone} &   \\

\texttt{movable\_object.pushable\_pullable} & \texttt{Pushable/Pullable Object} &   \\
\rowcolor{gray!20}
\texttt{movable\_object.debris} & \texttt{Debris} &  The class reference is unclear and has been deleted.  \\

\texttt{static\_object.bicycle\_rack} & \texttt{Bicycle Rack} &   \\
\bottomrule
\end{tabular}
}
\end{center}
\end{table*}

\begin{table*}[ht]
\caption{Original and modified labels for Cityscapes.}
\label{tab: cityscapes labels}
\begin{center}
\resizebox{11.5cm}{!}{
\begin{tabular}{ccc} 
\toprule
\textbf{Original Labels} & \textbf{Modified Labels} & \textbf{Remark} \\ 
\hline
\rowcolor{gray!20}
\texttt{road} & \texttt{road} &   \\
\texttt{sidewalk} & \texttt{sidewalk} & \\
\rowcolor{gray!20}
\texttt{parking} & \texttt{parking} &   \\
\texttt{rail track} & \texttt{rail track} &   \\
\rowcolor{gray!20}
\texttt{person} & \texttt{person} &   \\
\texttt{rider} & \texttt{rider} &   \\
\rowcolor{gray!20}
\texttt{car} & \texttt{car} &   \\
\texttt{truck} & \texttt{truck} &   \\
\rowcolor{gray!20}
\texttt{bus} & \texttt{bus} &   \\
\texttt{on rails} & \texttt{-} & The class reference is unclear and has been deleted.  \\
\rowcolor{gray!20}
\texttt{motorcycle} & \texttt{motorcycle} &   \\
\texttt{bicycle} & \texttt{bicycle} &   \\
\rowcolor{gray!20}
\texttt{caravan} & \texttt{caravan} &   \\
\texttt{trailer} & \texttt{trailer} &   \\
\rowcolor{gray!20}
\texttt{building} & \texttt{building} &   \\
\texttt{wall} & \texttt{wall} &   \\
\rowcolor{gray!20}
\texttt{fence} & \texttt{fence} &   \\
\texttt{guard rail} & \texttt{guard rail} &   \\
\rowcolor{gray!20}
\texttt{bridge} & \texttt{bridge} &   \\
\texttt{tunnel} & \texttt{tunnel} &   \\
\rowcolor{gray!20}
\texttt{pole} & \texttt{pole} &   \\
\texttt{pole group} & \texttt{-} & Pole group and pole are semantically duplicated, we choose pole for retention.  \\
\rowcolor{gray!20}
\texttt{traffic sign} & \texttt{traffic sign} &   \\
\texttt{traffic light} & \texttt{traffic light} &   \\
\rowcolor{gray!20}
\texttt{vegetation} & \texttt{vegetation} &   \\
\rowcolor{gray!20}
\texttt{terrain} & \texttt{-} & The class reference is unclear and has been deleted.  \\
\texttt{sky} & \texttt{-} & Background classes we exclude.  \\
\rowcolor{gray!20}
\texttt{ground} & \texttt{-} &  The class reference is unclear and has been deleted. \\
\texttt{dynamic} & \texttt{-} & The class reference is unclear and has been deleted.  \\
\rowcolor{gray!20}
\texttt{static} & \texttt{-} & The class reference is unclear and has been deleted.  \\
\bottomrule
\end{tabular}
}

\end{center}
\end{table*}


\section{Theoretical Justification of Contrastive Scores as a Mutual Information Surrogate}
\label{appendix: MI_CS_proof}
Let an image $x$ be given, and construct a candidate set $\mathcal{C} = \{c_1, \dots, c_N\}$. In this set, exactly one element is drawn from the conditional distribution $p(c \mid x)$, while the remaining $N-1$ elements are independently sampled from the marginal distribution $p(c)$. We denote the random variable $\mathcal{I} \in \{1, \dots, N\}$ as the index indicating which position contains the positive sample.

Consequently, given the image $x$ and the index $\mathcal{I}=i$, the joint density of the candidate set $\mathcal{C}$ is given by:

\begin{equation}
        p(\mathcal{C} \mid x, \mathcal{I}=i)=p(c_i \mid x) \prod_{j \neq i} p(c_j).
\end{equation}

Assuming a uniform prior where each position is equally likely to be the positive sample, \textit{i.e.}, $p(\mathcal{I}=i) = 1/N$, it follows from Bayes' rule that, given the image $x$ and the candidate set $\mathcal{C}$, the true posterior probability of position $i$ being the positive sample is:

\begin{equation}
        p(\mathcal{I}=i \mid x, \mathcal{C}) = \frac{p(\mathcal{C} \mid x, \mathcal{I}=i)p(\mathcal{I}=i)}{\sum _{k=1}^N p(\mathcal{C}\mid x, \mathcal{I}=k)p(\mathcal{I}=k)}.
\end{equation}

Since $p(\mathcal{I}=i) = 1/N$ is constant for all $i$, it cancels out, yielding:

\begin{equation}
        p(\mathcal{I}=i \mid x, \mathcal{C}) = \frac{p(c_i \mid x) \prod_{j \neq i}p(c_j)}{\sum_{k=1}^N p(c_k \mid x)\prod_{j \neq k}p(c_j)}.
\end{equation}

Next, we divide both the numerator and the denominator by $\prod_{j=1}^N p(c_j)$. Note that:

\begin{equation}
        \frac{p(c_i\mid x)\prod_{j \neq i}p(c_j)}{\prod_{j=1}^N p(c_j)} = \frac{p(c_i\mid x)}{p(c_i)}.
\end{equation}

Therefore

\begin{equation}
        p(\mathcal{I}=i\mid x, \mathcal{C}) = \frac{\frac{p(c_i \mid x)}{p(c_i)}}{\sum_{k=1}^N \frac{p(c_k \mid x)}{p(c_k)}}.
    \label{appendix:eq_true_posterior_prob}
\end{equation}

This indicates that the true Bayes-optimal classifier has a posterior probability proportional to the density ratio $\frac{p(c\mid x)}{p(c)}$. On the other hand, the contrastive learning model employs a scoring function $f(x,c)$ and defines the Softmax posterior as:

\begin{equation}
        q_\theta(\mathcal{I}=i\mid x, \mathcal{C}) = \frac{\exp (f(x, c_i)/\tau)}{\sum_{k=1}^N \exp(f(x, c_k)/\tau)}.
    \label{appendix:eq_classifier_posterior}
\end{equation}

The InfoNCE loss is precisely the cross-entropy for this $N$-way classification problem:

\begin{equation}
        \mathcal{L}_{\text{NCE}} = - \mathbb{E}[\log q_\theta (\mathcal{I}\mid x, \mathcal{C})].
\end{equation}

When the model capacity is sufficient, the optimal solution for the cross-entropy objective satisfies:

\begin{equation}
        q^*_\theta (\mathcal{I} = i\mid x, \mathcal{C}) = p(\mathcal{I}=i \mid x, \mathcal{C}).
\end{equation}

By comparing Eq.~\ref{appendix:eq_true_posterior_prob} with Eq.~\ref{appendix:eq_classifier_posterior}, it follows that in the optimal case, we must have:

\begin{equation}
        \frac{\exp (f^*(x, c_i)/\tau)}{\sum_{k=1}^N \exp (f^*(x, c_k)/\tau)} = \frac{\frac{p(c_i\mid x)}{p(c_i)}}{\sum_{k=1}^N \frac{p(c_k\mid x)}{p(c_l)}}.
\end{equation}

For the two softmax expressions to be identical for any arbitrary candidate set $\mathcal{C}$, a necessary and sufficient condition is that their unnormalized terms are proportional. That is, there exists a positive function $A(x, \mathcal{C})$, depending only on $x$ and $\mathcal{C}$, such that:

\begin{equation}
    \exp (f^*(x, c_i)/\tau) = A(x, \mathcal{C})\frac{p(c_i \mid x)}{p(c_i)}, \forall i.
\end{equation}

However, since each term on the left-hand side depends only on the pair $(x, c_i)$, it should not explicitly depend on other candidates $c_j$ within $\mathcal{C}$. For this relationship to hold consistently across any arbitrary candidate set, the scaling factor must degenerate into a term that depends on $x$ but is independent of the specific candidates $c_i$. Thus, it can be written as:

\begin{equation}
    \exp (f^*(x, c)/\tau) = A(x) \frac{p(c\mid x)}{p(c)}.
\end{equation}

This yields the following proportionality:

\begin{equation}
    \exp (f^*(x, c)/\tau) \propto \frac{p(c \mid x)}{p(c)}.
\end{equation}

Taking the logarithm of both sides, we have:

\begin{equation}
    \frac{f^*(x, c)}{\tau} = \log \frac{p(c\mid x)}{p(c)} + \log A(x).
\end{equation}

By denoting $a(x) := \log A(x)$, we obtain:

\begin{equation}
    f^*(x, c) = \tau \log \frac{p(c \mid x)}{p(c)} + \tau a(x).
\end{equation}

Furthermore, since

\begin{equation}
    \log \frac{p(c \mid x)}{p(c)} = \log \frac{p(x, c)}{p(x)p(c)} = \text{PMI}(x, c),
\end{equation}

it follows that:

\begin{equation}
    f^*(x, c) = \tau \text{PMI}(x, c) + b(x), \quad \text{where } b(x):=\tau a(x).
\end{equation}

This provides a rigorous proof that, under the InfoNCE optimal condition, the scoring function recovers a linear transformation of the PMI. Consequently,

\begin{equation}
    \exp (f(x, c)/ \tau) \propto \frac{p(c\mid x)}{p(c)}.
\end{equation}

\section{Statistical Evidence for the OOD Nature of OODBench in Modern MLLMs}
\label{appendix: ood nature evidence}

To verify whether OODBench constitutes OOD data within existing MLLMs, we conducted systematic statistical analysis across two complementary dimensions: open-source models and closed-source models. In the open-source model section, we calculated the semantic overlap rate between the model training corpus and the COCO-train-OOD partition, which was constructed according to the same rules as OODBench. We then evaluated the consistency and robustness of this partition using methods such as exact binomial intervals, Bayesian priors, and Two One-Sided Tests (TOST). In the closed-source model section, we further establish independent chains of evidence based on both the “significance” and “magnitude consistency” of performance results. This is achieved through significance testing for performance degradation (ID $\rightarrow$ OOD) and equivalence assessments against open-source OOD behavioral signatures. Through these complementary statistical inferences from both open-source and closed-source models, we jointly demonstrate that OODBench indeed exhibits verifiable OOD characteristics for current MLLMs, examining both the nature of data distributions and model response behaviors.

\subsection{OOD Evidence from Open-Source Models}
Open-Source MLLMs refer to models that disclose their core components to varying degrees within the research community, thereby enabling independent auditability, inspectability, reproducibility, or reusability. It is crucial to emphasize that open-source is not a binary attribute but exists along a continuum of openness. As shown in Table~\ref{tab:openness_levels}, we categorize current open-source MLLMs into four primary levels based on the scope and granularity of disclosed information, where each level exhibits a hierarchical increase in model weights, architectural details, training code, training data transparency, and reproducibility.

To rigorously assess whether the proposed OODBench constitutes out-of-distribution (OOD) data for existing open-source MLLMs, we conducted a systematic statistical analysis. We first quantified the overlap ratio between OODBench and each model’s corresponding training corpus. Then we incorporated statistical inference procedures to estimate the population-level proportion of OOD samples across models. To enhance the representativeness and cross-model comparability of these evaluations, we adopted the COCO dataset—one of the most commonly used and widely standardized datasets in multimodal model training—as the unified reference source. This provides a consistent framework for estimating overlap ratios and determining the OOD characteristics of our constructed benchmark.

\subsubsection{Principles for Selecting Model Samples}
We first conducted a systematic survey of current mainstream open-source MLLMs, covering diverse model families, parameter scales, and architectural paradigms. By comprehensively collecting publicly released versions, model repositories, and official documentation, we identified 28 model families corresponding to a candidate set of 93 models spanning parameter sizes from 0.5B to 424B. To ensure transparency and auditability at the training data level, we restricted our selection to robust open-source models at Levels 3 and 4, meaning those that either publicly disclose their training data inventories or possess strict reproducibility. In this process, we excluded the following two categories of models that did not meet these criteria:
\begin{itemize}
    \item Models using in-house or non-public data.
    \item Models with incomplete training data lists often lack precise entries, which prevents line-by-line verification.
\end{itemize}

\begin{table*}[ht]
\centering
\caption{Current open-source MLLM models are categorized into four primary tiers based on the scope and granularity of publicly disclosed information. Each level exhibits progressively enhanced transparency and reproducibility across model weights, architectural details, training code, training data, and overall reproducibility.}
\label{tab:openness_levels}
\resizebox{17.5cm}{!}{%
\begin{tabular}{
     >{\centering\arraybackslash}m{1cm}
    |>{\centering\arraybackslash}m{5cm}
    |>{\centering\arraybackslash}m{2cm}
    |>{\centering\arraybackslash}m{2cm}
    |>{\centering\arraybackslash}m{2.5cm}
    |>{\centering\arraybackslash}m{2.5cm}
    |>{\centering\arraybackslash}m{2.5cm}
}
\toprule
Level & Name & Weights & Architecture & Training Code & Training Data & Reproducibility \\
\midrule
0  & Non-Open Source 
   & \textcolor{red}{\ding{55}} & \textcolor{red}{\ding{55}} & \textcolor{red}{\ding{55}} & \textcolor{red}{\ding{55}} & \textcolor{red}{\ding{55}} \\
\midrule
1A & Architecture Open Source 
   & \textcolor{red}{\ding{55}} & \textcolor{green}{\ding{51}} & Partial & \textcolor{red}{\ding{55}} & \textcolor{red}{\ding{55}}\\
\midrule
1B & Weights Open but Architecture Not Fully Open
   & \textcolor{green}{\ding{51}} & Partial & Partial & \textcolor{red}{\ding{55}} & \textcolor{red}{\ding{55}} \\
\midrule
1C & Data Partially Open
   & Optional & Optional & Optional & Partial & \textcolor{red}{\ding{55}} \\
\midrule
2  & Weights Open 
   & \textcolor{green}{\ding{51}} & \textcolor{green}{\ding{51}} & \textcolor{green}{\ding{51}} & \textcolor{red}{\ding{55}} / Rough & \textcolor{green}{\ding{51}} (Partial) \\
\midrule
3  & Strongly Open (Reproducible)
   & \textcolor{green}{\ding{51}} & \textcolor{green}{\ding{51}} & \textcolor{green}{\ding{51}} & \textcolor{green}{\ding{51}} (Clean) & \textcolor{green}{\ding{51}} (Partial) \\
\midrule
4  & Fully Open Source (Full Pipeline Open)
   & \textcolor{green}{\ding{51}} & \textcolor{green}{\ding{51}} & \textcolor{green}{\ding{51}} & \textcolor{green}{\ding{51}} (Fully Open) & \textcolor{green}{\ding{51}}\textcolor{green}{\ding{51}}\textcolor{green}{\ding{51}} (Strict) \\
\bottomrule
\end{tabular}
}
\end{table*}

\begin{table*}[!htbp]
  \centering
  \caption{Complete summary of collected models. Includes models ultimately included in the analysis as well as those excluded during the screening process.}
  \label{tab:open_models_overview}
  \resizebox{17.5cm}{!}{%
  \begin{tabular}{
      >{\centering\arraybackslash}m{0.5cm}|
      >{\centering\arraybackslash}m{2.5cm}|
      >{\centering\arraybackslash}m{0.8cm}|
      >{\centering\arraybackslash}m{12cm}|
      >{\centering\arraybackslash}m{1.8cm}|
      >{\centering\arraybackslash}m{1.2cm}|
      >{\centering\arraybackslash}m{1.2cm}|
      >{\centering\arraybackslash}m{3cm}
    }
    \toprule
    No. & Family & Years & Models & Institution & COCO Used & Openness Level & Included in Analysis\\
    \midrule
    1 & Flamingo~\cite{alayrac2022flamingo} & 2022 & Flamingo-3B, Flamingo-9B & DeepMind  & \textcolor{red}{\ding{55}} & L3 & \textcolor{red}{\ding{55}} \\
    \midrule
    2 & Qwen-VL~\cite{bai2023qwenvlversatilevisionlanguagemodel} & 2023 & Qwen-VL(7B), Qwen-VL-Chat(7B), Qwen-VL-Plus(7B), Qwen-VL-Max(7B) & Alibaba Group & \textcolor{green}{\ding{51}}  & L3 & \textcolor{green}{\ding{51}}  \\
    \midrule   
    3 & PaLI-X~\cite{chen2023pali} & 2023 & PaLI-X(55B) & Google Research & \textcolor{red}{\ding{55}}  & L1C & \textcolor{red}{\ding{55}}  \\
    \midrule   
    4 & LLaVA-v1~\cite{liu2023visual} & 2023 & LLaVA-v1-Vicuna(13B), LLaVA-v1-LLama(13B), LLaVA-v1-LLama(7B) & UW-Madison & \textcolor{green}{\ding{51}}  & L3 & \textcolor{green}{\ding{51}}  \\
    \midrule   
    5 & InstructBLIP~\cite{dai2023instructblip} & 2023 & InstructBLIP (FlanT5XL)(3B), InstructBLIP (FlanT5XXL)(11B), InstructBLIP (Vicuna-7B), InstructBLIP (Vicuna-13B) & Salesforce Research & \textcolor{green}{\ding{51}}  & L3 & \textcolor{green}{\ding{51}}  \\
    \midrule   
    6 & MiniGPT-4~\cite{zhu2023minigpt-4} & 2023 & MiniGPT-4 (Vicuna 13B), MiniGPT-4 (Vicuna 7B), MiniGPT-4 (LLaMA-2 Chat 7B),  & KAUST & \textcolor{red}{\ding{55}}  & L3 & \textcolor{red}{\ding{55}}  \\
    \midrule   
    7 & MiniGPT-v2~\cite{chen2023minigpt-v2} & 2023 & MiniGPT-v2-LLaMA2-7B, MiniGPT-v2-Vicuna-7B, MiniGPT-v2-Vicuna-13B & KAUST & \textcolor{green}{\ding{51}}  & L3 & \textcolor{green}{\ding{51}}  \\
    \midrule   
    8 & OpenFlamingo~\cite{awadalla2023openflamingo} & 2023 & OpenFlamingo-3B, OpenFlamingo-3B (Instruct), OpenFlamingo-4B, OpenFlamingo-4B (Instruct), OpenFlamingo-9B  & UW & \textcolor{red}{\ding{55}}  & L1C & \textcolor{red}{\ding{55}}  \\
    \midrule   
    9 & InternVL~\cite{chen2024internvl} & 2024 & InternVL-C(7B), InternVL-G(7B), InternVL-Chat-13B, InternVL-Chat-19B & OpenGVLab & \textcolor{green}{\ding{51}}  & L3 & \textcolor{green}{\ding{51}}  \\
    \midrule   
    10 & DeepSeek-VL~\cite{lu2024deepseekvl} & 2024 & DeepSeek-VL-1.3B-base, DeepSeek-VL-1.3B-chat, DeepSeek-VL-7B-base, DeepSeek-VL-7B-chat & DeepSeek-AI & \textcolor{red}{\ding{55}}  & L3 & \textcolor{red}{\ding{55}}  \\
    \midrule   
    11 & LLaVA-1.5~\cite{liu2024improved} & 2024 & LLaVA-1.5-7B, LLaVA-1.5-13B & UW-Madison & \textcolor{green}{\ding{51}}  & L3 & \textcolor{green}{\ding{51}}  \\
    \midrule   
    12 & InternVL~\cite{chen2024internvl1.5} & 2024 & InternVL-1.2(40B), InternVL-1.5(26B) & Shanghai AI Laboratory & \textcolor{green}{\ding{51}}  & L3 & \textcolor{green}{\ding{51}}  \\
    \midrule   
    13 & CogVLM~\cite{wang2024cogvlm} & 2024 & cogvlm-chat-v1.1-Vicuna-7B, cogvlm-chat-v1.1-Llama3-8B, cogvlm-base-224(7B), cogvlm-base-490(7B), cogvlm-grounding-generalist(7B) & Zhipu AI & \textcolor{green}{\ding{51}}  & L3 & \textcolor{green}{\ding{51}}  \\
    \midrule   
    14 & MiniCPM-V~\cite{yao2024minicpm} & 2024 & MiniCPM-V 1.0(2.8B), MiniCPM-V 2.0(2.8B), MiniCPM-Llama3-V 2.5(8.5B)& OpenBMB & \textcolor{green}{\ding{51}}  & L3 & \textcolor{green}{\ding{51}}  \\
    \midrule   
    15 & Emu3~\cite{wang2024emu3} & 2024 & Emu3-Chat(8B), Emu3-Gen(8B), Emu3-DPO(8B) & BAAI & \textcolor{red}{\ding{55}}  & L1C & \textcolor{red}{\ding{55}}  \\
    \midrule   
    16 & Monkey~\cite{li2024monkey} & 2024 & Monkey(9.8B) & HUST & \textcolor{green}{\ding{51}}  & L3 & \textcolor{green}{\ding{51}}  \\
    \midrule   
    17 & Qwen2-VL~\cite{wang2024qwen2-vl} & 2024 & Qwen2-VL-2B, Qwen2-VL-7B, Qwen2-VL-72B & Alibaba Group & unknown  & L1C & \textcolor{red}{\ding{55}}  \\
    \midrule
    18 & CogVLM~\cite{wang2024cogvlm} & 2024 & CogVLM2-LLaMA3(8B), GLM-4V-9B & Zhipu AI & \textcolor{red}{\ding{55}}  & L1C & \textcolor{red}{\ding{55}}  \\
    \midrule
    19 & DeepSeek-VL2~\cite{wu2024deepseek-vl2} & 2024 & DeepSeek-VL2-tiny(3B), DeepSeek-VL2-small(16B), DeepSeek-VL2(27B) & DeepSeek-AI & unknown  & L1C & \textcolor{red}{\ding{55}}  \\
    \midrule
    20 & LLaVA-NeXT~\cite{li2024llavanext-strong} & 2024 & LLaVA-NeXT-Interleave-0.5B, LLaVA-NeXT-Interleave-7B, LLaVA-NeXT-Interleave-14B, LLaVA-NeXT-Interleave-7B-DPO & ByteDance & \textcolor{green}{\ding{51}}  & L1C & \textcolor{red}{\ding{55}}  \\
    \midrule
    21 & Open-Qwen2-VL~\cite{Open-Qwen2VL} & 2025 & Open-Qwen2-VL(2B) & UC Santa Barbara & \textcolor{green}{\ding{51}}  & L3 & \textcolor{green}{\ding{51}}  \\
    \midrule
    22 & BLIP-3~\cite{xue2025blip} & 2025 & BLIP-3-4B-SI, BLIP-3-4B-MI, BLIP-3-14B-SI, BLIP-3-14B-MI & Salesforce AI Research & \textcolor{green}{\ding{51}}  & L3 & \textcolor{green}{\ding{51}}  \\
    \midrule
    23 & BAGEL~\cite{deng2025bagel} & 2025 & BAGEL-1.5B, BAGEL-7B & ByteDance Seed & \textcolor{green}{\ding{51}}  & L3 & \textcolor{red}{\ding{55}}  \\
    \midrule
    24 & InternVL2~\cite{chen2024internvl2.5} & 2025 & InternVL2-2B, InternVL2-4B, InternVL2-8B, InternVL2-26B, InternVL2-40B, InternVL2-Llama3-76B & Shanghai AI Laboratory & \textcolor{green}{\ding{51}}  & L3 & \textcolor{green}{\ding{51}}  \\
    \midrule
    25 & InternVL2.5~\cite{chen2024internvl2.5} & 2025 & InternVL2.5-1B, InternVL2.5-2B, InternVL2.5-4B, InternVL2.5-8B, InternVL2.5-26B, InternVL2.5-38B, InternVL2.5-78B & Shanghai AI Laboratory & \textcolor{green}{\ding{51}}  & L3 & \textcolor{green}{\ding{51}}  \\
    \midrule
    26 & Gemma-3~\cite{team2025gemma} & 2025 & Gemma-3-1B, Gemma-3-4B, Gemma-3-12B, Gemma-3-27B & Google DeepMind & unknown  & L1C & \textcolor{red}{\ding{55}}  \\
    \midrule
    27 & ERNIE 4.5~\cite{ernie2025tERNIE4.5} & 2025 & ERNIE-4.5-VL-28B-A3B-Thinking, ERNIE-4.5-VL-424B-A47B-Base, ERNIE-4.5-VL-424B-A47B, ERNIE-4.5-VL-28B-A3B-Base, ERNIE-4.5-VL-28B-A3B & Baidu & unknown  & L1C & \textcolor{red}{\ding{55}}  \\
    \midrule
    28 & Qwen2.5-VL~\cite{bai2025qwen2.5-vl} & 2025 & Qwen2.5-VL-3B, Qwen2.5-VL-7B, Qwen2.5-VL-72B & Alibaba Group & unknown  & L1C & \textcolor{red}{\ding{55}}  \\

    \bottomrule
  \end{tabular}
  }
\end{table*}

After obtaining a set of candidate models that meet the training data transparency requirements, we further exclude models that did not use COCO as a training data source. Since our objective is to quantify the overlap ratio between OODBench-COCO and each MLLM's training data, a model with no COCO inclusion in its training corpus would yield a trivially zero overlap ratio. This would render it incapable of providing any meaningful validation for our proposed overlap ratio analysis workflow. Consequently, such models are unsuitable as statistical test samples. After these two rounds of screening, we ultimately obtained 50 qualifying MLLMs for subsequent overlap ratio calculations and statistical inference analysis. Table~\ref{tab:open_models_overview} provides a comprehensive summary of the collected models, including both those ultimately included in the analysis and those excluded during the screening process.

\subsubsection{Semantic-Level OOD Pairing and Overlap Rate Analysis}
We first select an image and perform semantic consistency matching for its image-category within the OOD division. Specifically, we collect all annotated versions of this image across various MLLM training corpora (\textit{e.g.}, captions, OCR, conversations, detailed descriptions, complex reasoning, REC, REG, grounding, \textit{etc.}). We then systematically examine each annotation to determine whether it contains semantic expressions matching the OOD category we assigned. If the category appears in any annotated version, it indicates the model has learned the corresponding image–category semantic relationship, and this image–category pair is considered ID data. Conversely, if the category is absent from all annotated versions, the image–category pair is classified as OOD data at the semantic level of learning. It is important to note that OODBench utilizes the COCO val dataset, while most MLLM training data is based on the COCO train version. These datasets have no overlap at the image level, making it impossible to calculate overlap rates using image matching rules directly. This version discrepancy may result in false-zero overlap. To address this issue, we regenerate OOD data using the COCO train image set, applying the same division criteria as OODBench. We then calculate the overlap rate with each MLLM's training data based on this division. This approach verifies whether our categorized samples genuinely constitute an OOD distribution at the training semantic level.

\subsubsection{Population-Level OOD Assessment}
\label{subsubsection: population_level ood assessment}
\textbf{Settings and Symbols.} We selected $N$ auditable open-source MLLMs as test subjects. Our constructed dataset is based on COCO-train and generates OOD data at the semantic level according to OODBench's division strategy, denoted as "COCO-train-OOD".

For each model $\textbf{m}$, its training overlap rate on "COCO-train-OOD" is defined as:
\begin{equation}
    \begin{split}
        r_m = \frac{\mathcal{H}_m}{|C|},
    \end{split}
\end{equation}

\noindent where $\mathcal{H}_m$ denotes the number of samples in the model training set that achieve semantic overlap with the "COCO-train-OOD" image-category pairs, while $|C|$ represents the total number of samples in "COCO-train-OOD". Setting the threshold $\varepsilon$ as the maximum acceptable overlap rate, the model's OOD decision criterion (pre-registered criterion) is defined as:

\begin{equation}
    \begin{split}
        Z_m = \mathds{1}[\mathrm{UCI}(r_m)<\varepsilon],
    \end{split}
\end{equation}

when the 95\% upper confidence bound of the training overlap rate for model $\textbf{m}$ does not exceed $\varepsilon$, we classify it as ‘treating this data as OOD’ ($Z_m=1$); if the upper confidence bound exceeds $\varepsilon$, it is classified as non-OOD ($Z_m=0$).

\textbf{Exact Clopper–Pearson Interval (Frequentist Lower Bound).} To estimate, at the population level, the proportion of models that satisfy the preregistered criterion, we treat the set of $N$ evaluated open-source models as a finite population. Among them, $k$ models satisfy $Z_m = 1$. We therefore compute the lower confidence bound for the binomial proportion based on $k$ successes out of $n=N$ trials. \newline
The lower Clopper–Pearson bound is defined as:

\begin{equation}
    \begin{split}
        \mathrm{LCP}=\mathrm{BetaInv}(0.025; k, n - k + 1).
    \end{split}
\end{equation}

The calculation results are as follows:
\begin{itemize}
    \item When $\varepsilon = 5\%$, $k=49$, $n=50$, \quad $\mathrm{LCP}=\mathrm{BetaInv}(0.025; 49,2) \approx 0.894.$
	\item When $\varepsilon=2\%$, $k=27$, $n=50$, \quad $\mathrm{LCP}=\mathrm{BetaInv}(0.025; 27,24) \approx 0.393.$
\end{itemize}

Therefore, under the criterion of "$95\%$ upper confidence bound $\le 5\%$ for training overlap rate," it can be asserted with $95\%$ confidence that at least approximately $89.4\%$ of comparable open-source MLLMs would classify this data as OOD. Even under the stricter criterion of "$\le 2\%$," at least approximately $39.3\%$ of models still meet this condition.

\textbf{Bayesian Inference with a Beta(1,1) Prior (Posterior Credible Interval).} To validate the above results from a probabilistic reasoning perspective, we employ a uniform prior "Beta(1,1)" to compute the lower bound of the $95\%$ confidence interval under the posterior distribution:

\begin{equation}
    \begin{split}
        L_{\mathrm{Bayes}} =\mathrm{BetaInv}(0.025; k+1, n-k+1).
    \end{split}
\end{equation}

The result is:
\begin{itemize}
    \item When $\varepsilon = 5\%$, $k=49$, $n=50$, \quad $L_{\mathrm{Bayes}} =\mathrm{BetaInv}(0.025; 50,2) \approx 0.896.$
	\item When $\varepsilon=2\%$, $k=27$, $n=50$, \quad $L_{\mathrm{Bayes}} =\mathrm{BetaInv}(0.025; 28, 24) \approx 0.403.$
\end{itemize}

Therefore, from a Bayesian perspective, it can be asserted with $95\%$ confidence that: when the criterion is "$95\%$ upper confidence bound $\le 5\%$," at least approximately $89.6\%$ of similar models will classify the data as OOD; when the criterion is "$\le 2\%$," at least approximately $40.3\%$ of models will classify the data as OOD.

\textbf{Population-Level Conclusion.} From two complementary statistical perspectives, the Clopper–Pearson method ensures frequentist coverage, while the Bayesian Beta method provides posterior credibility. Both methods align in direction and magnitude, indicating that, \textbf{with $95\%$ confidence, at least approximately $89\text{–}90\%$ of comparable open-source MLLMs perceive COCO-train-OOD data as semantically OOD during training.} These open-source MLLMs encompass diverse families, parameter scales, architectures, and training strategies. Thus, this division can be robustly regarded as an "out-of-distribution relative to training semantics" sample set at the population level.

\subsubsection{Distributional Equivalence Between COCO-Train-OOD and COCO-Val-OOD}
Let the random variable be:
\begin{equation}
    \begin{split}
       (X, C) \in \mathcal{X} \times \mathcal{C},
    \end{split}
\end{equation}

where $X$ denotes images and $C$ denotes discrete categorical labels. We consider separately:

\begin{itemize}
    \item $P_{\text{train}}(X, C)$: The joint distribution of OOD subsets divided according to OODBench rules on COCO-train.
    \item $P_{\text{val}}(X, C)$: The joint distribution of OOD subsets divided according to the same rules on COCO-val.
\end{itemize}

Section~\ref{subsubsection: population_level ood assessment}, based on the exact Clopper–Pearson interval and the Bayesian Beta(1,1) posterior interval, infers at the $95\%$ confidence level that at least approximately $89\text{–}90\%$ of open-source MLLMs will treat COCO-train-OOD data as OOD for training semantics. To extend this population-level OOD classification to COCO-val-OOD, it is necessary to verify further whether COCO-train-OOD and COCO-val-OOD belong to the same statistical distribution family.

\textbf{Definition of Distribution Equivalence.} We adopt the kernel mean embedding framework and utilize Maximum Mean Discrepancy (MMD) to measure the distance between two joint distributions.

For any image $x$, features are extracted using the self-supervised visual model DINOv2~\cite{oquab2024dinov2}.

\begin{equation}
    \begin{split}
        \phi(X)=\mathbb{R}^d.
    \end{split}
\end{equation}

The reason for selecting DINOv2 is that its training process does not rely on COCO labels, making its feature space "semantically neutral" and thus avoiding the introduction of unnecessary supervised bias. Category labels $c \in C$ are represented using one-hot vectors $e_c \in \mathbb{R}^t$.

We select a kernel function on the joint space $\mathcal{Z} = \mathcal{X} \times \mathcal{C}$:

\begin{equation}
    \begin{split}
        k((x,c), (x', c')).
    \end{split}
\end{equation}

The nature of the kernel guarantees the existence of a Hilbert space (RKHS) $\mathcal{H}$ and a mapping:

\begin{equation}
    \begin{split}
        \exists \ \mathcal{H}, \Phi: \mathcal{Z} \rightarrow \mathcal{H} \quad s.t. \quad k(z, z')=\langle \Phi (z), \Phi(z')\rangle _\mathcal{H}.
    \end{split}
\end{equation}

where $z=(x, c), z'=(x', c')$.

For any distribution $P$ over $\mathcal{Z}$, we can define its mean embedding in $\mathcal{H}$:

\begin{equation}
    \begin{split}
        \mu _P:= \mathbb{E}_{Z\sim P} [\Phi(\mathcal{Z})]\in \mathcal{H}.
    \end{split}
\end{equation}

We hereby order:

\begin{equation}
    \begin{split}
        k((x, c),(x',c'))=k_v(\phi(x), \phi(x'))\cdot \mathds{1}[c=c'],
    \end{split}
\end{equation}

where $\phi(x) \in \mathbb{R}^d$ represents image features extracted via self-supervised visual models \textit{(e.g.}, DINOv2), $k_v$ denotes the RBF kernel applied to visual features, and $\mathds{1}[\cdot]$ is the indicator function. This design constructs a block-diagonal joint kernel in the joint space $(X, C)$, where kernel values between different categories are zero, making them mutually orthogonal in the RKHS. Meanwhile, similarity within the same category is entirely determined by the RBF kernel of the visual features. In other words, the label variable $C$ participates in joint modeling through the discrete structure of "intra-class/cross-class," while the visual subspace provides the intra-class geometric structure.

Based on the above joint kernel, the maximum mean displacement (MMD) between the joint distributions $P_{\text{train}}$ and $P_{\text{val}}$ can be expressed as:

\begin{equation}
    \begin{split}
        \mathrm{MMD}^2 (P_{\text{train}}, P_{\text{val}}) = \|\mu_{\text{train}} - \mu_{\text{val}}\|_\mathcal{H}^2,
    \end{split}
\end{equation}

where $\mu_{\text{train}}, \mu_{\text{val}} \in \mathcal{H}$ represent the kernel mean embeddings of $P_{\text{train}}(X, C)$ and $P_{\text{val}}(X, C)$ in the reproducing kernel Hilbert space $\mathcal{H}$, respectively. That is, $\mu_P = \mathbb{E}_{(X, C)\sim P} [\Phi(X, C)]$, where $\Phi$ is the feature map induced by the joint kernel k.

\textbf{Statistical Test for Distribution Equivalence (Two-Sided Test).} We used the equivalence test (TOST) to verify the following hypothesis:

\begin{itemize}
    \item $\text{H}_0$: $\mathrm{MDD}^2(P_{\text{train}}, P_{\text{val}})\geq \xi.$ (The difference is at least $\xi$ and cannot be considered equivalent)
    \item $\text{H}_1$: $\mathrm{MDD}^2(P_{\text{train}}, P_{\text{val}}) < \xi.$ (The difference is less than $\xi$ and can be considered equivalent)
\end{itemize}

To determine the value of $\xi$, we construct a 'homogeneous baseline' within COCO-train-OOD as follows:

\begin{itemize}
    \item Randomly partition COCO-train-OOD into two mutually exclusive subsets: \newline
    \[ (D_{\text{train}}^{(1)}, D_{\text{train}}^{(2)}). \]
    \item Calculate the $\mathrm{MMD}^2$ between them: \newline
    \[ \mathrm{MMD}^2(D_{\text{train}}^{(1)}, D_{\text{train}}^{(2)}). \]
    \item Repeat multiple times to obtain a set of $\mathrm{MMD}^2$ values within the same distribution.
    \item Take the $95th$ percentile of this distribution as $\xi$: \newline
    \[
        \xi=\mathrm{Quantile}_{0.95}\{ \mathrm{MMD}^2(D_{\text{train}}^{(1)}, D_{\text{train}}^{(2)})\}.
    \]
\end{itemize}

$\xi$ denotes the upper bound of the finite-sample fluctuations of $\mathrm{MMD}^2$ observed between two independent samples originating from the same underlying distribution. Consequently, if the $\mathrm{MMD}^2$ between COCO-train-OOD and COCO-val-OOD falls below $\xi$, the two partitions can be statistically regarded as independent realizations of a common source distribution, \textit{i.e.}, as distributionally isomorphic.

To further quantify the uncertainty in $\mathrm{MMD}^2$ estimation, we employ a permutation test to construct a reference distribution for its null hypothesis. Specifically, we merge all samples from COCO-train-OOD and COCO-val-OOD into a single dataset, randomly permute their group labels (train/val), and recalculate $\mathrm{MMD}^2$ based on the permuted grouping structure. Repeating this process B times (B=500 in this experiment) yields the empirical permutation distribution of $\mathrm{MMD}^2$ under the null hypothesis $\text{H}_0$ (equivalence of the two distributions):

\begin{equation}
    \begin{split}
        \mathcal{D}_\mathrm{perm}=\{\mathrm{MMD}_{\mathrm{perm}}^2 (b)\}_{(b=1)}^B.
    \end{split}
\end{equation}

Thus, we construct an upper bound for the $\mathrm{MMD}^2$ by taking the $1-\alpha$-quantile of this distribution:

\begin{equation}
    \begin{split}
        \mathrm{UCI}_{(1-\alpha)}=\mathrm{Quantile}_{(1-\alpha)} (\mathcal{D}_\mathrm{perm}).
    \end{split}
\end{equation}

Under the Two One-Sided Tests (TOST) procedure, if the following condition is satisfied:

\begin{equation}
    \begin{split}
        \mathrm{UCI}_{0.95} (\mathrm{MMD}^2 (P_{\text{train}}, P_{\text{val}}))<\xi,
    \end{split}
\end{equation}

then it can be asserted:

\begin{equation}
    \begin{split}
        P_{\text{train}}^{\text{OOD}} \approx P_{\text{val}}^{\text{OOD}},
    \end{split}
\end{equation}

then, at significance level $\alpha$, the null hypothesis $\text{H}_0$ that the difference is at least $\xi$ can be rejected, and the equivalence hypothesis $\text{H}_1$ can be accepted, indicating that the two joint distributions are statistically equivalent within the prescribed tolerance.

\textbf{Observation Conclusion:} Based on the homogenous baseline constructed from COCO-train-OOD internal partition, the upper bound of $\mathrm{MMD}^2$'s $95\%$ co-distribution fluctuation is determined to be:
\[
    \xi = 4.47 \times 10^{-4}.
\]

After estimating the joint distribution of COCO-train-OOD and COCO-val-OOD, we further obtained the $95\%$ confidence interval for $\mathrm{MMD}^2$ via bootstrap sampling:

\[
    \mathrm{CI}_{95} (\mathrm{MMD}^2)=[1.84 \times 10^{-4}, 4.21 \times 10^{-4}],
\]

Its upper confidence bound is significantly lower than the homogeneity threshold $\xi$.

Simultaneously, the TOST equivalence test constructed based on the empirical null distribution from B=500 permutations also indicates that the $95\%$ upper confidence bound of $\mathrm{MMD}^2$ satisfies:

\[
    \mathrm{UCI}_{0.95} (\mathrm{MMD}^2 (P_{\text{train}}, P_{\text{val}}))< \xi,
\]

At the significance level $\alpha = 0.05$, we reject the null hypothesis $\text{H}_0$ —that the difference is at least $\xi$ —and accept the equivalence hypothesis $\text{H}_1$. This confirms that COCO-train-OOD and COCO-val-OOD are statistically distributionally equivalent within the prescribed tolerance. Since a model’s OOD response is fundamentally determined by the degree to which the test data deviates from its training semantic distribution, and train-OOD has already been shown in the previous section to constitute semantic OOD for open-source MLLMs, the statistical indistinguishability between val-OOD and train-OOD indicates that both deviate from the training distribution to the same order of magnitude. Consequently, COCO-val-OOD should exhibit the same OOD behavior for open-source MLLMs and can thus be regarded as semantic OOD data equivalent to train-OOD.

\subsection{OOD Evidence from Closed-Source Models}
\textbf{Problem Setup and Assumptions.} Let
\begin{itemize}
    \item $D_{\text{ID}}$: Conventional ID test set.
    \item $D_{\text{OOD}}$: A semantic OOD test set constructed according to OODBench semantic rules (non-main semantic objects, semantic variants, and anomalous image–category pairings), with its image subset sourced from COCO-val.
\end{itemize}

In the open-source model section, we have established the following two key premises:

\begin{itemize}
    \item [1.] \textbf{Training Out-of-Distribution Properties.}\newline
    By calculating the semantic overlap rate between the training corpora of open-source MLLMs and COCO-train-OOD, and combining precise binomial intervals with Bayesian credible intervals using uniform priors, we obtain the following: At the $95\%$ confidence level, at least $90\%$ of open-source models exhibit an overlap rate with COCO-train-OOD not exceeding $5\%$. This result indicates that, \textbf{from the perspective of training distribution, COCO-train-OOD can be robustly regarded as OOD data for the collection of open-source models.}
    \item [2.] \textbf{Distributional Equivalence Between COCO-train-OOD and COCO-val-OOD.}\newline
    Under the same semantic division rules, we construct COCO-train-OOD and COCO-val-OOD and perform an equivalence test in the joint space of image features and category labels using kernel MMD combined with the TOST procedure. The results indicate that \textbf{the two subsets are statistically indistinguishable and can be regarded as two independent samples drawn from the same underlying distribution, \textit{i.e.}, they exhibit distributional equivalence.}
\end{itemize}

Therefore, in subsequent reasoning, it can be directly treated as a known premise:

\begin{itemize}
    \item \textbf{For the open-source model collection $\mathcal{M}_{\text{open}}$, $D_{\text{OOD}}$ is OOD in terms of the training distribution.}
\end{itemize}

For closed-source models (such as GPT-4o, Gemini, \textit{etc.}), since their training data is inaccessible, we cannot directly determine whether $D_{\text{OOD}}$ falls outside their training distribution. To address this, we employ behavioral consistency inference:

\begin{itemize}
    \item First, inductively derive observable "open-source model OOD behavioral patterns"— \textit{i.e.}, OOD behavioral signatures—from the inference results of open-source models on $D_{\text{OOD}}$.
    \item Subsequently, verify whether the behavior of closed-source models on the same data $D_{\text{OOD}}$ aligns with these known "open-source OOD behavioral patterns".
\end{itemize}

Suppose its behavior aligns with the OOD behavioral signature of the open-source model. In that case, it can be inferred that the closed-source model's response pattern on $D_{\text{OOD}}$ corresponds to its performance on typical OOD data. This indicates that it exhibits behavioral characteristics consistent with OOD scenarios on that dataset.

\subsubsection{Behavioral Statistics and Performance Degradation}
For any model $m$ (either open-source or closed-source) and for any data split $S \in \{ID, OOD\}$, we define the basic classification metrics as follows:

\begin{equation}
    \begin{split}
        T_{(m,S)}^{(k)} \in \{\text{ACC, F1, Precision, Recall, MCC}\},\quad k=1,\cdots, K,
    \end{split}
\end{equation}

For example, $T_{(m,S)}^{(1)} = \text{ACC}_{m, S}$ and$ T_{(m,S)}^{(2)} = \text{F1}_{m, S}$ correspond to five performance statistics respectively.

For each metric dimension $k$, we define the ID $\rightarrow$ OOD performance degradation measure:

\begin{equation}
    \begin{split}
        \Delta_m^{(k)} :=T_{m, ID}^{(k)} - T_{m, OOD}^{(k)}.
    \end{split}
\end{equation}

Intuitively, $\Delta_m^{(k)} > 0$ indicates that the model's performance degrades on semantically OOD data; the larger the degradation, the more pronounced the OOD behavior in this dimension. In practice, we perform bootstrap resampling on the model's outputs across each data division to obtain the $95\%$ confidence interval for $\Delta_m^{(k)}$:

\begin{equation}
    \begin{split}
        \mathrm{CI}_{95}(\Delta_m^{(k)})=[L_m^{(k)}, U_m^{(k)}].
    \end{split}
\end{equation}

As demonstrated by the aforementioned distribution homogeneity analysis, for the open-source model ensemble, $D_{\text{OOD}}$ is indeed OOD on the training distribution. We can further abstract a set of observable OOD behavioral patterns from its performance on $D_{\text{OOD}}$, namely the OOD behavioral signatures of open-source models.

Let the open-source model collection be denoted as $\mathcal{M}_{\text{open}}$. For each metric dimension $k$, we define a unified degradation band from the confidence intervals $[L_m^{(k)}, U_m^{(k)}]$ of all open-source models:

\begin{equation}
    \begin{split}
        I_{(k)}^{\text{gap}} := [\min_{m \in \mathcal{M}_{\text{open}}} L_m^{(k)},  \max_{m \in \mathcal{M}_{\text{open}}} U_m^{(k)}].
    \end{split}
\end{equation}

This interval $I_{(k)}^{\text{gap}}$ represents the typical performance degradation range exhibited by the open-source model on the $k$th performance metric dimension for $D_{\text{OOD}}$ instances that have been confirmed as OOD.

Let the set of degeneracy bands for all metric dimensions be denoted as:

\begin{equation}
    \begin{split}
        A := \{I_{(k)}^{\text{gap}} | k = 1,\cdots,K\}.
    \end{split}
\end{equation}

We refer to $A$ as the behavioral signature of the open-source model when processing semantic OOD inputs. It reflects the systematic decline pattern in the model's performance from ID to OOD when confronted with semantic OOD. It is crucial to emphasize that Signature $A$ is not the definition of OOD itself; rather, the definition of OOD stems from data construction and training distribution analysis. Signature $A$ is merely an observable "behavioral fingerprint" of OOD's effect on model behavior.

\subsubsection{Behavioral Verification of Closed-Source Models}
For closed-source model $c$, we cannot observe its training data. Therefore, we employ behavioral-side statistical tests to address two questions:
\begin{itemize}
    \item [1.] Does the closed-source model exhibit significant ID $\rightarrow$ OOD degradation on $D_{\text{OOD}}$?
    \item [2.] Does the degradation magnitude of the closed-source model align with the OOD behavioral signature $A$ of the open-source model?
\end{itemize}
    
\textbf{(1) Significance Test of Performance Differences Between Closed-Source Models in ID vs OOD.} For each metric dimension $k$, we define the same degeneracy measure for closed-source models:

\begin{equation}
    \begin{split}
        \Delta_{c, obs}^{(k)} := T_{c, ID}^{(k)} - T_{c, OOD}^{(k)},
    \end{split}
\end{equation}

where $\Delta_{c, obs}^{(k)}$ denotes the observed degradation amount calculated based on the valid ID/OOD classification.

To verify whether this represents OOD behavior rather than random fluctuations, we construct the following hypothesis test for each $k$:

\begin{itemize}
    \item Null hypothesis $H_0$: The closed-source model exhibits no performance difference between ID and OOD:
    \[
        H_0: T_{c, ID}^{(k)} = T_{c, OOD}^{(k)}.
    \]
    \item Alternative hypothesis $H_1$: The closed-source model performs worse on OOD (degradation exists):
    \[
        H_1: T_{c, ID}^{(k)} > T_{c, OOD}^{(k)}.
    \]
\end{itemize}

We achieve this through a permutation test of ID and OOD sample labels. Under the null hypothesis $H_0$, ID/OOD labels should be interchangeable, yielding the null distribution of the degradation amount $\Delta_c^{(k)}$. If the observed $\Delta_c^{(k)}$ falls in the extreme tail of this distribution (\textit{e.g.}, beyond the top $5\%$), reject $H_0$ and conclude that degradation is significant. Therefore, we randomly permute the ID/OOD labels of the original data and recalculate the degradation amount for each permutation:

\[
        \delta_b^{(k)},\quad b=1,\cdots, B,
\]

where $B$ represents the number of samples. These constitute the empirical distribution under the null hypothesis:

\[
        \{\Delta_b^{(k)}\}_{b=1)}^B.
\]

The standard Monte-Carlo p-value correction formula for permutation tests calculated via this empirical distribution is:

\begin{equation}
    \begin{split}
        p_c^{(k)} = P_{H_0}(\Delta^{(k)} \geq \Delta_{c, obs}^{(k)}) \approx \frac{1+\sum_{b=1}^B\mathds{1}(\Delta_b^{(k)} \geq \Delta_{c, obs}^{(k)})}{B+1},
    \end{split}
\end{equation}

when $p_c^{(k)} < 0.05$, we reject $H_0$ as statistically significant. That is, the closed-source model exhibits significant ID $\rightarrow$ OOD performance degradation on $D_{\text{OOD}}$, which is a typical behavioral signal when semantic OOD is applied to the model.

\begin{figure*}[t]
  \centering
   \includegraphics[width=0.9\linewidth]{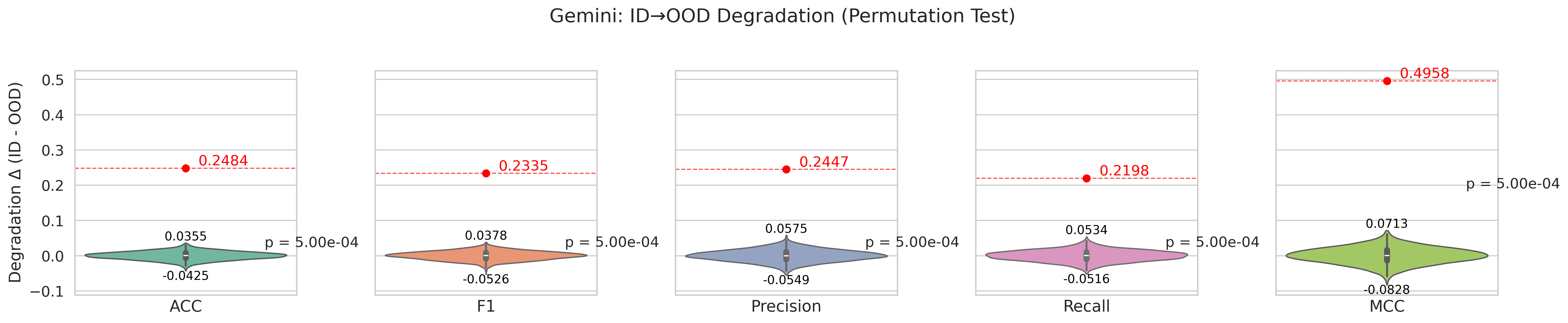}
   \includegraphics[width=0.9\linewidth]{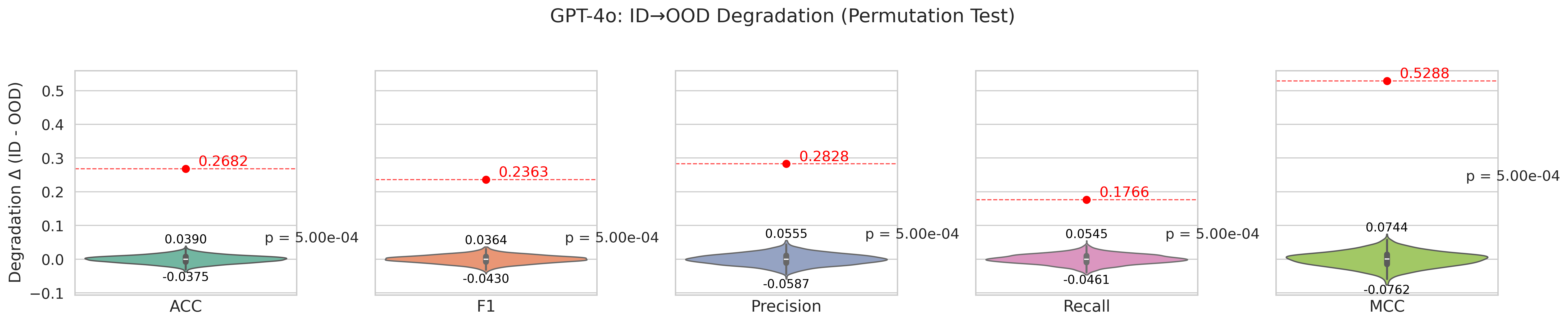}

   \caption{This figure illustrates the distribution of degradation amounts under the null hypothesis $H_0$ (no performance difference between ID and OOD) and the positions of the observed values. As illustrated, for both closed-source models and across all metrics, the observed degradation falls in the extreme tail of the permutation distribution. The permutation p-values are all below $10^{-3}$, and out of 2000 permutations, none produced a degradation larger than the observed value.}
   \label{fig:closed_model_violin_plot}
\end{figure*}

\textbf{Observation Conclusions:} For closed-source models (GPT-4o and Gemini), we calculated the observed measure of ID $\rightarrow$ OOD performance degradation $\Delta_{c, obs}^{(k)}$ across five performance dimensions (Accuracy, F1, Precision, Recall, MCC). We constructed the corresponding null hypothesis distribution based on $B=2000$ label permutations. Figure~\ref{fig:closed_model_violin_plot} presents violin plots illustrating the distribution of degradation amounts and the positions of observed values for each metric under the null hypothesis $H_0$ (no performance difference between ID and OOD). It is evident that for all metrics and both closed-source models, the observed degradation consistently resides in the extreme tail of the permutation distribution: their permutation p-values are all below $10^{-3}$, and across all 2000 permutations, not a single permutation resulted in degradation exceeding the actual observed values. From a statistical inference perspective, this implies: at the commonly used significance level $\alpha=0.05$, the stricter $\alpha=0.01$, and even $\alpha=0.001$, we have sufficient evidence to reject the null hypothesis $H_0: T_{c, ID}^{(k)} = T_{c, OOD} ^{(k)}$ and accept the one-tailed alternative hypothesis $H_0: T_{c, ID}^{(k)} > T_{c, OOD} ^{(k)}$. In other words, under the current data and testing framework, \textbf{the closed-source model exhibits statistically significant performance differences between the ID subset and the OOD-labeled subset, and this difference is highly unlikely to be caused by random label perturbations.}

\textbf{(2) Consistency of OOD Behavioral Signatures Between Closed-Source and Open-Source Models.} We aggregate the bootstrapped degradation values of the open-source models to obtain population-level statistics:
\begin{itemize}
    \item Open-source mean degradation: $\mu_{\text{open}^{(k)}}$.
    \item Open-source variability (natural fluctuation scale): $\sigma_{\text{open}}^{(k)}$.
\end{itemize}

where $\mu_{\text{open}^{(k)}}$ expresses the typical degradation magnitude of open-source models on semantic OOD; $\sigma_{\text{open}}^{(k)}$ represents the natural fluctuation scale of degradation within the open-source community (model variance + sample noise). Furthermore, we define this natural fluctuation scale as the tolerance threshold:

\[
    \eta_k:=\sigma_{\text{open}}^{(k)}.
\]

Suppose the performance degradation of closed-source models falls within the same order of magnitude as the average degradation of open-source models. In that case, we consider their behavior equivalent in this metric dimension.

Comparing the difference between closed-source and open-source models:

\begin{equation}
    \begin{split}
        \delta^{(k)}:=\Delta_c^{(k)} - \mu_{\text{open}}^{(k)}.
    \end{split}
\end{equation}

\begin{itemize}
    \item If $\delta^{(k)} = 0$: closed-source degradation = open-source average degradation;
    \item If $|\delta^{(k)}|$ is small (\textit{e.g.}, within $\pm \eta_k$): closed-source degradation and open-source degradation are "of the same order of magnitude";
    \item If $|\delta^{(k)}|$ is much larger than $\eta_k$, closed-source behavior and open-source OOD behavior are significantly inconsistent.
\end{itemize}

To obtain the distribution of $\delta^{(k)}$ and its $90\%$ confidence interval, we perform $B$ bootstrap resamples on both ID and OOD samples. In each resample, we simultaneously estimate the following quantities:
\begin{itemize}
    \item \textbf{Degradation of the closed-source model:\newline
    $\Delta_{c,b}^{(k)}$, denoting the performance degradation from ID to OOD for the closed-source model in the $b$th resample;}
    \item \textbf{The average degradation of the open-source model population:}\newline
    $\mu_{open,b}^{(k)}$, representing the average degradation of the open-source model population during the $b$th resampling (averaged across the two open-source models).
\end{itemize}

The difference between the two is defined as:

\begin{equation}
    \begin{split}
        \delta^{(k)}_b:=\Delta_{c,b}^{(k)} - \mu_{\text{open},b}^{(k)}, \quad b=1, \cdots, B.
    \end{split}
\end{equation}

Thus we obtain a set $\{\delta^{(k)}_b\}_{b=1}^B$, approximating the sampling distribution of $\delta^{(k)}$.

Equivalence test hypotheses:
\begin{itemize}
    \item Null hypothesis $H_0$ (non-equivalence):
    \[
        H_0: |\delta^{(k)}| \geq \eta_k
    \]
    \textit{i.e.}, the difference between closed-source degradation and open-source degradation on metric $k$ is at least $\eta_k$, indicating behavioral non-equivalence.
    \item Alternative hypothesis $H_1$ (equivalence):
    \[
        H_1: |\delta^{(k)}| < \eta_k
    \]
    \textit{i.e.,} the difference is constrained within the interval $(-\eta_k, \eta_k)$, which can be regarded as equivalent.
\end{itemize}

In practice, we construct a confidence interval using $\{\delta^{(k)}_b \}$ and employ CI to implement the TOST decision. We adopt the standard equivalent test framework, using a $90\%$ confidence interval at a significance level of $\alpha = 0.05$. Specifically, we extract the 5th and 95th percentiles from $\{\delta^{(k)}_b \}$ to obtain the $90\%$ CI:

\begin{equation}
    \begin{split}
        \mathrm{CI}_{90}(\delta^{(k)})=[L_\delta^{(k)},U_\delta^{(k)}].
    \end{split}
\end{equation}

Adopting the TOST equivalence criterion, if:

\begin{equation}
    \begin{split}
        \mathrm{CI}_{90}(\delta^{(k)}) \subset (-\eta_k, \eta_k),
    \end{split}
    \label{appendix:equ_26}
\end{equation}

At the $\alpha=0.05$ significance level, we reject the null hypothesis $H_0: |\delta^{(k)}| \geq \eta_k$ and accept $H_1: |\delta^{(k)}| < \eta_k$. This indicates that: \newline
The difference between closed-source degradation and open-source average degradation is constrained within the "open-source natural fluctuation scale $\sigma_{\text{open}}^{(k)}$" in dimension $k$, rendering it equivalent to open-source OOD behavior.

\begin{figure*}[t]
  \centering
   \includegraphics[width=0.9\linewidth]{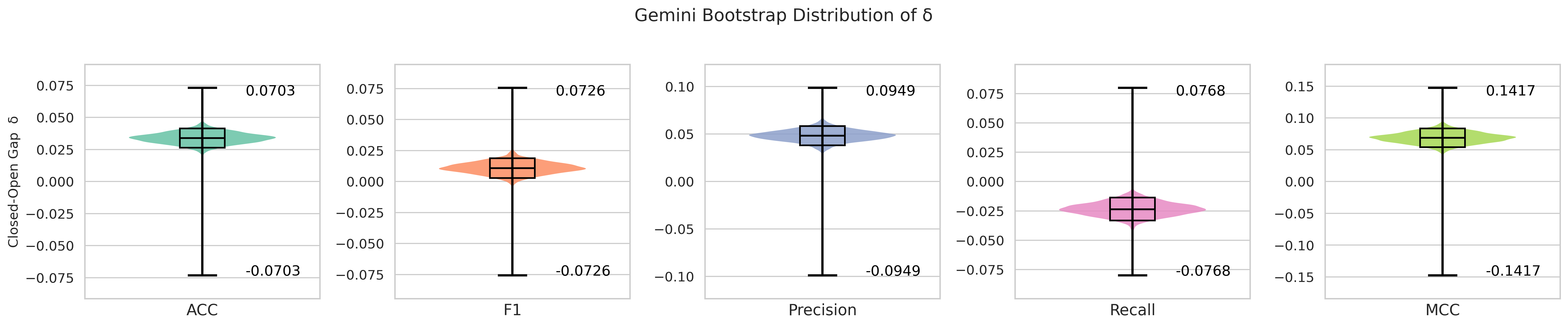}
   \includegraphics[width=0.9\linewidth]{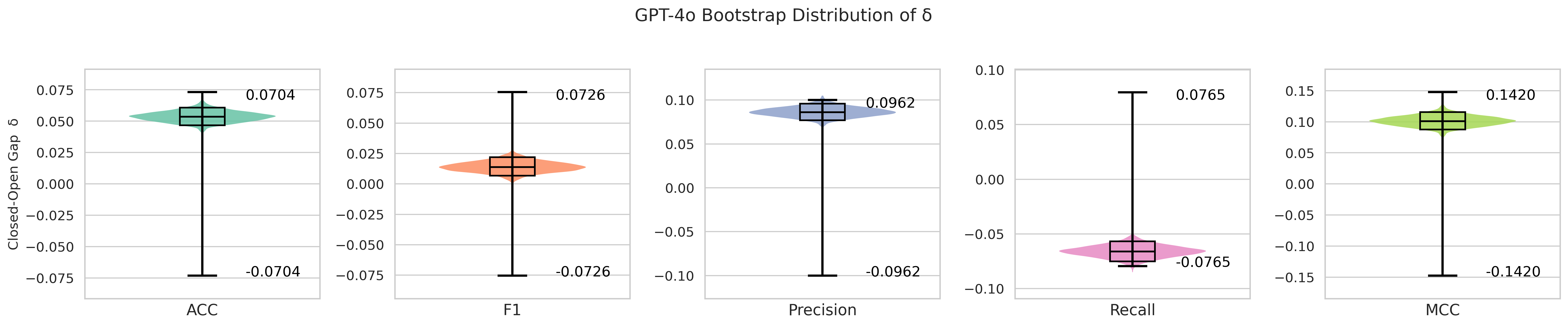}

   \caption{We selected InternVL2-8B and InternVL2.5-8B as open-source reference models and constructed their behavioral signatures based on bootstrap estimates of performance degradation. We then applied the TOST equivalence testing framework to assess the behavioral consistency of closed-source models (GPT-4o and Gemini) under semantic OOD conditions. The bootstrap distributions of $\delta^{(k)}$ for the five evaluation metrics (Accuracy, F1, Precision, Recall, MCC), along with their $90\%$ confidence intervals (CI90) and the corresponding tolerance thresholds $\eta_k$.
}
   \label{fig:closed_model_box_plot}
\end{figure*}

\textbf{Observational Conclusion:} In this study, we selected InternVL2-8B and InternVL2.5-8B as open-source reference models. Based on their bootstrap degeneracy estimates, we constructed open-source behavioral signatures. We then employed the TOST equivalence testing framework to evaluate the behavioral consistency of closed-source models (GPT-4o and Gemini) on semantic OOD data. Figure~\ref{fig:closed_model_box_plot} displays the $\delta^{(k)}$ bootstrap distributions of closed-source models across five metrics (Accuracy, F1, Precision, Recall, MCC), their $90\%$ confidence intervals ($\text{CI}_{90}$), and corresponding tolerance thresholds ($\eta_k$).

The result from GPT-4o is:
\[
\begin{aligned}
&\text{Accuracy:} && \text{CI}_{90} = [4.67\%,\, 6.09\%], && \eta_k = 7.04\% \\
&\text{F1:}       && \text{CI}_{90} = [0.68\%,\, 2.19\%], && \eta_k = 7.26\% \\
&\text{Precision:}&& \text{CI}_{90} = [7.69\%,\, 9.58\%], && \eta_k = 9.62\% \\
&\text{Recall:}   && \text{CI}_{90} = [-7.51\%,\, -5.68\%], && \eta_k = 7.65\% \\
&\text{MCC:}      && \text{CI}_{90} = [8.74\%,\, 11.55\%], && \eta_k = 14.20\%
\end{aligned}
\]

As shown, the $90\%$ confidence intervals for all five metrics fall within the corresponding tolerance ranges $(-\eta_k, \eta_k)$, \textit{i.e.}, Equ.~\ref{appendix:equ_26}, satisfying the equivalence criterion. In other words, across all performance dimensions, the ID $\rightarrow$ OOD degradation of GPT-4o fluctuates at the same order of magnitude as the average degradation of the open-source model population, without exceeding the natural variability exhibited by the open-source models.

Gemini's results also yielded highly consistent conclusions:
\[
\begin{aligned}
&\text{Accuracy:} && \text{CI}_{90} = [2.65\%,\, 4.14\%], && \eta_k = 7.03\% \\
&\text{F1:}       && \text{CI}_{90} = [0.28\%,\, 1.89\%], && \eta_k = 7.26\% \\
&\text{Precision:}&& \text{CI}_{90} = [3.80\%,\, 5.84\%], && \eta_k = 9.49\% \\
&\text{Recall:}   && \text{CI}_{90} = [-3.32\%,\, -1.36\%], && \eta_k = 7.68\% \\
&\text{MCC:}      && \text{CI}_{90} = [5.39\%,\, 8.33\%], && \eta_k = 14.17\%
\end{aligned}
\]
Similarly, all metrics satisfy Equ.~\ref{appendix:equ_26}, indicating that Gemini's degradation also falls stably within the natural fluctuation range of degradation observed in open-source communities.

In summary, GPT-4o and Gemini are statistically equivalent to the open-source OOD behavioral signature across all five performance dimensions—Accuracy, F1, Precision, Recall, and MCC. Their ID $\rightarrow$ OOD degradation magnitudes remain within the same order of magnitude as those of the open-source models, exhibiting no deviations beyond the natural variability scale $(\sigma_{\text{open}})$. These findings indicate that the closed-source models display OOD response patterns on our semantic OOD data that are consistent with those observed in the open-source models.

From the two types of analyses above, we obtain the following conclusions.
Combining (1) the significance tests on ID vs. OOD performance differences for closed-source models and (2) the consistency between closed-source behavior and the open-source OOD behavioral signature, we observe that:
\begin{itemize}
    \item [(1).] Closed-source models exhibit statistically significant ID $\rightarrow$ OOD performance degradation on $D_{\text{OOD}}$.
    \item [(2).] The magnitude of this degradation across all key metrics falls within the natural OOD variability range of the open-source models, consistent with the open-source OOD behavioral signature.
\end{itemize}
 
Thus, on the same semantic OOD dataset, the ID $\rightarrow$ OOD degradation patterns of closed-source models—both in direction and magnitude—align with those of open-source models on known OOD tests. This supports the view that \textbf{closed-source models treat $D_{\text{OOD}}$ as OOD data in a manner equivalent to open-source models.}

\section{Differences between OODBench and Hard Sample}
\label{appendix: hard sample}
Hard samples refer to data instances that exhibit high uncertainty, low confidence, or are prone to misclassification during model training or inference. Although they originate from the training distribution (in-distribution, ID), they typically reside in the tail or low-density regions — the marginal distribution portion. Typical hard samples include: blurred, occluded, or low-resolution images; instances with ambiguous semantic boundaries or highly similar categories (\textit{e.g.}, dogs and wolves); and minority class samples under long-tail distributions. They essentially represent “challenging points” within the distribution, often amenable to fitting through enhanced generalization capabilities, sample reweighting, or increased sample size. In contrast, this paper refines the definition of out-of-distribution (OOD) samples from a human perception perspective, categorizing them into two types: (1) objects in an image that are non-main objects and semantically unrelated to the main object; and (2) variants or abnormal forms of the target object. These samples fundamentally represent “semantic OOD” data that multimodal large models struggle to cover during semantic learning.

To investigate the distinction between OODBench and hard samples, we drew inspiration from the OHEM method~\cite{shrivastava2016training} to construct a set of hard samples on the COCO dataset and evaluated them using nine state-of-the-art models. By comparing the performance (\textit{i.e.}, behavioral patterns) of different models on hard samples versus OODBench, we aim to reveal the differences between the two. In the data partitioning process, we adopted an offline hard sample mining approach, with the specific algorithm detailed in Algorithm~\ref{alg:hard_sample}. First, a sufficiently converged detector is selected to define sample difficulty (YOLOv8-xlarge~\cite{yolov8_ultralytics} was used as the detector in this study). Perform inference on the input image to generate a set of candidate boxes (proposals). For each candidate box, compute the prediction distribution $p(c\mid x)$, where $c \in \{1, \dots, C\}$, and determine the predicted class $\hat{c} = \argmax  p(c\mid x)$. Subsequently, we computed a hardness score for each sample, using the commonly employed Focal Loss~\cite{lin2017focal} value as the metric:

\begin{equation}
    \begin{split}
        h(x) = (1 - p(\hat{c} \mid x))^\gamma \cdot (-\log p(\hat{c} \mid x)), \quad \gamma = 2
    \end{split}
\end{equation}

among these, scores for easily classifiable samples are significantly suppressed, retaining only hard samples. Next, for all proposals within each image, we group them by predicted category and select the top-k samples with the highest hardness within each category to add to the candidate hard sample list. After traversing the entire dataset, the top-k hard samples from each category within every image are aggregated. Finally, the hard sample list is globally sorted by hardness score, and the top-q\% samples from each category are selected to form the final hard sample set.

\begin{algorithm}[h]
   \caption{Hard Sample Mining via Focal Loss Scoring}
   \label{alg:hard_sample}
\begin{algorithmic}
   \STATE {\bfseries Input:} Converged detector $D$, Image set $\mathcal{I}$, Focal loss parameter $\gamma$, Top-$k$ per class, Selection threshold $q\%$.
   \STATE {\bfseries Output:} Hard sample list $\mathcal{H}$
   \STATE Initialize empty list $\mathcal{H} \leftarrow \emptyset$
   \FOR{each image $x \in \mathcal{I}$}
       \STATE Generate proposal set $\mathcal{P}(x)$ using $D$
       \FOR{each proposal $p \in \mathcal{P}(x)$}
           \STATE Obtain class probability distribution $p(c|x), c \in \{1, \dots, C\}$
           \STATE Predicted class $\hat{c} \leftarrow \arg\max_c p(c|x)$
           \STATE Compute hardness score: $h(x) \leftarrow (1 - p(\hat{c}|x))^\gamma \cdot (-\log p(\hat{c}|x))$
       \ENDFOR
       \STATE Group all proposals by predicted class $c$
       \STATE For each class $c$, select top-$k$ proposals with highest $h(x)$
       \STATE $\mathcal{H} \leftarrow \mathcal{H} \cup \{\text{selected proposals}\}$
   \ENDFOR
   \STATE Sort $\mathcal{H}$ by hardness score $h(x)$ in descending order
   \STATE For each class $c$, retain top-$q\%$ proposals as final hard samples
   \STATE \textbf{return} $\mathcal{H}$
\end{algorithmic}
\end{algorithm}

\begin{table*}[ht]
\centering
\caption{This table compares the performance of ID, OOD-H data, and in-Hard samples—segmented based on the COCO dataset—across nine leading SOTA MLLMs. We ensured that all testing configurations remained identical except for the data division method, guaranteeing that any observed differences stem solely from the inherent characteristics of the data.}
\label{tab: hard_sample}
\resizebox{13.5cm}{!}{
\begin{tabular}{cccccccc}
\toprule 
\hline
\centering \multirow{2}{*}{Model} & \centering \multirow{2}{*}{Data Type} & \multicolumn{6}{c}{\textbf{COCO-OOD-Hard Sample} Performance} \\
\cline{3-8}
  & & Num & Accuracy($\%$) & F1($\%$) & Precision($\%$) & Recall($\%$) & MCC($\%$ \\
 \hline
 \centering Random Chance & - & - & $\text{50.00}\%$ & $\text{50.00}\%$ & $\text{50.00}\%$ & $\text{50.00}\%$ & $\text{0.00}\%$ \\
 
 \hline
\textbf{Open-source Models} & & & & & \\
 \hline

 & {ID}  & 6000 & 89.90 & 90.17 & 87.83 & 92.63 & 79.92 \\
 & {OOD-H} & 6000 & 74.45 & 69.53 & 86.12 & 58.30 & 51.67 \\
\multirow{-3}{*}{LLaVA-NeXT-8B} & {Hard Sample} & 5282 & 56.74 & 30.27 & 77.99 & 18.78 & 20.71 \\
 
 \rowcolor{gray!20}
 & {ID} & 6000 & 92.65 & 92.84 & 90.48 & 95.33 & 85.42  \\
 \rowcolor{gray!20}
 & {OOD-H} & 6000 & 76.57 & 74.09 & 82.85 & 67.00 & 54.13  \\
 \rowcolor{gray!20}
\multirow{-3}{*}{DeepSeek-VL-7B-Chat} & {Hard Sample} & 5282 & 62.23 & 47.38 & 78.09 & 34.00 & 29.63 \\

 & {ID} & 6000 & 92.83 & 92.85 & 92.69 & 93.00 & 85.67   \\
 & {OOD-H} & 6000 & 65.18 & 64.92 & 65.41 & 64.43 & 30.37\\
\multirow{-3}{*}{\centering InternVL2-8B} & {Hard Sample} & 5282 & 85.12 & 84.29 & 89.28 & 79.82 & 70.64 \\

\rowcolor{gray!20}
 & {ID} & 6000 & 97.02 & 97.01 & 97.25 & 96.77 & 94.03  \\
 \rowcolor{gray!20}
 & {OOD-H} & 6000 & 79.70 & 78.65 & 82.95 & 74.77 & 59.69 \\
 \rowcolor{gray!20}
\multirow{-3}{*}{InternVL2.5-8B} & {Hard Sample} & 5282 & 85.38 & 84.23 & 91.48 & 78.04 & 71.54 \\

 & {ID} & 6000 & 97.02 & 97.04 & 96.14 & 97.97 & 94.05 \\
 & {OOD-H} & 6000 & 82.02 & 82.91 & 78.99 & 87.23 & 64.38  \\
\multirow{-3}{*}{Llama-3.2-11B-Vision-Instruct} & {Hard Sample} & 5282 & 72.09 & 71.11 & 73.71 & 68.69 & 44.29 \\

\rowcolor{gray!20}
 & {ID} & 6000 & 89.48 & 90.11 & 85.01 & 95.87 & 79.62 \\
 \rowcolor{gray!20}
 & {OOD-H} & 6000 & 77.67 & 75.48 & 83.66 & 68.77 & 56.23 \\
 \rowcolor{gray!20}
\multirow{-3}{*}{Qwen2-VL-7B-Instruct} & {Hard Sample} & 5284 & 72.88 & 66.56 & 86.79 & 53.97 & 49.43 \\

 & {ID} & 6000 & 95.92 & 95.79 & 98.93 & 92.83 & 92.01 \\
 & {OOD-H} & 6000 & 78.28 & 74.04 & 92.03 & 61.93 & 59.86 \\
\multirow{-3}{*}{Qwen2.5-VL-7B-Instruct} & {Hard Sample} & 5282 & 64.27 & 47.69 & 89.03 & 32.56 & 36.93 \\

 \hline
\textbf{Close-source Models} & & & & & \\
 \hline
 & {ID} & 5996 & 91.86 & 92.06 & 89.84 & 94.40 & 83.83  \\
 & {OOD-H} & 5994 & 66.80 & 68.46 & 65.20 & 72.07 & 33.79  \\
\multirow{-3}{*}{Gemini} & {Hard Sample} & 5276 & 85.90 & 85.91 & 85.84 & 85.97 & 71.80 \\

\rowcolor{gray!20}
 & {ID} & 5998 & 93.33 & 93.45 & 91.82 & 95.13 & 86.72 \\
\rowcolor{gray!20}
 & {OOD-H} & 6000 & 70.53 & 72.06 & 68.51 & 76.00 & 41.31 \\
 \rowcolor{gray!20}
\multirow{-3}{*}{GPT-4o} & {Hard Sample} & 5282 & 82.49 & 81.94 & 84.60 & 79.44 & 65.10 \\

 \hline

\bottomrule
\end{tabular}
}
\end{table*}

\begin{figure*}[ht]
    \centering
        \includegraphics[width=0.8\linewidth]{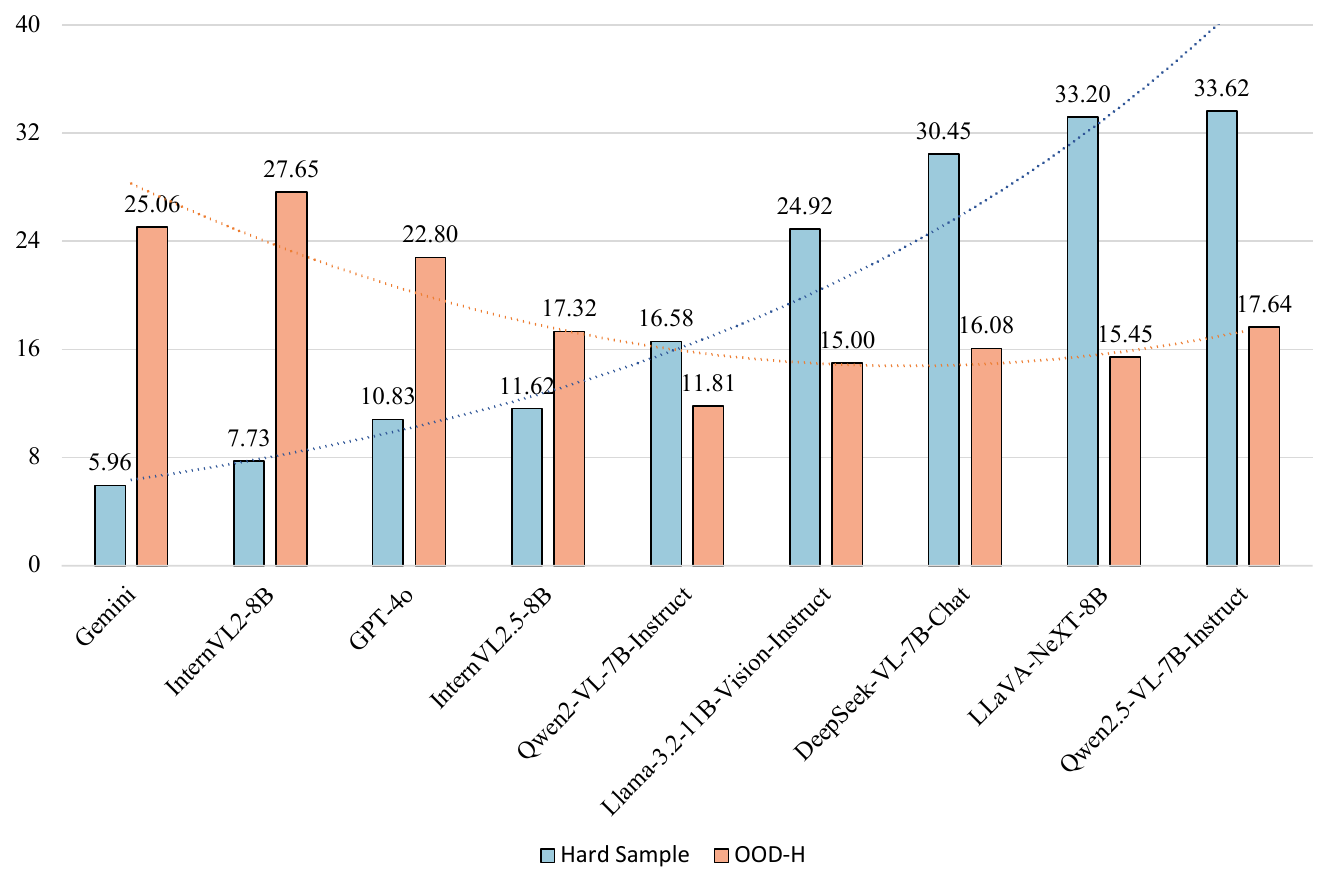}
    \caption{The figure displays the histogram of accuracy difference distributions between ID data and Hard samples, as well as OOD-H data on the COCO dataset. The bars are sorted in ascending order based on the Hard sample difference, with an exponential trendline fitted to characterize the overall trend.}
    \label{fig: hard sample vs oodbench bin}
\end{figure*}

Table~\ref{tab: hard_sample} demonstrates the performance of ID data, OODBench data, and hard samples across nine state-of-the-art multimodal large language models (MLLMs). As shown in Figure~\ref{fig: hard sample vs oodbench bin}, we plotted the performance differences of each model on hard samples and OOD-H compared to in-distribution data, then sorted the results in ascending order based on the magnitude of performance decline on hard samples. Analysis reveals significant variations in the challenge posed by hard samples across models: for instance, Gemini's performance declined by only 5.96\%, while Qwen2.5-VL-Instruct experienced a substantial 33.62\% drop. This indicates that hard samples remain in-distribution data, with performance degradation primarily dependent on a model's robustness and generalization capabilities. Specifically, certain models exhibited pronounced performance degradation on specific hard samples (\textit{e.g.}, DeepSeek-VL-7B-Chat, Llama-3.2-11B-Vision-Instruct, and Qwen2.5-VL-7B-Instruct, all of which exceeded a 30\% decline), while others remained largely unaffected (\textit{e.g.}, Gemini and InternVL2-8B). In contrast, OODBench-defined OOD samples reveal a consistent performance decline across all nine models, with even robust models failing to significantly mitigate this effect. For example, GPT-4o experienced a 27.65\% performance drop on OOD samples. This result indicates that OOD samples exhibit a distribution shift property consistent across models, rather than relying solely on individual model robustness.

\subsection{Statistical distinction between Hard and OOD}
\textbf{Setup.} For each model $m = 1, \dots, M$ (here $M=9$), we compute the performance degradation on hard samples $\Delta_\text{Hard}(m)$ and the performance degradation on our dataset $\Delta_\text{OOD}(m)$. We analyze the following aspects: (i) variability across models, (ii) alignment of performance degradation patterns, and (iii) overlap rate between data partitioned by OODBench and OHEM.

\textbf{Evidence 1 — Variance Test on Full Sets (Hard vs OOD)} \newline
We compare the variance across models using an F-test:
\begin{equation}
    \begin{split}
        F = \frac{\text{Var}(\Delta_{\text{Hard}})}{\text{Var}(\Delta_{\text{OOD}})} \sim F(\nu_1, \nu_2), \nu_1=\nu_2=M-1
    \end{split}
\end{equation}
In our experiments, \(F(8,8) = 4.59\), \(p = 0.0227\), indicating that the variance in performance degradation caused by Hard samples is significantly greater than that caused by OOD samples.\newline
\textbf{Explanation:} Hard samples exhibit greater model dependency, while OOD samples induce a more consistent (model-independent) performance degradation.

\textbf{Evidence 2 — Correlation on full sets (Hard vs OOD)}\newline
We measure the alignment of performance degradation patterns through correlation analysis:

\begin{equation}
    \begin{split}
        r = \text{corr}(\Delta_\text{Hard}, \Delta_\text{OOD})
    \end{split}
\end{equation}

We observe negative correlations (Pearson r = -0.65, p = 0.059; Spearman $\rho$ = -0.58, p = 0.099). These results indicate inconsistencies in the failure patterns of models on Hard and OOD samples: models exhibiting significant performance degradation on Hard samples do not necessarily show similar degradation on OOD samples, and vice versa. Furthermore, from the performance trend lines shown in Figure~\ref{fig: hard sample vs oodbench bin}, we observe an X-shaped decline pattern, where models exhibit significant fluctuations or reversals at certain points. This phenomenon aligns with the correlation test results, further validating the misaligned patterns of performance shifts across Hard and OOD samples. The X-shaped trend lines reflect the models' markedly divergent responses to different sample types (Hard vs. OOD).

\textbf{Hypotheses}
\begin{itemize}
    \item \(\mathbf{H_0}\) (Hard Hypothesis): The OODBench dataset shares the same distribution as the in-distribution Hard subset. In other words, the statistical characteristics of its cross-model performance degradation vector $\Delta V$ (including variance and correlation/alignment with the baseline Hard) should fall within the “empirical distribution of the Hard subset.”
    \item \(\mathbf{H_1}\) (OOD Hypothesis): OODBench does not belong to the Hard subset, and the same-distribution assumption does not hold; its statistical characteristics will significantly deviate from the “Hard baseline.”
\end{itemize}

\textbf{Test statistics. } Record the cross-model performance decline vector (ID performance - Hard sample performance) for the baseline Hard model as $\Delta_\text{Hard}$ (length equal to the number of models $M=9$).

\textbf{Resample details.} We performed resampling based on Algorithm~\ref{alg:hard_sample} to construct the empirical baseline for the Hard subset. Specifically, B=1000 independent samples were conducted, each randomly selecting 500 Hard samples from the COCO validation pool. All samples were selected according to a unified criterion: scoring using focal loss and choosing the top-k samples within each category. Each sampling process was independent, yielding distinct sets of 1000 Hard samples, thereby ensuring the randomness and diversity of the baseline distribution.

\begin{figure*}[t]
    \centering
        \includegraphics[width=\linewidth]{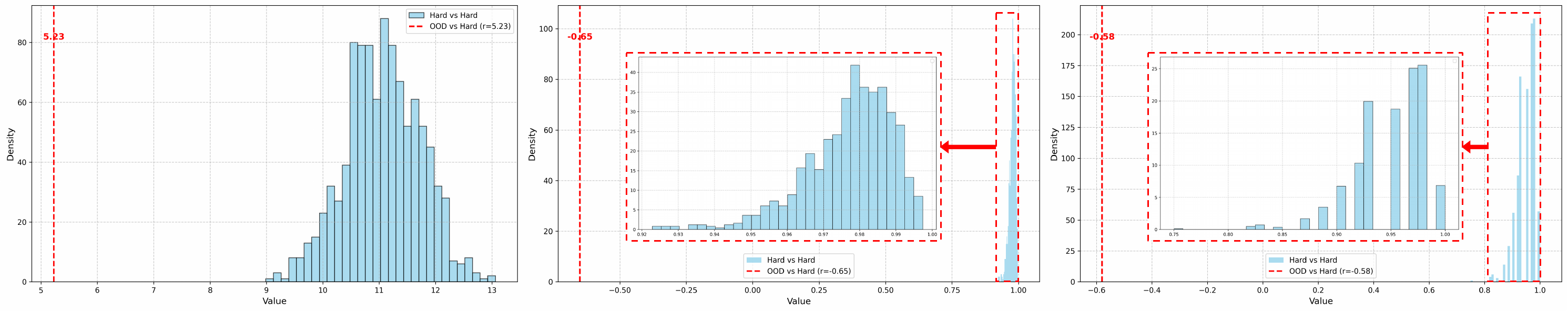}
    \caption{The figure displays the empirical distribution results for variance (left), Pearson correlation coefficient (center), and Spearman correlation coefficient (right). The x-axis of the variance subplot represents variance values, while the x-axes of the Pearson and Spearman subplots respectively denote the corresponding correlation coefficients. The y-axes of all three subplots indicate sample sizes. The red dashed lines in each subplot mark the statistical values of the out-of-distribution (OOD) data.}
    \label{fig: std_correlation_distribution}
\end{figure*}

\textbf{Evidence 3 — Empirical variance test}
\begin{itemize}
\item \textbf{Statistical Measure}: $s^2_\text{OOD} = \text{Var}(\Delta_\text{OOD})$
\item \textbf{Test Direction (one side)}: left tail. Hard samples are typically more model-dependent, exhibiting greater cross-model variance; if $s^2_\text{OOD}$ is significantly smaller, reject $\mathbf{H_0}$.
\item \textbf{Baseline Construction}: Resample $H^{(b)}$ from the COCO validation set using the same mechanism as Algorithm~\ref{alg:hard_sample}'s rule.
\item \textbf{Baseline Distribution}: $S_b=\text{Var}(\Delta_{H^{(b)}})$
\item \textbf{Decision}: If the empirical p-value $\hat{p}_\text{emp} < 0.05$ (\textit{i.e.}, OOD falls below the 5th percentile of the Hard baseline distribution), reject $\mathbf{H_0}$, concluding that the candidate set exhibits greater model-agnostic consistency in cross-model score drops, consistent with OOD characteristics. Results are shown in Figure~\ref{fig: std_correlation_distribution} (left). The empirical p-value is less than 1/(B+1) (\textit{i.e.}, $\hat{p}_\text{emp} < 0.001$), strongly rejecting $\mathbf{H_0}$. This indicates that the statistical characteristics of the candidate set significantly deviate from the Hard baseline.
\end{itemize}

\textbf{Evidence 4 — Empirical correlation test}
\begin{itemize}
\item \textbf{Statistical Measure}: $r_\text{OOD} = \text{corr}(\Delta_\text{Hard}, \Delta_\text{OOD})$
\item \textbf{Test Direction (one side)}: left tail. Hard vs Hard drop-off modes are typically non-negative or positively aligned; if $r_\text{OOD}$ is significantly low, reject $\mathbf{H_0}$.
\item \textbf{Baseline Construction}: Resample $B$ hard subsets $H^{(b)}$ from the COCO validation set and compute their correlation with the baseline hard set.
\item \textbf{Baseline Distribution}: $r_b = \text{corr}(\Delta_\text{Hard}, \Delta_\text{OOD})$
\item \textbf{Decision}: If the empirical p-value $\hat{p}_\text{emp} < 0.05$ (\textit{i.e.}, $r_\text{OOD}$ falls below the 5th percentile of the Hard baseline distribution), reject $\mathbf{H_0}$, indicating that the candidate set does not align with Hard's failure pattern, consistent with OOD's "different failure mode." Results are shown in Figure~\ref{fig: std_correlation_distribution}: The middle subfigure displays the distribution of Pearson correlation coefficients from 1000 resamples, all exceeding 0.9; The right subfigure shows the distribution of Spearman correlation coefficients, all exceeding 0.8. In contrast, the OOD correlations with Hard are -0.65 (Pearson) and -0.58 (Spearman), significantly below the Hard baseline distribution. The empirical p-value is less than 0.001, strongly rejecting $\mathbf{H_0}$ and further demonstrating that the statistical characteristics of the candidate set significantly deviate from the Hard baseline.
\end{itemize}

\textbf{Evidence 5 — Overlap Rate Between Hard Sample and OODBench}
\begin{itemize}
\item \textbf{Statistical Measure}: We calculated the overlap rate between the full COCO-H dataset (5,079 samples) in OODBench and the Hard sample collection (2,641 samples) obtained by sampling according to Algorithm 1. This metric evaluates the consistency between our classified OOD data and the hard samples generated by conventional hard sample mining methods.
\item \textbf{Results}: Statistical analysis revealed only 206 overlapping samples, accounting for 0.0406\% of the total COCO-H dataset and 0.0780\% of the total hard samples.
\item \textbf{Conclusion}: Such a low overlap rate indicates that the OOD data categorized by OODBench fundamentally differs from the hard samples obtained through traditional hard sample mining methods. In other words, OODBench captures more universally applicable out-of-distribution patterns rather than conventional in-distribution hard samples.
\end{itemize}

Taken together, the five lines of evidence consistently demonstrate that the OODBench data cannot be regarded as conventional hard samples. Evidence 1 (variance test on full sets) shows that performance degradation on Hard samples exhibits significantly larger variance across models than on OOD, indicating stronger model dependency for Hard. Evidence 2 (correlation analysis) reveals negative associations between Hard and OOD drop patterns, further corroborated by the X-shaped trend lines, highlighting misaligned failure modes. Evidence 3 (empirical variance test) confirms that the variance of OOD drops is significantly smaller than the empirical Hard baseline, rejecting $\mathbf{H_0}$. Evidence 4 (empirical correlation test) further establishes that OOD correlations with Hard (-0.65, -0.58) fall far below the Hard-vs-Hard baseline (all $>0.9$ Pearson, $>0.8$ Spearman), again strongly rejecting $\mathbf{H_0}$. Finally, Evidence 5 (overlap rate) shows that the intersection between OODBench samples and mined hard samples is minimal (206 out of 5,079), underscoring their fundamental distinction.

Across all five evidences, the results are significant and directionally consistent: OODBench captures distributional shifts and model-agnostic failure modes, whereas Hard samples remain in-distribution and model-dependent. Therefore, our OOD data are not merely hard samples but instead represent a fundamentally different category aligned with out-of-distribution characteristics.

\section{Differences between OODBench and Hallucination}
\label{appendix: hallucination}
\textbf{Out-of-Distribution (OOD) may cause hallucinations, but hallucinations are not necessarily caused by OOD.} We analyse in depth the differences between OOD and hallucinations in terms of concepts, interrelationships and experimental results to demonstrate the distinctiveness of the both.

\textbf{Hallucination definitions in the Large Vision- Language Model (LVLM)}: Hallucination in LVLM denotes a disagreement between factual content of images and corresponding generated textual content, akin to the solely textual hallucination encountered in large language models~\cite{liu2024survey, bai2024hallucination}. \newline
\textbf{Definition of OOD}: The OOD is defined as a test example originating from a distribution different from the training data~\cite{hendrycks2016baseline}. 

In LVLMs, hallucination refers to the inconsistency between the generated text description and the visual content. Our experiment shows that OOD data does not necessarily lead to hallucinations and that it is not fully equivalent to hallucinations. As shown in Tab.~\ref{tab:Main_Results}, OOD data does not necessarily trigger hallucinations. For example, on the OOD-H data, the accuracy of GPT-4o was 65.13\%, implying that 100\% - 65.13\% = 34.87\% of the OOD data led to hallucinations, while the remaining 65.13\% of the OOD data did not trigger hallucinations. This indicates that there is not a one-to-one correspondence between OOD and hallucinations. Second, the causes of hallucinations are multiple, and not all hallucinations are caused by OOD data. For example, the ID data in Tab.~\ref{tab:Main_Results} also triggered hallucinations. This phenomenon was also verified in the BAP experiments in Tab.~\ref{tab: BAP}. Taking GPT-4o as an example, we asked questions on ID data according to Basic-to-Advanced Progress, including existential description, counting description, and logical reasoning, and the performance was 94.84\%, 48.26\%, and 30.83\% in decreasing order. This indicates that as the complexity of the inference task increases, the model becomes more hallucinatory and the match between the generated text and the visual content decreases. The results of the above experiments demonstrate that there is some intersection between OOD and hallucinations, but they are not equivalent.

Prior research~\cite{zhang2023multimodal} has shown that CoT (Chain of Thought) is an effective way to alleviate hallucinations. In our experiments, CoT did not perform as well on OOD data. For example, in Tab.~\ref{tab:Main_Results}, we observed that several models such as DeepSeek-VL-7B-Chat, DeepSeek-VL2-Small, Llama3.2-11B-Vision, Qwen2-7B-Instruct, and GPT-4o failed to improve their performance, but instead, their performance decreased to different degrees after using CoT. This phenomenon may stem from the fact that the distribution of the OOD data deviates from the training distribution, which makes CoT reason with out-of-distribution data as the starting point, thus reinforcing false assumptions, amplifying the inference bias, and ultimately reducing the model performance. We begin by hypothesising that there is a high degree of overlap between the data in OODBench and the hallucinations. Further, CoT, as a typical method to mitigate the hallucination problem, should improve the model performance on this benchmark. However, the experimental results were contrary to expectations: the model performance decreased rather than increased after the introduction of CoT. This phenomenon directly refutes our initial hypothesis that the OODBench data highly overlaps with hallucinations. Instead, it experimentally validates that there is an essential difference between our divided OOD dataset and hallucinations. Together, these experimental results demonstrate that the OOD dataset we constructed is significantly different from the hallucination.

\section{OODBench Sub-Dataset Performance}
\label{appendix: sub-dataset performance}

\textbf{Subdataset classification performance.} Tab.~\ref{tab: coco}, Tab.~\ref{tab: lvis}, Tab.~\ref{tab: nuscenes} and Tab.~\ref{tab: cityscapes} show the performance of the different models on publicly available datasets collected in natural and autonomous driving scenarios, respectively. Where, OOD-S` and OOD-H` denote the initial benchmark datasets. In the Time column, the time superscript indicates the number of graphics cards used in the test, and we used NVIDIA L40s graphics cards in the testing phase.


\begin{table*}[tp]
\centering
\caption{\textbf{Performance on COCO-ID-OOD.} We report the performance of the \textbf{10} leading VLMs on OODBench. All models perform significantly lower on OOD-H than on ID.}
\label{tab: coco}
\resizebox{17cm}{!}{
\begin{tabular}{cccp{2.8cm}ccccccc}
\toprule 
\hline
\centering \multirow{2}{*}{Model} & \centering \multirow{2}{*}{Image Encoder} & \centering \multirow{2}{*}{ Language Model} & \centering \multirow{2}{*}{Data Type} & \multicolumn{6}{c}{\textbf{COCO-ID-OOD} Performance} \\
\cline{5-11}
  & & & & Num & Accuracy($\%$) & F1($\%$) & Precision($\%$) & Recall($\%$) & MCC($\%$) & Time\\
 \hline
 \centering Random Chance & - & - & ID/OOD-S/OOD-H & - & $\text{50.00}\%$ & $\text{50.00}\%$ & $\text{50.00}\%$ & $\text{50.00}\%$ & $\text{0.00}\%$ & -\\
 \hline

 & & \centering \textbf{Open-source Models} & & & & & & \\
 \hline

& & & \multicolumn{1}{c}{ID} & 6000 & 89.90 & 90.17 & 87.83 & 92.63 & 79.92 & 00:24:24$^\text{\textbf{\textit{2}}}$ \\
 & & & \multicolumn{1}{c}{OOD-S} & 6000 & 74.32 & 70.26 & 83.45 & 60.67 & 50.55 & 00:27:16$^\text{\textbf{\textit{2}}}$  \\
 & & & \multicolumn{1}{c}{OOD-H} & 6000 & 74.45 & 69.53 & 86.12 & 58.30 & 51.67 & 00:23:49$^\text{\textbf{\textit{2}}}$  \\
 & & & \multicolumn{1}{c}{OOD-S'} & 7972 & 74.34 & 70.09 & 83.99 & 60.14 & 50.76 & 00:57:40$^\text{\textbf{\textit{1}}}$  \\
 & & & \multicolumn{1}{c}{OOD-H'} & 10158 & 75.31 & 70.67 & 87.01 & 59.50 & 53.36 & 01:07:40$^\text{\textbf{\textit{1}}}$  \\
 & & & \multicolumn{1}{c}{$\text{ID}^{\text{CoT}}$} & 6000 & 74.92 & 79.02 & 67.91 & 94.47 & 54.14 & 00:22:57$^\text{\textbf{\textit{2}}}$  \\
& & & \multicolumn{1}{c}{$\text{OOD-S}^{\text{CoT}}$} & 6000 & 73.43 & 70.13 & 80.09 & 62.37 & 48.06 & 00:28:34$^\text{\textbf{\textit{2}}}$  \\
\multirow{-6}{*}{ LLaVA-NeXT-8B} & \multirow{-6}{*}{ CLIP-L-14} & \multirow{-6}{*}{Llama-3-8B-Instruct} & \multicolumn{1}{c}{$\text{OOD-H}^{\text{CoT}}$} & 6000 & 76.65 & 71.52 & 91.66 & 58.63 & 57.14 & 00:27:14$^\text{\textbf{\textit{2}}}$  \\

\rowcolor{gray!20}
 & & & \multicolumn{1}{c}{ID} & 6000 & 92.65 & 92.84 & 90.48 & 95.33 & 85.42 & 00:32:42$^\text{\textbf{\textit{1}}}$ \\
\rowcolor{gray!20}
 & & & \multicolumn{1}{c}{OOD-S} & 6000 & 75.43 & 72.67 & 81.87 & 65.33 & 51.94 & 00:32:28$^\text{\textbf{\textit{1}}}$  \\
 \rowcolor{gray!20}
& & & \multicolumn{1}{c}{OOD-H} & 6000 & 76.57 & 74.09 & 82.85 & 67.00 & 54.13 & 00:32:26$^\text{\textbf{\textit{1}}}$  \\
\rowcolor{gray!20}
 & & & \multicolumn{1}{c}{OOD-S'} & 7972 & 75.60 & 72.82 & 82.20 & 65.35 & 52.31 & 00:42:57$^\text{\textbf{\textit{1}}}$  \\
 \rowcolor{gray!20}
& & & \multicolumn{1}{c}{OOD-H'} & 10158 & 77.13 & 74.79 & 83.32 & 67.85 & 55.22 & 00:53:33$^\text{\textbf{\textit{1}}}$  \\
\rowcolor{gray!20}
 & & & \multicolumn{1}{c}{$\text{ID}^{\text{CoT}}$} & 6000 & 92.62 & 92.65 & 92.21 & 93.10 & 85.24 & 01:11:34$^\text{\textbf{\textit{1}}}$  \\
\rowcolor{gray!20}
 & & & \multicolumn{1}{c}{$\text{OOD-S}^{\text{CoT}}$} & 6000 & 74.82 & 72.37 & 80.15 & 65.97 & 50.43 & 01:30:32$^\text{\textbf{\textit{1}}}$  \\
 \rowcolor{gray!20}
\multirow{-8}{*}{\centering DeepSeek-VL-7B-Chat} & \multirow{-8}{*}{\centering SigLIP-L $\&$ SAM-B} & \multirow{-8}{*}{\centering DeepSeek-LLM-7B} & \multicolumn{1}{c}{$\text{OOD-H}^{\text{CoT}}$} & 6000 & 73.80 & 70.95 & 79.60 & 64.00 & 48.54 & 01:09:51$^\text{\textbf{\textit{1}}}$  \\

 & & & \multicolumn{1}{c}{ID} & 6000 & 92.88 & 92.74 & 94.62 & 90.93 & 85.83 & 00:29:10$^\text{\textbf{\textit{1}}}$ \\
 & & & \multicolumn{1}{c}{OOD-S} & 6000 & 73.97 & 67.74 & 89.03 & 54.67 & 51.96 & 00:27:58$^\text{\textbf{\textit{1}}}$  \\
& & & \multicolumn{1}{c}{OOD-H} & 6000  & 75.17 & 68.63 & 93.14 & 54.33 & 55.37 & 00:31:00$^\text{\textbf{\textit{1}}}$  \\
 & & & \multicolumn{1}{c}{OOD-S'} & 7972 & 74.02 & 67.82 & 89.10 & 54.74 & 52.07 & 00:43:44$^\text{\textbf{\textit{1}}}$  \\
& & & \multicolumn{1}{c}{OOD-H'} & 10158 & 75.59 & 69.32 & 93.24 & 55.17 & 56.06 & 01:00:14$^\text{\textbf{\textit{1}}}$  \\
 & & & \multicolumn{1}{c}{$\text{ID}^{\text{CoT}}$} & 6000 & 90.90 & 89.99 & 99.96 & 81.83 & 83.18 & 00:57:42$^\text{\textbf{\textit{1}}}$  \\
 & & & \multicolumn{1}{c}{$\text{OOD-S}^{\text{CoT}}$} & 6000 & 69.43 & 57.17 & 95.48 & 40.80 & 47.41 & 00:52:15$^\text{\textbf{\textit{1}}}$  \\
\multirow{-8}{*}{\centering DeepSeek-VL2-Small} & \multirow{-8}{*}{\centering SigLIP-SO400M} & \multirow{-8}{*}{\centering DeepSeekMoE LLM} & \multicolumn{1}{c}{$\text{OOD-H}^{\text{CoT}}$} & 6000 & 68.45 & 54.35 & 98.26 & 37.57 & 46.92 & 00:57:07$^\text{\textbf{\textit{1}}}$  \\

 \rowcolor{gray!20}
& & & \multicolumn{1}{c}{ID} & 6000 & 92.83 & 92.85 & 92.69 & 93.00 & 85.67 & 00:40:41$^\text{\textbf{\textit{2}}}$ \\
 \rowcolor{gray!20}
 & & & \multicolumn{1}{c}{OOD-S} & 6000 & 77.22 & 76.40 & 79.23 & 73.77 & 54.56 & 00:39:38$^\text{\textbf{\textit{2}}}$ \\
  \rowcolor{gray!20}
 & & & \multicolumn{1}{c}{OOD-H} & 6000 & 65.18 & 64.92 & 65.41 & 64.43 & 30.37 & 00:37:58$^\text{\textbf{\textit{2}}}$ \\
  \rowcolor{gray!20}
 & & & \multicolumn{1}{c}{OOD-S'} & 7972 & 77.35 & 76.60 & 79.21 & 74.16 & 54.80 & 00:52:50$^\text{\textbf{\textit{2}}}$ \\
  \rowcolor{gray!20}
 & & & \multicolumn{1}{c}{OOD-H'} & 10160 & 66.22 & 65.99 & 66.44 & 65.55 & 32.44 & 01:03:50$^\text{\textbf{\textit{2}}}$ \\
  \rowcolor{gray!20}
 & & & \multicolumn{1}{c}{$\text{ID}^{\text{CoT}}$} & 6000 & 96.18 & 96.07 & 99.01 & 93.30 & 92.52 & 01:20:50$^\text{\textbf{\textit{2}}}$ \\
  \rowcolor{gray!20}
 & & & \multicolumn{1}{c}{$\text{OOD-S}^{\text{CoT}}$} & 6000 & 77.88 & 73.92 & 90.04 & 62.70 & 58.53 & 01:21:19$^\text{\textbf{\textit{2}}}$ \\
  \rowcolor{gray!20}
 \multirow{-8}{*}{\centering InternVL2-8B} & \multirow{-8}{*}{\centering InternViT-300M-448px} & \multirow{-8}{*}{\centering InternLM2.5-7b-chat} & \multicolumn{1}{c}{$\text{OOD-H}^{\text{CoT}}$} & 6000 & 81.48 & 78.17 & 95.21 & 66.30 & 66.09 & 01:19:54$^\text{\textbf{\textit{2}}}$ \\

 & & & \multicolumn{1}{c}{ID} & 6000 & 97.02 & 97.01 & 97.25 & 96.77 & 94.03 & 00:25:17$^\text{\textbf{\textit{2}}}$ \\
 & & & \multicolumn{1}{c}{OOD-S} & 6000 & 84.88 & 84.18 & 88.29 & 80.43 & 70.04 & 00:26:28$^\text{\textbf{\textit{2}}}$  \\
 & & & \multicolumn{1}{c}{OOD-H} & 6000 & 79.70 & 78.65 & 82.95 & 74.77 & 59.69 & 00:25:20$^\text{\textbf{\textit{2}}}$  \\
 & & & \multicolumn{1}{c}{OOD-S'} & 7972 & 85.00 & 84.31 & 88.35 & 80.63 & 70.26 & 00:35:11$^\text{\textbf{\textit{2}}}$  \\
 & & & \multicolumn{1}{c}{OOD-H'} & 10160 & 80.65 & 79.55 & 84.33 & 75.30 & 61.65 & 00:42:53$^\text{\textbf{\textit{2}}}$  \\
 & & & \multicolumn{1}{c}{$\text{ID}^{\text{CoT}}$} & 6000 & 96.88 & 96.87 & 97.40 & 96.33 & 93.77 & 01:59:05$^\text{\textbf{\textit{2}}}$  \\
 & & & \multicolumn{1}{c}{$\text{OOD-S}^{\text{CoT}}$} & 6000 & 82.02 & 80.31 & 88.71 & 73.37 & 65.01 & 01:48:03$^\text{\textbf{\textit{2}}}$  \\
\multirow{-8}{*}{\centering InternVL2.5-8B} & \multirow{-8}{*}{\centering InternViT-300M-448px-V2.5} & \multirow{-8}{*}{\centering InternLM2.5-7b-chat} & \multicolumn{1}{c}{$\text{OOD-H}^{\text{CoT}}$} & 6000 & 81.92 & 80.08 & 89.13 & 72.70 & 64.95 & 01:49:08$^\text{\textbf{\textit{2}}}$  \\

 \rowcolor{gray!20}
& & & \multicolumn{1}{c}{ID} & 6000 & 97.02 & 97.04 & 96.14 & 97.97 & 94.05 & 00:56:59$^\text{\textbf{\textit{1}}}$ \\
 \rowcolor{gray!20}
 & & & \multicolumn{1}{c}{OOD-S} & 6000 & 82.87 & 83.05 & 82.16 & 83.97 & 65.75 & 01:00:41$^\text{\textbf{\textit{1}}}$ \\
  \rowcolor{gray!20}
 & & & \multicolumn{1}{c}{OOD-H} & 6000 & 82.02 & 82.91 & 78.99 & 87.23 & 64.38 & 01:01:03$^\text{\textbf{\textit{1}}}$\\
  \rowcolor{gray!20}
 & & & \multicolumn{1}{c}{OOD-S'} & 7972 & 75.70 & 77.81 & 71.60 & 85.20 & 52.36 & 01:10:42$^\text{\textbf{\textit{1}}}$ \\
  \rowcolor{gray!20}
 & & & \multicolumn{1}{c}{OOD-H'} & 10158 & 71.93 & 75.64 & 66.81 & 87.16 & 46.06 & 01:30:22$^\text{\textbf{\textit{1}}}$\\
  \rowcolor{gray!20}
 & & & \multicolumn{1}{c}{$\text{ID}^{\text{CoT}}$} & 6000 & 78.32 & 80.58 & 72.97 & 89.97 & 58.24 & 02:26:01$^\text{\textbf{\textit{1}}}$ \\
  \rowcolor{gray!20}
 & & & \multicolumn{1}{c}{$\text{OOD-S}^{\text{CoT}}$} & 6000 & 67.13 & 69.04 & 65.25 & 73.30 & 34.53 & 02:19:54$^\text{\textbf{\textit{1}}}$ \\
  \rowcolor{gray!20}
 \multirow{-8}{*}{\centering Llama-3.2-11B-Vision-Instruct} & \multirow{-8}{*}{\centering ViT-H-14 } & \multirow{-8}{*}{\centering Llama3.1-8B} & \multicolumn{1}{c}{$\text{OOD-H}^{\text{CoT}}$}& 6000 & 68.78 & 72.68 & 64.62 & 83.03 & 39.19 & 02:16:51$^\text{\textbf{\textit{1}}}$ \\

 & & & \multicolumn{1}{c}{ID} & 6000 & 89.48 & 90.11 & 85.01 & 95.87 & 79.62 & 00:06:54$^\text{\textbf{\textit{2}}}$ \\
 & & & \multicolumn{1}{c}{OOD-S} & 6000 & 81.57 & 79.98 & 87.52 & 73.63 & 63.94 & 00:06:59$^\text{\textbf{\textit{2}}}$ \\ 
 & & & \multicolumn{1}{c}{OOD-H} & 6000 & 77.67 & 75.48 & 83.66 & 68.77 & 56.23 & 00:07:00$^\text{\textbf{\textit{2}}}$ \\
 & & & \multicolumn{1}{c}{OOD-S'} & 7972 & 81.18 & 79.49 & 87.35 & 72.93 & 63.24 & 00:09:13$^\text{\textbf{\textit{2}}}$  \\
 & & & \multicolumn{1}{c}{OOD-H'} & 10160 & 78.74 & 76.78 & 84.58 & 70.30 & 58.32 & 00:11:44$^\text{\textbf{\textit{2}}}$ \\
 & & & \multicolumn{1}{c}{$\text{ID}^{\text{CoT}}$} & 6000 & 92.08 & 92.43 & 88.53 & 96.70 & 84.53 & 00:29:54$^\text{\textbf{\textit{2}}}$  \\
 & & & \multicolumn{1}{c}{$\text{OOD-S}^{\text{CoT}}$} & 6000 & 82.43 & 81.73 & 85.13 & 78.60 & 65.06 & 00:45:11$^\text{\textbf{\textit{2}}}$ \\
\multirow{-8}{*}{\centering Qwen2-VL-7B-Instruct} & \multirow{-8}{*}{\centering CLIP-L-14} & \multirow{-8}{*}{\centering Qwen2-7B} & \multicolumn{1}{c}{$\text{OOD-H}^{\text{CoT}}$} & 6000 & 76.82 & 75.87 & 79.10 & 72.90 & 53.80 & 00:44:57$^\text{\textbf{\textit{2}}}$  \\

\rowcolor{gray!20}
 & & & \multicolumn{1}{c}{ID} & 6000 & 95.92 & 95.79 & 98.93 & 92.83 & 92.01 & 00:11:15$^\text{\textbf{\textit{1}}}$ \\
\rowcolor{gray!20}
 & & & \multicolumn{1}{c}{OOD-S} & 6000 & 79.10 & 75.00 & 93.30 & 62.70 & 61.61 & 00:11:10$^\text{\textbf{\textit{1}}}$ \\ 
 \rowcolor{gray!20}
 & & & \multicolumn{1}{c}{OOD-H} & 6000 & 78.28 & 74.04 & 92.03 & 61.93 & 59.86 & 00:11:16$^\text{\textbf{\textit{1}}}$ \\
\rowcolor{gray!20}
 & & & \multicolumn{1}{c}{OOD-S'} & 7972 & 78.81 & 74.54 & 93.39 & 62.02 & 61.18 & 00:14:19$^\text{\textbf{\textit{1}}}$  \\
 \rowcolor{gray!20}
 & & & \multicolumn{1}{c}{OOD-H'} & 10160 & 79.42 & 75.52 & 93.13 & 63.52 & 62.05 & 00:18:40$^\text{\textbf{\textit{1}}}$ \\
\rowcolor{gray!20}
 & & & \multicolumn{1}{c}{$\text{ID}^{\text{CoT}}$} & 6000 & 95.45 & 95.23 & 99.96 & 90.93 & 91.27 & 01:22:11$^\text{\textbf{\textit{1}}}$  \\
\rowcolor{gray!20}
 & & & \multicolumn{1}{c}{$\text{OOD-S}^{\text{CoT}}$} & 6000 & 76.48 & 70.35 & 95.17 & 55.80 & 58.18 & 01:24:46$^\text{\textbf{\textit{1}}}$ \\
 \rowcolor{gray!20}
\multirow{-8}{*}{\centering Qwen2.5-VL-7B-Instruct} & \multirow{-8}{*}{\centering ViT(SwiGLU, RMSNorm)} & \multirow{-8}{*}{\centering Qwen2.5 LLM} & \multicolumn{1}{c}{$\text{OOD-H}^{\text{CoT}}$} & 6000 & 77.20 & 70.89 & 98.00 & 55.53 & 60.36 & 01:26:11$^\text{\textbf{\textit{1}}}$  \\

 \hline

& & \centering \textbf{Closed-source Models} & & & & & & & \\

\hline

 & & & \multicolumn{1}{c}{ID} & 5996 & 91.86 & 92.06 & 89.84 & 94.40 & 83.83 & 01:41:54 \\
 & & & \multicolumn{1}{c}{OOD-S} & 5998 & 75.88 & 76.44 & 74.70 & 78.26 & 51.81 & 01:47:49  \\
& & & \multicolumn{1}{c}{OOD-H} & 5994 & 66.80 & 68.46 & 65.20 & 72.07 & 33.79 & 01:49:24  \\
 & & & \multicolumn{1}{c}{OOD-S'} & 7966 & 75.50 & 76.28 & 73.91 & 78.81 & 51.10 & 02:16:19 \\
 & & & \multicolumn{1}{c}{OOD-H'} & 10150 & 67.02 & 68.71 & 65.37 & 72.41 & 34.25 & 02:59:57 \\
 & & & \multicolumn{1}{c}{$\text{ID}^{\text{CoT}}$} & 5996 & 96.28 & 96.22 & 97.70 & 94.80 & 92.60 & 02:21:18  \\
 & & & \multicolumn{1}{c}{$\text{OOD-S}^{\text{CoT}}$} & 5998 & 78.26 & 76.04 & 84.69 & 68.99 & 57.52 & 02:29:46  \\
\multirow{-8}{*}{\centering Gemini} & \multirow{-8}{*}{\centering -} & \multirow{-8}{*}{\centering -} & \multicolumn{1}{c}{$\text{OOD-H}^{\text{CoT}}$} & 5994 & 79.53 & 77.85 & 84.82 & 71.94 & 59.75 & 02:26:29  \\

 \rowcolor{gray!20}
 & & & \multicolumn{1}{c}{ID} & 5998 & 93.33 & 93.45 & 91.82 & 95.13 & 86.72 & 03:53:40 \\
 \rowcolor{gray!20}
 & & & \multicolumn{1}{c}{OOD-S} & 5992 & 76.72 & 78.22 & 73.48 & 83.61 & 53.95 & 04:38:08 \\
 \rowcolor{gray!20}
 & & & \multicolumn{1}{c}{OOD-H} & 6000 & 70.53 & 72.06 & 68.51 & 76.00 & 41.31 & 03:47:18 \\
 \rowcolor{gray!20}
 & & & \multicolumn{1}{c}{OOD-S'} & 7958 & 77.51 & 78.97 & 74.16 & 84.44 & 55.55 & 05:35:00 \\
 \rowcolor{gray!20}
 & & & \multicolumn{1}{c}{OOD-H'} & 10146 & 66.51 & 69.82 & 63.54 & 77.47 & 33.84 & 07:08:36 \\
 \rowcolor{gray!20}
 & & & \multicolumn{1}{c}{$\text{ID}^{\text{CoT}}$} & 6000 & 82.92 & 84.48 & 77.39 & 93.00 & 67.21 & 06:37:40 \\
 \rowcolor{gray!20}
 & & & \multicolumn{1}{c}{$\text{OOD-S}^{\text{CoT}}$} & 5994 & 67.52 & 66.39 & 68.78 & 64.16 & 35.11 & 06:50:23 \\
 \rowcolor{gray!20}
 \multirow{-8}{*}{\centering GPT-4o} & \multirow{-8}{*}{\centering - } & \multirow{-8}{*}{\centering GPT-4 } & \multicolumn{1}{c}{$\text{OOD-H}^{\text{CoT}}$} & 5996 & 69.31 & 69.72 & 68.81 & 70.65 & 38.64 & 05:58:43 \\

 \hline
\bottomrule
\end{tabular}
}
\end{table*}


\begin{table*}[tp]
\centering
\caption{\textbf{Performance on LVIS-ID-OOD.} We report the performance of the \textbf{10} leading VLMs on OODBench. All models perform significantly lower on OOD-H than on ID.}
\label{tab: lvis}
\resizebox{17cm}{!}{
\begin{tabular}{cccp{2.8cm}ccccccc}
\toprule 
\hline
\centering \multirow{2}{*}{Model} & \centering \multirow{2}{*}{Image Encoder} & \centering \multirow{2}{*}{ Language Model} & \centering \multirow{2}{*}{Data Type} & \multicolumn{6}{c}{\textbf{LVIS-ID-OOD} Performance} \\
\cline{5-11}
  & & & & Num & Accuracy($\%$) & F1($\%$) & Precision($\%$) & Recall($\%$) & MCC($\%$) & Time\\
 \hline
 \centering Random Chance & - & - & ID/OOD-S/OOD-H & - & $\text{50.00}\%$ & $\text{50.00}\%$ & $\text{50.00}\%$ & $\text{50.00}\%$ & $\text{0.00}\%$ & -\\
 
 \hline
 & & \centering \textbf{Open-source Models} & & & & & & \\
 \hline

 & & & \multicolumn{1}{c}{ID} & 6000 & 85.17 & 85.06 & 85.67 & 84.47 & 70.34 & 00:24:54$^\text{\textbf{\textit{2}}}$ \\
 & & & \multicolumn{1}{c}{OOD-S} & 6000 & 47.90 & 35.20 & 46.55 & 28.30 & -4.57 & 00:29:54$^\text{\textbf{\textit{2}}}$  \\
 & & & \multicolumn{1}{c}{OOD-H} & 2436 & 40.23 & 30.86 & 36.60 & 26.68 & -20.30 & 00:10:45$^\text{\textbf{\textit{2}}}$  \\
 & & & \multicolumn{1}{c}{OOD-S'} & 13282 & 47.10 & 35.08 & 45.40 & 28.58 & -6.24 & 01:49:00$^\text{\textbf{\textit{1}}}$  \\
 & & & \multicolumn{1}{c}{$\text{ID}^{\text{CoT}}$} & 6000 & 72.45 & 75.57 & 67.88 & 85.23 & 46.44 & 00:25:28$^\text{\textbf{\textit{2}}}$  \\
 & & & \multicolumn{1}{c}{$\text{OOD-S}^{\text{CoT}}$} & 6000 & 49.65 & 41.62 & 49.52 & 35.90 & -0.73 & 00:36:39$^\text{\textbf{\textit{2}}}$  \\
\multirow{-6}{*}{ LLaVA-NeXT-8B} & \multirow{-6}{*}{ CLIP-L-14} & \multirow{-6}{*}{Llama-3-8B-Instruct} & \multicolumn{1}{c}{$\text{OOD-H}^{\text{CoT}}$} & 2436 & 48.07 & 46.47 & 47.95 & 45.07 & -3.87 & 00:12:55$^\text{\textbf{\textit{2}}}$  \\

\rowcolor{gray!20}
 & & & \multicolumn{1}{c}{ID} & 6000 & 89.85 & 90.14 & 87.63 & 92.80 & 79.84 & 00:27:58$^\text{\textbf{\textit{1}}}$ \\
\rowcolor{gray!20}
 & & & \multicolumn{1}{c}{OOD-S} & 6000 & 47.42 & 41.65 & 46.78 & 37.53 & -5.27 & 00:33:43$^\text{\textbf{\textit{1}}}$  \\
 \rowcolor{gray!20}
& & & \multicolumn{1}{c}{OOD-H} & 2436 & 42.69 & 41.39 & 42.35 & 40.48 & -14.63 & 00:08:51$^\text{\textbf{\textit{1}}}$  \\
\rowcolor{gray!20}
 & & & \multicolumn{1}{c}{OOD-S'} & 13282 & 47.33 & 42.06 & 46.74 & 38.23 & -5.42 & 00:49:51$^\text{\textbf{\textit{1}}}$  \\
 \rowcolor{gray!20}
\rowcolor{gray!20}
 & & & \multicolumn{1}{c}{$\text{ID}^{\text{CoT}}$} & 6000 & 88.42 & 88.23 & 89.65 & 86.87 & 76.87 & 01:15:31$^\text{\textbf{\textit{1}}}$  \\
\rowcolor{gray!20}
 & & & \multicolumn{1}{c}{$\text{OOD-S}^{\text{CoT}}$} & 6000 & 46.18 & 38.51 & 44.91 & 33.70 & -7.88 & 01:58:38$^\text{\textbf{\textit{1}}}$  \\
 \rowcolor{gray!20}
\multirow{-7}{*}{\centering DeepSeek-VL-7B-Chat} & \multirow{-7}{*}{\centering SigLIP-L $\&$ SAM-B} & \multirow{-7}{*}{\centering DeepSeek-LLM-7B} & \multicolumn{1}{c}{$\text{OOD-H}^{\text{CoT}}$} & 2436 & 35.18 & 30.16 & 32.69 & 28.00 & -29.95 & 00:41:00$^\text{\textbf{\textit{1}}}$  \\

 & & & \multicolumn{1}{c}{ID} & 6000 & 87.75 & 87.13 & 91.74 & 82.97 & 75.85 & 00:30:31$^\text{\textbf{\textit{1}}}$ \\
 & & & \multicolumn{1}{c}{OOD-S} & 6000 & 47.23 & 30.90 & 44.75 & 23.60 & -6.28 & 00:29:37$^\text{\textbf{\textit{1}}}$  \\
& & & \multicolumn{1}{c}{OOD-H} & 2436 & 39.66 & 27.08 & 34.21 & 22.41 & -22.04 & 00:12:38$^\text{\textbf{\textit{1}}}$  \\
 & & & \multicolumn{1}{c}{OOD-S'} & 13282 & 46.94 & 31.50 & 44.43 & 24.39 & -6.85 & 01:40:48$^\text{\textbf{\textit{1}}}$  \\
 & & & \multicolumn{1}{c}{$\text{ID}^{\text{CoT}}$} & 6000 & 82.47 & 78.82 & 99.49 & 65.27 & 69.15 & 00:51:51$^\text{\textbf{\textit{1}}}$  \\
 & & & \multicolumn{1}{c}{$\text{OOD-S}^{\text{CoT}}$} & 6000 & 44.30 & 15.61 & 32.19 & 10.30 & -15.55 & 00:49:32$^\text{\textbf{\textit{1}}}$  \\
\multirow{-7}{*}{\centering DeepSeek-VL2-Small} & \multirow{-7}{*}{\centering SigLIP-SO400M} & \multirow{-7}{*}{\centering DeepSeekMoE LLM} & \multicolumn{1}{c}{$\text{OOD-H}^{\text{CoT}}$} & 2436 & 34.85 & 8.64 & 14.45 & 6.16 & -36.99 & 00:24:25$^\text{\textbf{\textit{1}}}$  \\

 \rowcolor{gray!20}
 & & & \multicolumn{1}{c}{ID} & 6000 & 87.20 & 87.31 & 86.57 & 88.07 & 74.41 & 00:40:08$^\text{\textbf{\textit{2}}}$ \\
 \rowcolor{gray!20}
 & & & \multicolumn{1}{c}{OOD-S} & 6000 & 49.12 & 46.16 & 49.01 & 43.63 & -1.78 & 00:41:49$^\text{\textbf{\textit{2}}}$ \\
 \rowcolor{gray!20}
 & & & \multicolumn{1}{c}{OOD-H} & 2436 & 34.56 & 33.80 & 34.20 & 33.42 & -30.88 & 00:16:27$^\text{\textbf{\textit{2}}}$ \\
 \rowcolor{gray!20}
 & & & \multicolumn{1}{c}{OOD-S'} & 13284 & 48.34 & 45.32 & 48.14 & 42.82 & -3.33 & 01:32:38$^\text{\textbf{\textit{2}}}$ \\
 \rowcolor{gray!20}
 & & & \multicolumn{1}{c}{$\text{ID}^{\text{CoT}}$} & 6000 & 94.55 & 94.28 & 99.12 & 89.90 & 89.49 & 01:18:29$^\text{\textbf{\textit{2}}}$ \\
 \rowcolor{gray!20}
 & & & \multicolumn{1}{c}{$\text{OOD-S}^{\text{CoT}}$}  & 6000 & 48.23 & 38.32 & 47.40 & 32.17 & -3.73 & 01:31:17$^\text{\textbf{\textit{2}}}$ \\
 \rowcolor{gray!20}
 \multirow{-7}{*}{\centering InternVL2-8B} & \multirow{-7}{*}{\centering InternViT-300M-448px} & \multirow{-7}{*}{\centering InternLM2.5-7b-chat} & \multicolumn{1}{c}{$\text{OOD-H}^{\text{CoT}}$} & 2436 & 43.23 & 36.47 & 41.40 & 32.59 & -13.86 & 00:35:24$^\text{\textbf{\textit{2}}}$ \\

 & & & \multicolumn{1}{c}{ID} & 6000 & 94.17 & 94.09 & 95.28 & 92.93 & 88.36 & 00:25:50$^\text{\textbf{\textit{2}}}$ \\
 & & & \multicolumn{1}{c}{OOD-S} & 6000 & 52.87 & 47.65 & 53.58 & 42.90 & 5.85 & 00:28:34$^\text{\textbf{\textit{2}}}$  \\
 & & & \multicolumn{1}{c}{OOD-H} & 2436 & 39.24 & 36.04 & 38.05 & 34.24 & -21.62 & 00:10:55$^\text{\textbf{\textit{2}}}$  \\
 & & & \multicolumn{1}{c}{OOD-S'} & 13284 & 51.71 & 46.47 & 52.12 & 41.93 & 3.48 & 01:03:03$^\text{\textbf{\textit{2}}}$ \\
 & & & \multicolumn{1}{c}{$\text{ID}^{\text{CoT}}$} & 6000 & 94.52 & 94.40 & 96.42 & 92.47 & 89.11 & 01:47:40$^\text{\textbf{\textit{2}}}$ \\
 & & & \multicolumn{1}{c}{$\text{OOD-S}^{\text{CoT}}$} & 6000 & 52.30 & 46.24 & 52.97 & 41.03 & 4.72 & 01:48:51$^\text{\textbf{\textit{2}}}$  \\
\multirow{-7}{*}{\centering InternVL2.5-8B} & \multirow{-7}{*}{\centering InternViT-300M-448px-V2.5} & \multirow{-7}{*}{\centering InternLM2.5-7b-chat} & \multicolumn{1}{c}{$\text{OOD-H}^{\text{CoT}}$} & 2436 & 41.67 & 37.76 & 40.47 & 35.39 & -16.80 & 00:44:34$^\text{\textbf{\textit{2}}}$  \\

\rowcolor{gray!20}
 & & & \multicolumn{1}{c}{ID} & 6000 & 92.60 & 92.88 & 89.54 & 96.47 & 85.46 & 00:57:42$^\text{\textbf{\textit{1}}}$ \\
 \rowcolor{gray!20}
 & & & \multicolumn{1}{c}{OOD-S} & 6000 & 50.62 & 51.70 & 50.59 & 52.87 & 1.23 & 01:01:18$^\text{\textbf{\textit{1}}}$ \\
 \rowcolor{gray!20}
 & & & \multicolumn{1}{c}{OOD-H} & 2436 & 40.23 & 43.61 & 41.28 & 46.22 & -19.68 & 00:23:34$^\text{\textbf{\textit{1}}}$\\
 \rowcolor{gray!20}
 & & & \multicolumn{1}{c}{OOD-S'} & 13282 & 49.99 & 51.01 & 49.99 & 52.07 & -0.02 & 02:14:40$^\text{\textbf{\textit{1}}}$ \\
 \rowcolor{gray!20}
 & & & \multicolumn{1}{c}{$\text{ID}^{\text{CoT}}$} & 6000 & 75.57 & 78.71 & 69.74 & 90.33 & 53.52 & 02:18:56$^\text{\textbf{\textit{1}}}$ \\
 \rowcolor{gray!20}
 & & & \multicolumn{1}{c}{$\text{OOD-S}^{\text{CoT}}$} & 6000 & 49.87 & 51.31 & 49.87 & 52.83 & -0.27 & 02:16:41$^\text{\textbf{\textit{1}}}$ \\
 \rowcolor{gray!20}
 \multirow{-7}{*}{\centering Llama-3.2-11B-Vision-Instruct} & \multirow{-7}{*}{\centering ViT-H-14 } & \multirow{-7}{*}{\centering Llama3.1-8B} & \multicolumn{1}{c}{$\text{OOD-H}^{\text{CoT}}$} & 2436 & 41.67 & 37.76 & 40.47 & 35.39 & -16.80 & 00:44:34$^\text{\textbf{\textit{2}}}$ \\

 & & & \multicolumn{1}{c}{ID} & 6000 & 86.78 & 87.63 & 82.37 & 93.60 & 74.26 & 00:06:51$^\text{\textbf{\textit{2}}}$ \\
 & & & \multicolumn{1}{c}{OOD-S} & 6000 & 55.60 & 53.15 & 56.25 & 50.37 & 11.26 & 00:06:56$^\text{\textbf{\textit{2}}}$ \\ 
 & & & \multicolumn{1}{c}{OOD-H} & 2436 & 49.14 & 50.50 & 49.18 & 51.89 & -1.73 & 00:02:50$^\text{\textbf{\textit{2}}}$ \\
 & & & \multicolumn{1}{c}{OOD-S'} & 13284 & 54.91 & 52.80 & 55.39 & 50.45 & 9.86 & 00:15:26$^\text{\textbf{\textit{2}}}$  \\
 & & & \multicolumn{1}{c}{$\text{ID}^{\text{CoT}}$} & 6000 & 89.03 & 89.58 & 85.33 & 94.27 & 78.50 & 00:35:25$^\text{\textbf{\textit{2}}}$  \\
 & & & \multicolumn{1}{c}{$\text{OOD-S}^{\text{CoT}}$} & 6000 & 56.23 & 55.69 & 56.39 & 55.00 & 12.47 & 00:54:50$^\text{\textbf{\textit{2}}}$ \\
\multirow{-7}{*}{\centering Qwen2-VL-7B-Instruct} & \multirow{-7}{*}{\centering CLIP-L-14} & \multirow{-7}{*}{\centering Qwen2-7B} & \multicolumn{1}{c}{$\text{OOD-H}^{\text{CoT}}$} & 2436 & 48.11 & 49.68 & 48.22 & 51.23 & -3.78 & 00:19:20$^\text{\textbf{\textit{2}}}$  \\

\rowcolor{gray!20}
 & & & \multicolumn{1}{c}{ID} & 6000 & 93.03 & 92.81 & 95.94 & 89.87 & 86.24 & 00:11:14$^\text{\textbf{\textit{1}}}$ \\
\rowcolor{gray!20}
 & & & \multicolumn{1}{c}{OOD-S} & 6000 & 52.63 & 41.43 & 54.27 & 33.50 & 5.70 & 00:10:55$^\text{\textbf{\textit{1}}}$ \\ 
 \rowcolor{gray!20}
 & & & \multicolumn{1}{c}{OOD-H} & 2436 & 46.31 & 39.67 & 45.26 & 35.30 & -7.57 & 00:04:20$^\text{\textbf{\textit{1}}}$ \\
\rowcolor{gray!20}
 & & & \multicolumn{1}{c}{OOD-S'} & 13284 & 52.29 & 41.66 & 53.60 & 34.08 & 4.92 & 00:23:43$^\text{\textbf{\textit{1}}}$  \\
\rowcolor{gray!20}
 & & & \multicolumn{1}{c}{$\text{ID}^{\text{CoT}}$} & 6000 & 92.70 & 92.13 & 99.92 & 85.47 & 86.31 & 01:31:56$^\text{\textbf{\textit{1}}}$  \\
\rowcolor{gray!20}
 & & & \multicolumn{1}{c}{$\text{OOD-S}^{\text{CoT}}$} & 6000 & 51.38 & 35.05 & 52.78 & 26.23 & 3.20 & 01:48:30$^\text{\textbf{\textit{1}}}$ \\
 \rowcolor{gray!20}
\multirow{-7}{*}{\centering Qwen2.5-VL-7B-Instruct} & \multirow{-7}{*}{\centering ViT(SwiGLU, RMSNorm)} & \multirow{-7}{*}{\centering Qwen2.5 LLM} & \multicolumn{1}{c}{$\text{OOD-H}^{\text{CoT}}$} & 2436 & 43.19 & 32.36 & 39.98 & 27.18 & -14.39 & 00:46:05$^\text{\textbf{\textit{1}}}$  \\

 \hline
& & \centering \textbf{Closed-source Models} & & & & & & & \\
\hline

 & & & \multicolumn{1}{c}{ID} & 5996 & 87.68 & 88.13 & 84.98 & 91.53 & 75.57 & 01:41:01 \\
 & & & \multicolumn{1}{c}{OOD-S} & 5994 & 51.58 & 52.05 & 51.55 & 52.55 & 3.17 & 01:47:55  \\
& & & \multicolumn{1}{c}{OOD-H} & 2436 & 35.18 & 37.47 & 36.19 & 38.83 & -29.72 & 00:42:07  \\
 & & & \multicolumn{1}{c}{OOD-S'} & 13272 & 50.63 & 51.49 & 50.61 & 52.40 & 1.27 & 03:51:01 \\
 & & & \multicolumn{1}{c}{$\text{ID}^{\text{CoT}}$}& 5996 & 92.48 & 92.42 & 93.15 & 91.69 & 84.97 & 02:21:13  \\
 & & & \multicolumn{1}{c}{$\text{OOD-S}^{\text{CoT}}$} & 5994 & 50.10 & 43.41 & 50.13 & 38.27 & 0.21 & 02:38:14  \\
\multirow{-7}{*}{\centering Gemini} & \multirow{-7}{*}{\centering -} & \multirow{-7}{*}{\centering -} & \multicolumn{1}{c}{$\text{OOD-H}^{\text{CoT}}$} & 2436 & 40.72 & 37.27 & 39.58 & 35.22 & -18.67 & 01:03:47  \\

\rowcolor{gray!20}
 & & & \multicolumn{1}{c}{ID} & 5960 & 93.09 & 93.26 & 91.00 & 95.64 & 86.29 & 05:46:52 \\
 \rowcolor{gray!20}
 & & & \multicolumn{1}{c}{OOD-S} & 5990 & 58.30 & 60.15 & 57.59 & 62.94 & 16.67 & 04:29:12 \\
 \rowcolor{gray!20}
 & & & \multicolumn{1}{c}{OOD-H} & 2432 & 41.69 & 45.08 & 42.61 & 47.86 & -16.74 & 01:45:24 \\
 \rowcolor{gray!20}
 & & & \multicolumn{1}{c}{OOD-S'} & 13282 & 57.90 & 58.53 & 57.66 & 59.43 & 15.80 & 08:40:59 \\
 \rowcolor{gray!20}
 & & & \multicolumn{1}{c}{$\text{ID}^{\text{CoT}}$} & 6000 & 81.10 & 83.09 & 75.16 & 92.90 & 64.01 & 06:35:48 \\
 \rowcolor{gray!20}
 & & & \multicolumn{1}{c}{$\text{OOD-S}^{\text{CoT}}$} & 5958 & 51.88 & 48.98 & 52.12 & 46.19 & 3.78 & 08:42:01 \\
 \rowcolor{gray!20}
 \multirow{-7}{*}{\centering GPT-4o} & \multirow{-7}{*}{\centering - } & \multirow{-7}{*}{\centering GPT-4 } & \multicolumn{1}{c}{$\text{OOD-H}^{\text{CoT}}$} & 2392 & 45.07 & 46.76 & 45.36 & 48.24 & -9.89 & 04:43:32\\
 \hline

\bottomrule
\end{tabular}
}
\end{table*}


\begin{table*}[tp]
\centering
\caption{\textbf{Performance on nuScenes-ID-OOD.} We report the performance of the \textbf{10} leading VLMs on OODBench. All models perform significantly lower on OOD-H than on ID.}
\label{tab: nuscenes}
\resizebox{17cm}{!}{
\begin{tabular}{cccp{2.8cm}ccccccc}
\toprule 
\hline
\centering \multirow{2}{*}{Model} & \centering \multirow{2}{*}{Image Encoder} & \centering \multirow{2}{*}{ Language Model} & \centering \multirow{2}{*}{Data Type} & \multicolumn{6}{c}{\textbf{nuScenes-ID-OOD} Performance} \\
\cline{5-11}
  & & & & Num & Accuracy($\%$) & F1($\%$) & Precision($\%$) & Recall($\%$) & MCC($\%$) & Time\\
 \hline
 \centering Random Chance & - & - & ID/OOD-S/OOD-H & - & $\text{50.00}\%$ & $\text{50.00}\%$ & $\text{50.00}\%$ & $\text{50.00}\%$ & $\text{0.00}\%$ & -\\
 
 \hline
 & & \centering \textbf{Open-source Models} & & & & & & \\
 \hline

 & & & \multicolumn{1}{c}{ID} & 6000 & 81.43 & 80.05 & 86.49 & 74.50 & 63.48 & 00:27:02$^\text{\textbf{\textit{2}}}$ \\
 & & & \multicolumn{1}{c}{OOD-S} & 6000 & 62.68 & 45.38 & 84.62 & 31.00 & 32.79 & 00:31:30$^\text{\textbf{\textit{2}}}$  \\
 & & & \multicolumn{1}{c}{OOD-H} & 6000 & 59.03 & 40.34 & 74.20 & 27.70 & 23.18 & 00:30:31$^\text{\textbf{\textit{2}}}$  \\
 & & & \multicolumn{1}{c}{OOD-S'} & 25798 & 62.57 & 45.42 & 83.83 & 31.15 & 32.32 & 03:45:13$^\text{\textbf{\textit{1}}}$  \\
 & & & \multicolumn{1}{c}{OOD-H'} & 9882 & 60.58 & 42.24 & 79.02 & 28.82 & 27.41 & 01:23:29$^\text{\textbf{\textit{1}}}$  \\
 & & & \multicolumn{1}{c}{$\text{ID}^{\text{CoT}}$} & 6000 & 50.90 & 58.48 & 50.66 & 69.17 & 1.93 & 01:11:18$^\text{\textbf{\textit{1}}}$\\
 & & & \multicolumn{1}{c}{$\text{OOD-S}^{\text{CoT}}$} & 6000 & 52.12 & 41.14 & 53.38 & 33.47 & 4.56 & 01:06:26$^\text{\textbf{\textit{2}}}$  \\
\multirow{-6}{*}{ LLaVA-NeXT-8B} & \multirow{-6}{*}{ CLIP-L-14} & \multirow{-6}{*}{Llama-3-8B-Instruct} & \multicolumn{1}{c}{$\text{OOD-H}^{\text{CoT}}$} & 6000 & 54.87 & 42.58 & 58.51 & 33.47 & 10.77 & 00:57:57$^\text{\textbf{\textit{2}}}$  \\

\rowcolor{gray!20}
 & & & \multicolumn{1}{c}{ID} & 6000 & 83.40 & 82.09 & 89.14 & 76.07 & 67.53 & 00:33:53$^\text{\textbf{\textit{1}}}$ \\
\rowcolor{gray!20}
 & & & \multicolumn{1}{c}{OOD-S} & 6000 & 61.12 & 47.16 & 73.57 & 34.70 & 26.19 & 00:34:22$^\text{\textbf{\textit{1}}}$  \\
 \rowcolor{gray!20}
& & & \multicolumn{1}{c}{OOD-H} & 6000 & 52.95 & 28.77 & 59.19 & 19.00 & 8.04 & 00:23:41$^\text{\textbf{\textit{1}}}$  \\
\rowcolor{gray!20}
 & & & \multicolumn{1}{c}{OOD-S'} & 25798 & 60.47 & 46.13 & 72.38 & 33.85 & 24.73 & 02:03:01$^\text{\textbf{\textit{1}}}$  \\
 \rowcolor{gray!20}
 & & & \multicolumn{1}{c}{OOD-H'} & 9882 & 54.62 & 30.33 & 65.28 & 19.75 & 12.91 & 00:55:52$^\text{\textbf{\textit{1}}}$ \\
 \rowcolor{gray!20}
 & & & \multicolumn{1}{c}{$\text{ID}^{\text{CoT}}$} & 6000 & 74.20 & 71.15 & 80.68 & 63.63 & 49.52 & 02:20:00$^\text{\textbf{\textit{1}}}$\\
\rowcolor{gray!20}
 & & & \multicolumn{1}{c}{$\text{OOD-S}^{\text{CoT}}$} & 6000 & 59.18 & 42.66 & 71.68 & 30.37 & 22.47 & 03:04:27$^\text{\textbf{\textit{1}}}$  \\
 \rowcolor{gray!20}
\multirow{-8}{*}{\centering DeepSeek-VL-7B-Chat} & \multirow{-8}{*}{\centering SigLIP-L $\&$ SAM-B} & \multirow{-8}{*}{\centering DeepSeek-LLM-7B} & \multicolumn{1}{c}{$\text{OOD-H}^{\text{CoT}}$} & 6000 & 46.30 & 24.54 & 41.26 & 17.47 & -9.06 & 02:44:52$^\text{\textbf{\textit{1}}}$  \\

 & & & \multicolumn{1}{c}{ID} & 6000 & 76.08 & 70.42 & 92.27 & 56.93 & 56.47 & 00:43:54$^\text{\textbf{\textit{1}}}$ \\
 & & & \multicolumn{1}{c}{OOD-S} & 6000 & 58.73 & 32.64 & 88.76 & 20.00 & 27.62 & 00:41:44$^\text{\textbf{\textit{1}}}$  \\
& & & \multicolumn{1}{c}{OOD-H} & 6000 & 51.72 & 15.81 & 61.68 & 9.07 & 6.58 & 00:44:10$^\text{\textbf{\textit{1}}}$  \\
 & & & \multicolumn{1}{c}{OOD-S'} & 25798 & 58.09 & 30.67 & 88.69 & 18.54 & 26.44 & 03:27:02$^\text{\textbf{\textit{1}}}$  \\
 & & & \multicolumn{1}{c}{OOD-H'} & 9882 & 52.74 & 16.49 & 70.81 & 9.33 & 11.05 & 01:32:27$^\text{\textbf{\textit{1}}}$ \\
 & & & \multicolumn{1}{c}{$\text{ID}^{\text{CoT}}$} & 6000 & 77.77 & 71.59 & 99.12 & 56.03 & 61.66 & 00:55:41$^\text{\textbf{\textit{1}}}$\\
 & & & \multicolumn{1}{c}{$\text{OOD-S}^{\text{CoT}}$} & 6000 & 58.73 & 31.90 & 91.19 & 19.33 & 28.37 & 01:01:12$^\text{\textbf{\textit{1}}}$  \\
\multirow{-8}{*}{\centering DeepSeek-VL2-Small} & \multirow{-8}{*}{\centering SigLIP-SO400M} & \multirow{-8}{*}{\centering DeepSeekMoE LLM} & \multicolumn{1}{c}{$\text{OOD-H}^{\text{CoT}}$} & 6000 & 51.05 & 11.56 & 59.81 & 6.40 & 4.67 & 01:01:55$^\text{\textbf{\textit{1}}}$  \\

 \rowcolor{gray!20}
 & & & \multicolumn{1}{c}{ID} & 6000 & 89.77 & 89.13 & 95.09 & 83.87 & 80.09 & 00:29:17$^\text{\textbf{\textit{2}}}$ \\
 \rowcolor{gray!20}
 & & & \multicolumn{1}{c}{OOD-S} & 6000 & 71.27 & 68.14 & 76.45 & 61.47 & 43.37 & 00:31:04$^\text{\textbf{\textit{2}}}$ \\
 \rowcolor{gray!20}
 & & & \multicolumn{1}{c}{OOD-H} & 6000 & 51.08 & 43.20 & 51.50 & 37.20 & 2.26 & 00:30:46$^\text{\textbf{\textit{2}}}$ \\
 \rowcolor{gray!20}
 & & & \multicolumn{1}{c}{OOD-S'} & 25800 & 70.90 & 67.60 & 76.26 & 60.71 & 42.70 & 02:12:54$^\text{\textbf{\textit{2}}}$ \\
 \rowcolor{gray!20}
  & & & \multicolumn{1}{c}{OOD-H'} & 9884 & 52.08 & 44.57 & 52.86 & 38.53 & 4.33 & 00:50:14$^\text{\textbf{\textit{2}}}$ \\
  \rowcolor{gray!20}
 & & & \multicolumn{1}{c}{$\text{ID}^{\text{CoT}}$} & 6000 & 90.20 & 89.30 & 98.32 & 81.80 & 81.56 & 03:56:47$^\text{\textbf{\textit{2}}}$ \\
 \rowcolor{gray!20}
 & & & \multicolumn{1}{c}{$\text{OOD-S}^{\text{CoT}}$} & 6000 & 64.73 & 50.24 & 85.30 & 35.60 & 36.26 & 04:20:00$^\text{\textbf{\textit{2}}}$ \\
 \rowcolor{gray!20}
 \multirow{-8}{*}{\centering InternVL2-8B} & \multirow{-8}{*}{\centering InternViT-300M-448px} & \multirow{-8}{*}{\centering InternLM2.5-7b-chat} & \multicolumn{1}{c}{$\text{OOD-H}^{\text{CoT}}$} & 6000 & 62.23 & 45.48 & 81.75 & 31.50 & 31.02 & 04:24:46$^\text{\textbf{\textit{2}}}$ \\

 & & & \multicolumn{1}{c}{ID} & 6000 & 82.70 & 83.75 & 78.96 & 89.17 & 65.95 & 00:23:12$^\text{\textbf{\textit{2}}}$ \\
 & & & \multicolumn{1}{c}{OOD-S} & 6000 & 70.13 & 71.02 & 68.97 & 73.20 & 40.34 & 00:24:38$^\text{\textbf{\textit{2}}}$  \\
 & & & \multicolumn{1}{c}{OOD-H} & 6000 & 36.23 & 39.61 & 37.62 & 41.83 & -27.71 & 00:24:29$^\text{\textbf{\textit{2}}}$ \\
 & & & \multicolumn{1}{c}{OOD-S'} & 25800 & 69.65 & 70.57 & 68.50 & 72.78 & 39.38 & 01:45:43$^\text{\textbf{\textit{2}}}$ \\
 & & & \multicolumn{1}{c}{OOD-H'} & 9884 & 36.68 & 39.90 & 37.97 & 42.05 & -26.80 & 00:40:22$^\text{\textbf{\textit{2}}}$ \\
 & & & \multicolumn{1}{c}{$\text{ID}^{\text{CoT}}$} & 6000 & 78.00 & 79.16 & 75.19 & 83.57 & 56.35 & 02:53:33$^\text{\textbf{\textit{2}}}$\\
 & & & \multicolumn{1}{c}{$\text{OOD-S}^{\text{CoT}}$} & 6000 & 65.18 & 57.43 & 73.89 & 46.97 & 32.61 & 03:21:22$^\text{\textbf{\textit{2}}}$ \\
\multirow{-8}{*}{\centering InternVL2.5-8B} & \multirow{-8}{*}{\centering InternViT-300M-448px-V2.5} & \multirow{-8}{*}{\centering InternLM2.5-7b-chat} & \multicolumn{1}{c}{$\text{OOD-H}^{\text{CoT}}$} & 6000 & 54.82 & 48.99 & 56.24 & 43.40 & 9.89 & 03:24:39$^\text{\textbf{\textit{2}}}$  \\

\rowcolor{gray!20}
 & & & \multicolumn{1}{c}{ID} & 6000 & 88.78 & 88.74 & 89.06 & 88.43 & 77.57 & 01:11:40$^\text{\textbf{\textit{1}}}$ \\
 \rowcolor{gray!20}
 & & & \multicolumn{1}{c}{OOD-S} & 6000 & 69.58 & 64.56 & 77.34 & 55.40 & 40.84 & 01:15:19$^\text{\textbf{\textit{1}}}$ \\
 \rowcolor{gray!20}
 & & & \multicolumn{1}{c}{OOD-H} & 6000 & 61.72 & 60.02 & 62.81 & 57.47 & 23.52 & 01:18:18$^\text{\textbf{\textit{1}}}$ \\
 \rowcolor{gray!20}
 & & & \multicolumn{1}{c}{OOD-S'} & 25798 & 66.30 & 63.34 & 69.43 & 58.24 & 33.02 & 04:51:10$^\text{\textbf{\textit{1}}}$ \\
 \rowcolor{gray!20}
 & & & \multicolumn{1}{c}{OOD-H'} & 9882 & 58.86 & 60.28 & 58.28 & 62.42 & 17.77 & 01:51:33$^\text{\textbf{\textit{1}}}$ \\
 \rowcolor{gray!20}
 & & & \multicolumn{1}{c}{$\text{ID}^{\text{CoT}}$} & 6000 & 71.63 & 72.33 & 70.60 & 74.13 & 43.32 & 02:34:13$^\text{\textbf{\textit{1}}}$ \\
 \rowcolor{gray!20}
 & & & \multicolumn{1}{c}{$\text{OOD-S}^{\text{CoT}}$} & 6000 & 59.33 & 49.90 & 64.97 & 40.50 & 20.15 & 02:27:42$^\text{\textbf{\textit{1}}}$ \\
 \rowcolor{gray!20}
 \multirow{-8}{*}{\centering Llama-3.2-11B-Vision-Instruct} & \multirow{-8}{*}{\centering ViT-H-14 } & \multirow{-8}{*}{\centering Llama3.1-8B} & \multicolumn{1}{c}{$\text{OOD-H}^{\text{CoT}}$} & 6000 & 59.98 & 55.56 & 62.46 & 50.03 & 20.37 & 02:29:42$^\text{\textbf{\textit{1}}}$ \\

 & & & \multicolumn{1}{c}{ID}  & 6000 & 92.72 & 92.58 & 94.39 & 90.83 & 85.49 & 00:50:55$^\text{\textbf{\textit{2}}}$ \\
 & & & \multicolumn{1}{c}{OOD-S} & 6000 & 73.53 & 69.84 & 81.16 & 61.30 & 48.54 & 00:51:02$^\text{\textbf{\textit{2}}}$ \\ 
 & & & \multicolumn{1}{c}{OOD-H} & 6000 & 57.55 & 50.05 & 60.79 & 42.53 & 15.83 & 00:50:54$^\text{\textbf{\textit{2}}}$ \\
 & & & \multicolumn{1}{c}{OOD-S'} & 25800 & 73.12 & 69.32 & 80.75 & 60.72 & 47.74 & 03:40:30$^\text{\textbf{\textit{2}}}$ \\
 & & & \multicolumn{1}{c}{OOD-H'} & 9884 & 59.02 & 51.87 & 62.85 & 44.15 & 18.91 & 01:24:18$^\text{\textbf{\textit{2}}}$ \\
 & & & \multicolumn{1}{c}{$\text{ID}^{\text{CoT}}$} & 6000 & 87.82 & 88.04 & 86.47 & 89.67 & 75.69 & 01:35:02$^\text{\textbf{\textit{2}}}$ \\
 & & & \multicolumn{1}{c}{$\text{OOD-S}^{\text{CoT}}$} & 6000 & 68.58 & 59.95 & 82.66 & 47.03 & 41.19 & 02:02:32$^\text{\textbf{\textit{2}}}$ \\
\multirow{-8}{*}{\centering Qwen2-VL-7B-Instruct} & \multirow{-8}{*}{\centering CLIP-L-14} & \multirow{-8}{*}{\centering Qwen2-7B} & \multicolumn{1}{c}{$\text{OOD-H}^{\text{CoT}}$} & 6000 & 57.13 & 47.27 & 61.40 & 38.43 & 15.38 & 02:02:22$^\text{\textbf{\textit{2}}}$ \\

\rowcolor{gray!20}
 & & & \multicolumn{1}{c}{ID}  & 6000 & 80.62 & 81.20 & 78.82 & 83.73 & 61.35 & 01:09:27$^\text{\textbf{\textit{1}}}$ \\
\rowcolor{gray!20}
 & & & \multicolumn{1}{c}{OOD-S} & 6000 & 63.77 & 51.17 & 78.44 & 37.97 & 32.14 & 00:50:18$^\text{\textbf{\textit{1}}}$ \\ 
 \rowcolor{gray!20}
 & & & \multicolumn{1}{c}{OOD-H} & 6000 & 60.12 & 46.02 & 71.18 & 34.00 & 23.73 & 00:59:19$^\text{\textbf{\textit{1}}}$ \\
\rowcolor{gray!20}
 & & & \multicolumn{1}{c}{OOD-S'} & 25800 & 63.54 & 50.57 & 78.51 & 37.30 & 31.82 & 03:19:44$^\text{\textbf{\textit{1}}}$ \\
 \rowcolor{gray!20}
 & & & \multicolumn{1}{c}{OOD-H'} & 9884 & 61.72 & 47.97 & 74.85 & 35.30 & 27.61 & 01:20:31$^\text{\textbf{\textit{1}}}$ \\
\rowcolor{gray!20}
 & & & \multicolumn{1}{c}{$\text{ID}^{\text{CoT}}$} & 6000 & 91.20 & 90.35 & 100.00 & 82.40 & 83.71 & 02:05:30$^\text{\textbf{\textit{1}}}$ \\
\rowcolor{gray!20}
 & & & \multicolumn{1}{c}{$\text{OOD-S}^{\text{CoT}}$} & 6000 & 66.52 & 51.83 & 92.31 & 36.03 & 41.67 & 02:06:14$^\text{\textbf{\textit{1}}}$ \\
 \rowcolor{gray!20}
\multirow{-8}{*}{\centering Qwen2.5-VL-7B-Instruct} & \multirow{-8}{*}{\centering ViT(SwiGLU, RMSNorm)} & \multirow{-8}{*}{\centering Qwen2.5 LLM} & \multicolumn{1}{c}{$\text{OOD-H}^{\text{CoT}}$} & 6000 & 63.02 & 45.28 & 87.01 & 30.60 & 34.19 & 02:20:17$^\text{\textbf{\textit{1}}}$ \\

 \hline

& & \centering \textbf{Closed-source Models} & & & & & & & \\
\hline
 & & & \multicolumn{1}{c}{ID} & 6000 & 90.40 & 90.48 & 89.76 & 91.20 & 80.81 & 02:06:56 \\
 & & & \multicolumn{1}{c}{OOD-S} & 6000 & 77.25 & 76.71 & 78.57 & 74.93 & 54.56 & 02:15:23  \\
& & & \multicolumn{1}{c}{OOD-H} & 6000 & 56.95 & 56.74 & 57.02 & 56.47 & 13.90 & 02:17:32  \\
 & & & \multicolumn{1}{c}{OOD-S'} & 25798 & 77.19 & 76.68 & 78.42 & 75.03 & 54.43 & 08:43:47 \\
& & & \multicolumn{1}{c}{OOD-H'} & 9882 & 59.07 & 59.13 & 59.04 & 59.22 & 18.13 & 03:23:56 \\
 & & & \multicolumn{1}{c}{$\text{ID}^{\text{CoT}}$} & 6000 & 92.70 & 92.55 & 94.45 & 90.73 & 85.47 & 02:57:49  \\
 & & & \multicolumn{1}{c}{$\text{OOD-S}^{\text{CoT}}$} & 6000 & 68.55 & 62.16 & 78.01 & 51.67 & 39.42 & 03:03:12  \\
\multirow{-8}{*}{\centering Gemini} & \multirow{-8}{*}{\centering -} & \multirow{-8}{*}{\centering -} & \multicolumn{1}{c}{$\text{OOD-H}^{\text{CoT}}$} & 6000 & 68.93 & 64.36 & 75.47 & 56.10 & 39.18 & 02:58:06 \\

\rowcolor{gray!20}
 & & & \multicolumn{1}{c}{ID} & 5996 & 86.72 & 87.72 & 81.60 & 94.83 & 74.43 & 05:14:11 \\
 \rowcolor{gray!20}
 & & & \multicolumn{1}{c}{OOD-S} & 5994 & 74.94 & 74.55 & 75.73 & 73.41 & 49.91 & 05:51:09 \\
 \rowcolor{gray!20}
 & & & \multicolumn{1}{c}{OOD-H} & 5998 & 58.92 & 61.29 & 57.94 & 65.06 & 17.98 & 05:48:40 \\
 \rowcolor{gray!20}
 & & & \multicolumn{1}{c}{OOD-S'} & 25798 & 75.83 & 75.62 & 76.27 & 74.98 & 51.66 & 24:39:04 \\
 \rowcolor{gray!20}
 & & & \multicolumn{1}{c}{OOD-H'} & 9882 & 58.78 & 61.58 & 57.67 & 66.06 & 17.76 & 09:52:58 \\
 \rowcolor{gray!20}
 & & & \multicolumn{1}{c}{$\text{ID}^{\text{CoT}}$} & 6000 & 70.83 & 75.76 & 64.81 & 91.17 & 45.61 & 07:48:32\\
 \rowcolor{gray!20}
 & & & \multicolumn{1}{c}{$\text{OOD-S}^{\text{CoT}}$} & 5998 & 62.70 & 58.41 & 66.01 & 52.38 & 25.97 & 08:11:22\\
 \rowcolor{gray!20}
 \multirow{-8}{*}{\centering GPT-4o} & \multirow{-8}{*}{\centering - } & \multirow{-8}{*}{\centering GPT-4 } & \multicolumn{1}{c}{$\text{OOD-H}^{\text{CoT}}$} & 5992 & 61.35 & 60.50 & 61.85 & 59.21 & 22.72 & 07:21:03 \\

 \hline

\bottomrule
\end{tabular}
}

\end{table*}



\begin{table*}[tp]
\centering
\caption{\textbf{Performance on Cityscapes-ID-OOD.} We report the performance of the \textbf{10} leading VLMs on OODBench. All models perform significantly lower on OOD-H than on ID.}
\label{tab: cityscapes}
\resizebox{17cm}{!}{
\begin{tabular}{cccp{2.8cm}ccccccc}
\toprule 
\hline
\centering \multirow{2}{*}{Model} & \centering \multirow{2}{*}{Image Encoder} & \centering \multirow{2}{*}{ Language Model} & \centering \multirow{2}{*}{Data Type} & \multicolumn{6}{c}{\textbf{Cityscapes-ID-OOD} Performance} \\
\cline{5-11}
  & & & & Num & Accuracy($\%$) & F1($\%$) & Precision($\%$) & Recall($\%$) & MCC($\%$) & Time\\
 \hline
 \centering Random Chance & - & - & ID/OOD-S/OOD-H & - & $\text{50.00}\%$ & $\text{50.00}\%$ & $\text{50.00}\%$ & $\text{50.00}\%$ & $\text{0.00}\%$ & -\\
 
 \hline
 & & \centering \textbf{Open-source Models} & & & & & & \\
 \hline

 & & & \multicolumn{1}{c}{ID} & 3556 & 81.21 & 82.14 & 78.29 & 86.39 & 62.77 & 00:18:39$^\text{\textbf{\textit{2}}}$ \\
 & & & \multicolumn{1}{c}{OOD-S} & 3820 & 78.22 & 76.55 & 82.91 & 71.10 & 57.02 & 00:17:43$^\text{\textbf{\textit{2}}}$ \\
 & & & \multicolumn{1}{c}{OOD-H} & 3836 & 66.79 & 58.09 & 78.70 & 46.04 & 36.90 & 00:18:58$^\text{\textbf{\textit{2}}}$ \\
 & & & \multicolumn{1}{c}{$\text{ID}^{\text{CoT}}$} & 3554 & 59.23 & 69.67 & 55.47 & 93.64 & 25.44 & 00:32:01$^\text{\textbf{\textit{1}}}$  \\
 & & & \multicolumn{1}{c}{$\text{OOD-S}^{\text{CoT}}$} & 3820 & 65.58 & 70.01 & 62.02 & 80.37 & 32.61 & 00:20:07$^\text{\textbf{\textit{2}}}$  \\
\multirow{-6}{*}{ LLaVA-NeXT-8B} & \multirow{-6}{*}{ CLIP-L-14} & \multirow{-6}{*}{Llama-3-8B-Instruct} & \multicolumn{1}{c}{$\text{OOD-H}^{\text{CoT}}$} & 3836 & 70.52 & 65.47 & 79.00 & 55.89 & 42.91 & 00:19:56$^\text{\textbf{\textit{2}}}$  \\

\rowcolor{gray!20}
 & & & \multicolumn{1}{c}{ID} & 3554 & 91.73 & 91.37 & 95.52 & 87.56 & 83.75 & 00:25:48$^\text{\textbf{\textit{1}}}$ \\
\rowcolor{gray!20}
 & & & \multicolumn{1}{c}{OOD-S} & 3818 & 78.71 & 75.95 & 87.23 & 67.26 & 58.98 & 00:27:57$^\text{\textbf{\textit{1}}}$  \\
 \rowcolor{gray!20}
& & & \multicolumn{1}{c}{OOD-H} & 3834 & 61.53 & 54.99 & 66.25 & 47.00 & 24.10 & 00:26:57$^\text{\textbf{\textit{1}}}$  \\
\rowcolor{gray!20}
 & & & \multicolumn{1}{c}{$\text{ID}^{\text{CoT}}$} & 3554 & 85.45 & 85.18 & 86.80 & 83.62 & 70.95 & 00:49:00$^\text{\textbf{\textit{1}}}$  \\
\rowcolor{gray!20}
 & & & \multicolumn{1}{c}{$\text{OOD-S}^{\text{CoT}}$} & 3818 & 77.71 & 75.33 & 84.35 & 68.05 & 56.49 & 00:51:54$^\text{\textbf{\textit{1}}}$  \\
 \rowcolor{gray!20}
\multirow{-6}{*}{\centering DeepSeek-VL-7B-Chat} & \multirow{-6}{*}{\centering SigLIP-L $\&$ SAM-B} & \multirow{-6}{*}{\centering DeepSeek-LLM-7B} & \multicolumn{1}{c}{$\text{OOD-H}^{\text{CoT}}$} & 3834 & 63.43 & 54.60 & 71.99 & 43.97 & 29.16 & 01:25:13$^\text{\textbf{\textit{1}}}$ \\

 & & & \multicolumn{1}{c}{ID} & 3554 & 80.95 & 76.55 & 99.55 & 62.18 & 66.79 & 00:32:18$^\text{\textbf{\textit{1}}}$ \\
 & & & \multicolumn{1}{c}{OOD-S} & 3818 & 77.42 & 71.46 & 97.12 & 56.52 & 60.37 & 00:30:48$^\text{\textbf{\textit{1}}}$  \\
& & & \multicolumn{1}{c}{OOD-H} & 3834 & 65.39 & 47.49 & 98.36 & 31.30 & 42.07 & 00:31:18$^\text{\textbf{\textit{1}}}$  \\
 & & & \multicolumn{1}{c}{$\text{ID}^{\text{CoT}}$} & 3554 & 79.07 & 73.52 & 100.00 & 58.13 & 64.01 & 00:42:59$^\text{\textbf{\textit{1}}}$  \\
 & & & \multicolumn{1}{c}{$\text{OOD-S}^{\text{CoT}}$} & 3818 & 70.77 & 59.80 & 95.73 & 43.48 & 49.58 & 00:52:23$^\text{\textbf{\textit{1}}}$  \\
\multirow{-6}{*}{\centering DeepSeek-VL2-Small} & \multirow{-6}{*}{\centering SigLIP-SO400M} & \multirow{-6}{*}{\centering DeepSeekMoE LLM} & \multicolumn{1}{c}{$\text{OOD-H}^{\text{CoT}}$} & 3834 & 57.90 & 27.88 & 97.20 & 16.28 & 28.53 & 00:50:00$^\text{\textbf{\textit{1}}}$ \\

 \rowcolor{gray!20}
 & & & \multicolumn{1}{c}{ID} & 3556 & 95.95 & 95.82 & 98.98 & 92.86 & 92.08 & 00:17:34$^\text{\textbf{\textit{2}}}$ \\
 \rowcolor{gray!20}
 & & & \multicolumn{1}{c}{OOD-S} & 3820 & 86.94 & 86.20 & 91.38 & 81.57 & 74.30 & 00:21:15$^\text{\textbf{\textit{2}}}$ \\
 \rowcolor{gray!20}
 & & & \multicolumn{1}{c}{OOD-H} & 3836 & 77.48 & 74.78 & 84.95 & 66.79 & 56.25 & 00:19:21$^\text{\textbf{\textit{2}}}$ \\
 \rowcolor{gray!20}
 & & & \multicolumn{1}{c}{$\text{ID}^{\text{CoT}}$} & 3556 & 90.83 & 89.94 & 99.59 & 82.00 & 82.97 & 02:12:08$^\text{\textbf{\textit{2}}}$ \\
 \rowcolor{gray!20}
 & & & \multicolumn{1}{c}{$\text{OOD-S}^{\text{CoT}}$} & 3820 & 82.38 & 79.44 & 95.38 & 68.06 & 67.60 & 02:23:47$^\text{\textbf{\textit{2}}}$ \\
 \rowcolor{gray!20}
 \multirow{-6}{*}{\centering InternVL2-8B} & \multirow{-6}{*}{\centering InternViT-300M-448px} & \multirow{-6}{*}{\centering InternLM2.5-7b-chat} & \multicolumn{1}{c}{$\text{OOD-H}^{\text{CoT}}$} & 3836 & 77.69 & 71.66 & 98.19 & 56.41 & 61.18 & 02:27:58$^\text{\textbf{\textit{2}}}$ \\

 & & & \multicolumn{1}{c}{ID} & 3556 & 93.53 & 93.61 & 92.43 & 94.83 & 87.09 & 00:14:58$^\text{\textbf{\textit{2}}}$ \\
 & & & \multicolumn{1}{c}{OOD-S} & 3820 & 86.88 & 86.95 & 86.52 & 87.38 & 73.77 & 00:16:55$^\text{\textbf{\textit{2}}}$ \\
 & & & \multicolumn{1}{c}{OOD-H} & 3836 & 72.34 & 71.07 & 74.50 & 67.94 & 44.86 & 00:15:46$^\text{\textbf{\textit{2}}}$  \\
 & & & \multicolumn{1}{c}{$\text{ID}^{\text{CoT}}$} & 3556 & 93.98 & 93.95 & 94.38 & 93.53 & 87.97 & 01:46:20$^\text{\textbf{\textit{2}}}$ \\
 & & & \multicolumn{1}{c}{$\text{OOD-S}^{\text{CoT}}$} & 3820 & 85.73 & 84.80 & 90.75 & 79.58 & 72.01 & 01:55:57$^\text{\textbf{\textit{2}}}$ \\
\multirow{-6}{*}{\centering InternVL2.5-8B} & \multirow{-8}{*}{\centering InternViT-300M-448px-V2.5} & \multirow{-8}{*}{\centering InternLM2.5-7b-chat} & \multicolumn{1}{c}{$\text{OOD-H}^{\text{CoT}}$} & 3836 & 79.25 & 76.99 & 86.38 & 69.45 & 59.66 & 01:58:27$^\text{\textbf{\textit{2}}}$ \\

\rowcolor{gray!20}
 & & & \multicolumn{1}{c}{ID} & 3554 & 92.40 & 92.70 & 89.22 & 96.45 & 85.09 & 00:45:22$^\text{\textbf{\textit{1}}}$ \\
 \rowcolor{gray!20}
 & & & \multicolumn{1}{c}{OOD-S} & 3818 & 73.21 & 76.24 & 68.49 & 85.96 & 48.00 & 00:43:55$^\text{\textbf{\textit{1}}}$ \\
 \rowcolor{gray!20}
 & & & \multicolumn{1}{c}{OOD-H} & 3834 & 65.26 & 69.41 & 62.00 & 78.82 & 31.71 & 00:41:42$^\text{\textbf{\textit{1}}}$ \\
 \rowcolor{gray!20}
 & & & \multicolumn{1}{c}{$\text{ID}^{\text{CoT}}$} & 3554 & 71.05 & 70.41 & 72.00 & 68.88 & 42.13 & 01:32:09$^\text{\textbf{\textit{1}}}$ \\
 \rowcolor{gray!20}
 & & & \multicolumn{1}{c}{$\text{OOD-S}^{\text{CoT}}$} & 3818 & 64.64 & 62.42 & 66.61 & 58.72 & 29.49 & 01:37:42$^\text{\textbf{\textit{1}}}$ \\
 \rowcolor{gray!20}
 \multirow{-6}{*}{\centering Llama-3.2-11B-Vision-Instruct} & \multirow{-6}{*}{\centering ViT-H-14 } & \multirow{-6}{*}{\centering Llama3.1-8B} & \multicolumn{1}{c}{$\text{OOD-H}^{\text{CoT}}$} & 3834 & 61.22 & 61.29 & 61.17 & 61.40 & 22.43 & 01:34:27$^\text{\textbf{\textit{1}}}$ \\

 & & & \multicolumn{1}{c}{ID} & 3556 & 95.78 & 95.69 & 97.88 & 93.59 & 91.65 & 01:16:54$^\text{\textbf{\textit{2}}}$ \\
 & & & \multicolumn{1}{c}{OOD-S} & 3820 & 86.10 & 85.58 & 88.89 & 82.51 & 72.39 & 01:22:25$^\text{\textbf{\textit{2}}}$ \\ 
 & & & \multicolumn{1}{c}{OOD-H} & 3836 & 75.36 & 72.42 & 82.24 & 64.70 & 51.92 & 01:22:52$^\text{\textbf{\textit{2}}}$\\
 & & & \multicolumn{1}{c}{$\text{ID}^{\text{CoT}}$} & 3556 & 95.73 & 95.67 & 96.94 & 94.43 & 91.48 & 01:37:58$^\text{\textbf{\textit{2}}}$ \\
 & & & \multicolumn{1}{c}{$\text{OOD-S}^{\text{CoT}}$}& 3820 & 83.53 & 83.24 & 84.75 & 81.78 & 67.11 & 01:34:26$^\text{\textbf{\textit{2}}}$ \\
\multirow{-6}{*}{\centering Qwen2-VL-7B-Instruct} & \multirow{-6}{*}{\centering CLIP-L-14} & \multirow{-6}{*}{\centering Qwen2-7B} & \multicolumn{1}{c}{$\text{OOD-H}^{\text{CoT}}$} & 3836 & 70.62 & 68.77 & 73.39 & 64.70 & 41.53 & 01:47:12$^\text{\textbf{\textit{2}}}$ \\

\rowcolor{gray!20}
 & & & \multicolumn{1}{c}{ID} & 3556 & 92.82 & 92.28 & 99.87 & 85.76 & 86.52 & 00:44:03$^\text{\textbf{\textit{1}}}$ \\
\rowcolor{gray!20}
 & & & \multicolumn{1}{c}{OOD-S} & 3820 & 86.30 & 84.69 & 96.02 & 75.75 & 74.28 & 00:51:29$^\text{\textbf{\textit{1}}}$ \\ 
 \rowcolor{gray!20}
 & & & \multicolumn{1}{c}{OOD-H} & 3836 & 81.35 & 77.39 & 98.23 & 63.85 & 66.94 & 00:47:35$^\text{\textbf{\textit{1}}}$\\
\rowcolor{gray!20}
 & & & \multicolumn{1}{c}{$\text{ID}^{\text{CoT}}$} & 3556 & 94.15 & 93.78 & 100.00 & 88.29 & 88.91 & 02:01:18$^\text{\textbf{\textit{1}}}$ \\
\rowcolor{gray!20}
 & & & \multicolumn{1}{c}{$\text{OOD-S}^{\text{CoT}}$}& 3820 & 84.65 & 82.56 & 95.59 & 72.66 & 71.39 & 02:12:09$^\text{\textbf{\textit{1}}}$ \\
 \rowcolor{gray!20}
\multirow{-6}{*}{\centering Qwen2.5-VL-7B-Instruct} & \multirow{-6}{*}{\centering ViT(SwiGLU, RMSNorm)} & \multirow{-6}{*}{\centering Qwen2.5 LLM} & \multicolumn{1}{c}{$\text{OOD-H}^{\text{CoT}}$} & 3836 & 81.61 & 77.61 & 99.19 & 63.75 & 67.69 & 02:04:44$^\text{\textbf{\textit{1}}}$ \\

 \hline

& & \centering \textbf{Closed-source Models} & & & & & & & \\
\hline
 & & & \multicolumn{1}{c}{ID} & 3554 & 96.88 & 96.88 & 96.75 & 97.02 & 93.75 & 03:11:49 \\
 & & & \multicolumn{1}{c}{OOD-S} & 3818 & 88.45 & 88.48 & 88.27 & 88.69 & 76.90 & 03:35:00  \\
& & & \multicolumn{1}{c}{OOD-H} & 3834 & 86.78 & 86.96 & 85.79 & 88.16 & 73.58 & 03:38:15 \\
 & & & \multicolumn{1}{c}{$\text{ID}^{\text{CoT}}$} & 3554 & 97.16 & 97.14 & 97.89 & 96.40 & 94.33 & 03:32:09 \\
 & & & \multicolumn{1}{c}{$\text{OOD-S}^{\text{CoT}}$} & 3818 & 86.90 & 85.96 & 92.62 & 80.20 & 74.48 & 03:42:41 \\
\multirow{-6}{*}{\centering Gemini} & \multirow{-6}{*}{\centering -} & \multirow{-6}{*}{\centering -} & \multicolumn{1}{c}{$\text{OOD-H}^{\text{CoT}}$} & 3834 & 91.78 & 91.40 & 95.88 & 87.32 & 83.90 & 03:42:08 \\

\rowcolor{gray!20}
 & & & \multicolumn{1}{c}{ID} & 3552 & 96.54 & 96.59 & 95.09 & 98.14 & 93.12 & 07:12:17  \\
 \rowcolor{gray!20}
 & & & \multicolumn{1}{c}{OOD-S} & 3800 & 89.34 & 89.48 & 88.35 & 90.63 & 78.71 & 07:51:47  \\
 \rowcolor{gray!20}
 & & & \multicolumn{1}{c}{OOD-H} & 3830 & 81.28 & 82.32 & 77.99 & 87.15 & 63.00 & 07:21:43  \\
 \rowcolor{gray!20}
 & & & \multicolumn{1}{c}{$\text{ID}^{\text{CoT}}$} & 3514 & 77.32 & 80.81 & 70.03 & 95.50 & 58.66 & 10:25:24 \\
 \rowcolor{gray!20}
 & & & \multicolumn{1}{c}{$\text{OOD-S}^{\text{CoT}}$} & 3816 & 70.55 & 74.68 & 65.48 & 86.90 & 43.48 & 08:13:24 \\
 \rowcolor{gray!20}
 \multirow{-6}{*}{\centering GPT-4o} & \multirow{-6}{*}{\centering - } & \multirow{-6}{*}{\centering GPT-4 } & \multicolumn{1}{c}{$\text{OOD-H}^{\text{CoT}}$} & 3824 & 69.87 & 73.36 & 65.75 & 82.95 & 41.18 & 09:15:38 \\

 \hline

\bottomrule
\end{tabular}
}

\end{table*}



\textbf{Subdataset BAP performance.} Tab.~\ref{tab: COCO BAP}, ~\ref{tab: LVIS BAP}, ~\ref{tab: nuScenes BAP}, and ~\ref{tab: Cityscapes BAP} show the evaluation results of the BAP metrics on four datasets, namely COCO, LVIS, nuScenes, and Cityscapes, respectively. The BAP metrics progressively increase the complexity of the questions from the perspective of task construction, and consist of Existence, Counting, and Logic Reasoning, with the corresponding performance evaluation metrics of E-Acc, C-Acc, and L-Acc, respectively.

The results of the four datasets show a consistent trend. Under the same question type, as the data distribution migrates from ID to OOD-S to OOD-H, the model performance generally decreases. This phenomenon indicates that the performance of the multimodal large language model degrades significantly when the data distribution deviates from the training distribution, reflecting that its generalization ability in dealing with complex environments is still limited.

On the other hand, while keeping the test data unchanged, when the complexity of questioning is gradually increased from Existence to Counting to Logic Reasoning, the accuracy of the model on the three corresponding metrics (E-Acc, C-Acc, and L-Acc) also shows a decreasing trend, which indicates that the higher the requirement of reasoning ability is, the more difficult it is for the model to answer correctly.

Specifically, Tab.~\ref{tab: COCO BAP} to~\ref{tab: Cityscapes BAP} systematically show the performance of major mainstream VLMs on the four datasets of COCO, LVIS, nuScenes, and Cityscapes. The results are generally consistent, further validating the generality of the two trends mentioned above. This indicates that current VLMs still have significant performance bottlenecks when facing more challenging inference tasks and more complex data distributions.


\begin{table*}[tp]
\caption{Experimental performance of \textbf{10} leading VLMs on \textbf{COCO} for Basic-to-Advanced Progress.}
\label{tab: COCO BAP}
\begin{center}

\resizebox{10.5cm}{!}{
\begin{tabular}{cccccc} 
\toprule 
\hline
\centering \multirow{2}{*}{Model}  & \centering \multirow{2}{*}{Data Type} & \multicolumn{3}{c}{\textbf{COCO BAP} Performance} \\
\cline{3-6}
  & & E-Acc($\%$) & C-Acc($\%$) & L-Acc($\%$) & Time \\
 \hline
\midrule 
 & ID & 88.04 & 61.72 & 37.08 & 00:03:48$^\text{\textbf{\textit{2}}}$\\

 & OOD-S & 79.98 & 47.40 & 33.23 & 00:10:14$^\text{\textbf{\textit{2}}}$\\

\multirow{-3}{*}{\centering LLaVA-NeXT-8B} & OOD-H & 55.85 & 33.83 & 21.10 & 00:09:47$^\text{\textbf{\textit{2}}}$ \\
\rowcolor{gray!20}
 & ID & 91.39 & 58.61 & 44.26 & 00:02:58$^\text{\textbf{\textit{2}}}$\\
\rowcolor{gray!20}
 & OOD-S & 80.30 & 36.26 & 27.81 & 00:06:47$^\text{\textbf{\textit{2}}}$\\
\rowcolor{gray!20}
\multirow{-3}{*}{\centering DeepSeek-VL-7B-Chat} & OOD-H & 60.89 & 33.26 & 23.17 & 00:06:16$^\text{\textbf{\textit{2}}}$ \\

 & ID & 85.85 & 55.40 & 37.65 & 00:08:53$^\text{\textbf{\textit{1}}}$\\

 & OOD-S & 73.89 & 41.28 & 29.14 & 00:23:00$^\text{\textbf{\textit{1}}}$\\

\multirow{-3}{*}{\centering DeepSeek-VL2-Small} & OOD-H & 49.77 & 34.86 & 23.74 & 00:22:08$^\text{\textbf{\textit{1}}}$ \\
\rowcolor{gray!20}
 & ID & 88.04 & 69.14 & 57.18 & 00:05:34$^\text{\textbf{\textit{2}}}$\\
\rowcolor{gray!20}
 & OOD-S & 79.87 & 53.57 & 41.45 & 00:12:32$^\text{\textbf{\textit{2}}}$\\
\rowcolor{gray!20}
\multirow{-3}{*}{\centering InternVL2-8B} & OOD-H & 59.63 & 49.43 & 36.35 & 00:11:26$^\text{\textbf{\textit{2}}}$ \\

 & ID & 94.02 & 67.94 & 53.83 & 00:04:35$^\text{\textbf{\textit{2}}}$\\

 & OOD-S & 85.93 & 59.20 & 47.62 & 00:11:04$^\text{\textbf{\textit{2}}}$\\

\multirow{-3}{*}{\centering InternVL2.5-8B} & OOD-H & 70.87 & 49.31 & 38.76 & 00:10:10$^\text{\textbf{\textit{2}}}$ \\
\rowcolor{gray!20}
 & ID & 96.65 & 66.99 & 51.20 & 00:04:26$^\text{\textbf{\textit{2}}}$\\
\rowcolor{gray!20}
 & OOD-S & 86.90 & 56.17 & 35.61 & 00:10:17$^\text{\textbf{\textit{2}}}$\\
\rowcolor{gray!20}
\multirow{-3}{*}{\centering Llama-3.2-11B-Vision-Instruct} & OOD-H & 86.93 & 48.17 & 31.08 & 00:09:59$^\text{\textbf{\textit{2}}}$ \\

 & ID & 91.87 & 65.55 & 48.56 & 00:02:13$^\text{\textbf{\textit{2}}}$\\

 & OOD-S & 81.82 & 62.12 & 48.16 & 00:05:09$^\text{\textbf{\textit{2}}}$\\

\multirow{-3}{*}{\centering Qwen2-VL-7B-Instruct} & OOD-H & 66.06 & 45.53 & 33.60 & 00:05:39$^\text{\textbf{\textit{2}}}$ \\
\rowcolor{gray!20}
 & ID & 89.21 & 66.19 & 54.44 & 00:06:01$^\text{\textbf{\textit{1}}}$\\
\rowcolor{gray!20}
 & OOD-S & 78.11 & 59.05 & 45.50 & 00:15:26$^\text{\textbf{\textit{1}}}$\\
\rowcolor{gray!20}
\multirow{-3}{*}{\centering Qwen2.5-VL-7B-Instruct} & OOD-H & 58.83 & 47.36 & 35.78 & 00:15:06$^\text{\textbf{\textit{1}}}$ \\

\hline

 & ID & 88.49 & 75.78 & 42.21 & 02:57:38\\
 & OOD-S & 81.47 & 69.45 & 48.43 & 00:59:56\\
\multirow{-3}{*}{\centering Gemini} & OOD-H & 66.97 & 52.29 & 34.40 & 03:54:16 \\
\rowcolor{gray!20}

 & ID & 90.41 & 66.67 & 36.93 & 00:45:11\\
\rowcolor{gray!20}
 & OOD-S & 85.70 & 58.07 & 42.90 & 01:44:58\\
\rowcolor{gray!20}
\multirow{-3}{*}{\centering gpt-4o} & OOD-H & 74.77 & 48.17 & 33.72 & 01:45:12 \\

\bottomrule

\end{tabular}}
\end{center}
\end{table*}

\begin{table*}[tp]
\caption{Experimental performance of \textbf{10} leading VLMs on \textbf{LVIS} for Basic-to-Advanced Progress.}
\label{tab: LVIS BAP}
\begin{center}

\resizebox{10.5cm}{!}{
\begin{tabular}{cccccc} 
\toprule 
\hline
\centering \multirow{2}{*}{Model}  & \centering \multirow{2}{*}{Data Type} & \multicolumn{3}{c}{\textbf{LVIS BAP} Performance} \\
\cline{3-6}
  & & E-Acc($\%$) & C-Acc($\%$) & L-Acc($\%$) & Time \\
 \hline
\midrule 
 & ID & 81.27 & 52.07 & 29.84 & 00:07:14$^\text{\textbf{\textit{2}}}$\\

 & OOD-S & 49.04 & 35.01 & 17.39 & 00:10:15$^\text{\textbf{\textit{2}}}$\\

\multirow{-3}{*}{\centering LLaVA-NeXT-8B} & OOD-H & 31.25 & 27.73 & 14.06 & 00:02:45$^\text{\textbf{\textit{2}}}$ \\
\rowcolor{gray!20}
 & ID & 90.44 & 55.04 & 39.79 & 00:05:38$^\text{\textbf{\textit{2}}}$\\
\rowcolor{gray!20}
 & OOD-S & 53.36 & 22.66 & 16.19 & 00:06:24$^\text{\textbf{\textit{2}}}$\\
\rowcolor{gray!20}
\multirow{-3}{*}{\centering DeepSeek-VL-7B-Chat} & OOD-H & 50.78 & 23.05 & 13.67 & 00:01:56$^\text{\textbf{\textit{2}}}$ \\

 & ID & 77.23 & 43.86 & 28.07 & 00:23:36$^\text{\textbf{\textit{1}}}$\\

 & OOD-S & 51.14 & 34.21 & 19.93 & 00:25:16$^\text{\textbf{\textit{1}}}$\\

\multirow{-3}{*}{\centering DeepSeek-VL2-Small} & OOD-H & 29.41 & 26.67 & 12.94 & 00:08:02$^\text{\textbf{\textit{1}}}$ \\
\rowcolor{gray!20}
 & ID & 85.14 & 65.37 & 48.32 & 00:11:28$^\text{\textbf{\textit{2}}}$\\
\rowcolor{gray!20}
 & OOD-S & 55.16 & 35.97 & 24.82 & 00:12:22$^\text{\textbf{\textit{2}}}$\\
\rowcolor{gray!20}
\multirow{-3}{*}{\centering InternVL2-8B} & OOD-H & 42.19 & 37.89 & 28.12 & 00:03:34$^\text{\textbf{\textit{2}}}$ \\

 & ID & 90.18 & 61.50 & 46.51 & 00:08:46$^\text{\textbf{\textit{2}}}$\\

 & OOD-S & 56.12 & 38.13 & 25.06 & 00:10:41$^\text{\textbf{\textit{2}}}$\\

\multirow{-3}{*}{\centering InternVL2.5-8B} & OOD-H & 48.83 & 36.72 & 21.48 & 00:03:12$^\text{\textbf{\textit{2}}}$ \\
\rowcolor{gray!20}
 & ID & 96.12 & 62.02 & 41.99 & 00:08:27$^\text{\textbf{\textit{2}}}$\\
\rowcolor{gray!20}
 & OOD-S & 59.11 & 37.17 & 25.78 & 00:09:29$^\text{\textbf{\textit{2}}}$\\
\rowcolor{gray!20}
\multirow{-3}{*}{\centering Llama-3.2-11B-Vision-Instruct} & OOD-H & 62.50 & 32.42 & 21.88 & 00:02:56$^\text{\textbf{\textit{2}}}$ \\

 & ID & 91.60 & 63.95 & 45.48 & 00:04:38$^\text{\textbf{\textit{2}}}$\\

 & OOD-S & 57.07 & 44.36 & 28.78 & 00:05:36$^\text{\textbf{\textit{2}}}$\\

\multirow{-3}{*}{\centering Qwen2-VL-7B-Instruct} & OOD-H & 51.95 & 31.64 & 23.44 & 00:01:35$^\text{\textbf{\textit{2}}}$ \\
\rowcolor{gray!20}
 & ID & 88.75 & 59.38 & 43.73 & 00:11:42$^\text{\textbf{\textit{1}}}$\\
\rowcolor{gray!20}
 & OOD-S & 60.98 & 39.98 & 26.05 & 00:15:16$^\text{\textbf{\textit{1}}}$\\
\rowcolor{gray!20}
\multirow{-3}{*}{\centering Qwen2.5-VL-7B-Instruct} & OOD-H & 54.51 & 32.16 & 19.61 & 00:04:16$^\text{\textbf{\textit{1}}}$ \\

\hline

 & ID & 89.65 & 70.25 & 37.39 & 00:49:55\\

 & OOD-S & 57.21 & 50.72 & 25.72 & 04:56:20\\

\multirow{-3}{*}{\centering Gemini} & OOD-H & 54.12 & 41.57 & 23.53 & 00:17:33 \\
\rowcolor{gray!20}
 & ID & 94.44 & 63.39 & 39.07 & 01:29:08\\
\rowcolor{gray!20}
 & OOD-S & 63.03 & 41.54 & 21.37 & 01:39:38\\
\rowcolor{gray!20}
\multirow{-3}{*}{\centering gpt-4o} & OOD-H & 58.43 & 35.29 & 16.08 & 00:29:38 \\

\bottomrule

\end{tabular}}
\end{center}
\end{table*}

\begin{table*}[tp]
\caption{Experimental performance of \textbf{10} leading VLMs on \textbf{nuScenes} for Basic-to-Advanced Progress.}
\label{tab: nuScenes BAP}
\begin{center}

\resizebox{10.5cm}{!}{
\begin{tabular}{cccccc} 
\toprule 
\hline
\centering \multirow{2}{*}{Model}  & \centering \multirow{2}{*}{Data Type} & \multicolumn{3}{c}{\textbf{nuScenes BAP} Performance} \\
\cline{3-6}
  & & E-Acc($\%$) & C-Acc($\%$) & L-Acc($\%$) & Time \\
 \hline
\midrule 
 & ID & 68.31 & 21.38 & 15.54 & 00:08:03$^\text{\textbf{\textit{2}}}$\\

 & OOD-S & 65.15 & 40.88 & 28.65 & 00:07:10$^\text{\textbf{\textit{2}}}$\\

\multirow{-3}{*}{\centering LLaVA-NeXT-8B} & OOD-H & 35.10 & 20.65 & 13.27 & 00:08:46$^\text{\textbf{\textit{2}}}$ \\
\rowcolor{gray!20}
 & ID & 68.46 & 30.77 & 13.08 & 00:05:05$^\text{\textbf{\textit{2}}}$\\
\rowcolor{gray!20}
 & OOD-S & 59.85 & 34.67 & 19.89 & 00:04:29$^\text{\textbf{\textit{2}}}$\\
\rowcolor{gray!20}
\multirow{-3}{*}{\centering DeepSeek-VL-7B-Chat} & OOD-H & 27.14 & 22.57 & 15.34 & 00:05:23$^\text{\textbf{\textit{2}}}$ \\

 & ID & 56.39 & 5.39 & 3.54 & 00:22:42$^\text{\textbf{\textit{1}}}$\\

 & OOD-S & 57.04 & 43.88 & 22.30 & 00:25:28$^\text{\textbf{\textit{1}}}$\\

\multirow{-3}{*}{\centering DeepSeek-VL2-Small} & OOD-H & 20.21 & 22.71 & 10.18 & 00:28:18$^\text{\textbf{\textit{1}}}$ \\
\rowcolor{gray!20}
 & ID & 82.15 & 38.92 & 25.85 & 00:10:03$^\text{\textbf{\textit{2}}}$\\
\rowcolor{gray!20}
 & OOD-S & 69.89 & 41.42 & 25.73 & 00:07:42$^\text{\textbf{\textit{2}}}$\\
\rowcolor{gray!20}
\multirow{-3}{*}{\centering InternVL2-8B} & OOD-H & 44.99 & 41.30 & 28.76 & 00:09:43$^\text{\textbf{\textit{2}}}$ \\

 & ID & 85.69 & 38.92 & 31.54 & 00:05:56$^\text{\textbf{\textit{2}}}$\\

 & OOD-S & 75.73 & 45.26 & 34.85 & 00:05:11$^\text{\textbf{\textit{2}}}$\\

\multirow{-3}{*}{\centering InternVL2.5-8B} & OOD-H & 46.02 & 35.10 & 24.78 & 00:06:36$^\text{\textbf{\textit{2}}}$ \\
\rowcolor{gray!20}
 & ID & 87.38 & 37.69 & 19.54 & 00:08:15$^\text{\textbf{\textit{2}}}$\\
\rowcolor{gray!20}
 & OOD-S & 77.37 & 51.46 & 23.54 & 00:07:08$^\text{\textbf{\textit{2}}}$\\
\rowcolor{gray!20}
\multirow{-3}{*}{\centering Llama-3.2-11B-Vision-Instruct} & OOD-H & 58.11 & 28.47 & 17.11 & 00:08:38$^\text{\textbf{\textit{2}}}$ \\

 & ID & 88.00 & 34.00 & 24.62 & 00:08:47$^\text{\textbf{\textit{2}}}$\\

 & OOD-S & 75.73 & 50.36 & 40.51 & 00:07:29$^\text{\textbf{\textit{2}}}$\\

\multirow{-3}{*}{\centering Qwen2-VL-7B-Instruct} & OOD-H & 47.49 & 36.28 & 25.07 & 00:09:10$^\text{\textbf{\textit{2}}}$ \\
\rowcolor{gray!20}
 & ID & 81.51 & 32.05 & 23.42 & 00:22:13$^\text{\textbf{\textit{1}}}$\\
\rowcolor{gray!20}
 & OOD-S & 72.39 & 44.79 & 35.47 & 00:18:53$^\text{\textbf{\textit{1}}}$\\
\rowcolor{gray!20}
\multirow{-3}{*}{\centering Qwen2.5-VL-7B-Instruct} & OOD-H & 44.69 & 35.40 & 25.22 & 00:22:51$^\text{\textbf{\textit{1}}}$ \\

\hline

 & ID & 86.44 & 47.15 & 35.75 & 00:55:20\\

 & OOD-S & 79.71 & 53.38 & 40.95 & 00:43:04\\

\multirow{-3}{*}{\centering Gemini} & OOD-H & 62.09 & 46.31 & 29.94 & 00:55:34 \\
\rowcolor{gray!20}
 & ID & 92.91 & 36.21 & 29.12 & 01:53:19\\
\rowcolor{gray!20}
 & OOD-S & 82.97 & 47.07 & 36.08 & 01:34:33\\
\rowcolor{gray!20}
\multirow{-3}{*}{\centering gpt-4o} & OOD-H & 68.83 & 38.26 & 26.88 & 01:55:43 \\

\bottomrule

\end{tabular}}
\end{center}
\end{table*}

\begin{table*}[tp]
\caption{Experimental performance of \textbf{10} leading VLMs on \textbf{Cityscapes} for Basic-to-Advanced Progress.}
\label{tab: Cityscapes BAP}
\begin{center}

\resizebox{10.5cm}{!}{
\begin{tabular}{cccccc} 
\toprule 
\hline
\centering \multirow{2}{*}{Model}  & \centering \multirow{2}{*}{Data Type} & \multicolumn{3}{c}{\textbf{Cityscapes BAP} Performance} \\
\cline{3-6}
  & & E-Acc($\%$) & C-Acc($\%$) & L-Acc($\%$) & Time \\
 \hline
\midrule 
& ID & 87.37 & 28.44 & 12.73 & 00:11:48$^\text{\textbf{\textit{2}}}$\\

 & OOD-S & 69.53 & 17.17 & 10.75 & 00:13:41$^\text{\textbf{\textit{2}}}$\\

\multirow{-3}{*}{\centering LLaVA-NeXT-8B} & OOD-H & 43.62 & 10.86 & 7.05 & 00:12:38$^\text{\textbf{\textit{2}}}$ \\
\rowcolor{gray!20}
 & ID & 87.99 & 22.38 & 11.19 & 00:08:56$^\text{\textbf{\textit{2}}}$\\
\rowcolor{gray!20}
 & OOD-S & 70.19 & 15.19 & 8.21 & 00:09:51$^\text{\textbf{\textit{2}}}$\\
\rowcolor{gray!20}
\multirow{-3}{*}{\centering DeepSeek-VL-7B-Chat} & OOD-H & 45.33 & 12.95 & 6.86 & 00:09:43$^\text{\textbf{\textit{2}}}$ \\

 & ID & 68.76 & 8.12 & 2.67 & 00:50:22$^\text{\textbf{\textit{1}}}$\\

 & OOD-S & 67.64 & 12.08 & 4.43 & 01:07:38$^\text{\textbf{\textit{1}}}$\\

\multirow{-3}{*}{\centering DeepSeek-VL2-Small} & OOD-H & 41.09 & 5.15 & 3.43 & 00:45:52$^\text{\textbf{\textit{1}}}$ \\
\rowcolor{gray!20}
 & ID & 92.92 & 40.04 & 16.22 & 00:15:03$^\text{\textbf{\textit{2}}}$\\
\rowcolor{gray!20}
 & OOD-S & 83.30 & 31.98 & 15.57 & 00:27:59$^\text{\textbf{\textit{2}}}$\\
\rowcolor{gray!20}
\multirow{-3}{*}{\centering InternVL2-8B} & OOD-H & 65.43 & 32.38 & 21.71 & 00:18:06$^\text{\textbf{\textit{2}}}$ \\

 & ID & 94.87 & 27.93 & 12.63 & 00:10:02$^\text{\textbf{\textit{2}}}$\\

 & OOD-S & 85.94 & 17.83 & 12.92 & 00:11:15$^\text{\textbf{\textit{2}}}$\\

\multirow{-3}{*}{\centering InternVL2.5-8B} & OOD-H & 67.14 & 23.71 & 18.19 & 00:10:22$^\text{\textbf{\textit{2}}}$ \\
\rowcolor{gray!20}
 & ID & 97.23 & 33.47 & 13.86 & 00:13:29$^\text{\textbf{\textit{2}}}$\\
\rowcolor{gray!20}
 & OOD-S & 85.38 & 18.87 & 7.92 & 00:15:05$^\text{\textbf{\textit{2}}}$\\
\rowcolor{gray!20}
\multirow{-3}{*}{\centering Llama-3.2-11B-Vision-Instruct} & OOD-H & 77.24 & 19.62 & 8.29 & 00:15:06$^\text{\textbf{\textit{2}}}$ \\

 & ID & 94.25 & 31.72 & 13.35 & 00:33:15$^\text{\textbf{\textit{2}}}$\\

 & OOD-S & 83.77 & 19.43 & 11.98 & 00:35:41$^\text{\textbf{\textit{2}}}$\\

\multirow{-3}{*}{\centering Qwen2-VL-7B-Instruct} & OOD-H & 64.38 & 22.00 & 15.14 & 00:35:37$^\text{\textbf{\textit{2}}}$ \\
\rowcolor{gray!20}
 & ID & 86.43 & 22.40 & 10.69 & 00:50:03$^\text{\textbf{\textit{1}}}$\\
\rowcolor{gray!20}
 & OOD-S & 79.53 & 18.58 & 10.85 & 00:57:08$^\text{\textbf{\textit{1}}}$\\
\rowcolor{gray!20}
\multirow{-3}{*}{\centering Qwen2.5-VL-7B-Instruct} & OOD-H & 62.73 & 21.54 & 15.54 & 00:54:25$^\text{\textbf{\textit{1}}}$ \\

\hline

 & ID & 97.43 & 48.92 & 31.14 & 02:30:39\\

 & OOD-S & 89.91 & 37.74 & 24.72 & 03:00:34\\

\multirow{-3}{*}{\centering Gemini} & OOD-H & 87.70 & 32.98 & 22.12 & 03:46:30 \\
\rowcolor{gray!20}
 & ID & 98.36 & 36.38 & 22.82 & 05:50:51\\
\rowcolor{gray!20}
 & OOD-S & 90.85 & 23.96 & 15.85 & 06:37:16\\
\rowcolor{gray!20}
\multirow{-3}{*}{\centering gpt-4o} & OOD-H & 86.94 & 24.59 & 16.97 & 06:31:57 \\

\bottomrule

\end{tabular}}
\end{center}
\end{table*}



\section{Effect of Model Scale on OODBench Performance}
\label{appendix: model scale effect}
To further analyze the performance of models of different scales on OODBench under the same architecture, we conducted a systematic comparison of different parameter scales within the same model family. Specifically, we selected the representative Qwen2-VL series. We performed experiments on the complete sets and subsets of the four domains covered by OODBench (COCO, LVIS, nuScenes, Cityscapes) for its 2B and 7B versions. The results are shown in Table~\ref{tab: model scale effect}: The 2B model achieved accuracy scores of 80.6, 64.0, and 81.2 on the OOD-H metrics for COCO, nuScenes, and Cityscapes, respectively, all surpassing the 7B model's scores of 77.7, 57.6, and 75.4. However, on the LVIS dataset, the 7B model slightly outperformed the 2B model. These results reveal a distinct mixed trend, \textbf{indicating that model size does not maintain a monotonically increasing relationship with OOD robustness. Increasing parameter size does not consistently enhance model performance under semantic OOD conditions.}

This phenomenon is further validated in larger-scale models. For instance, as shown in Tables~\ref{tab:Main_Results}, ~\ref{tab: coco}, ~\ref{tab: lvis}, ~\ref{tab: nuscenes}, and ~\ref{tab: cityscapes}, models like GPT-4o and Gemini, which have significantly more parameters than 7B, do not systematically outperform smaller-scale models on OODBench. Overall, expanding model scale cannot fundamentally resolve semantic OOD issues. The root cause lies in the fact that when test samples deviate from the support domain of training data in the joint image-text distribution, their OOD characteristics remain independent of model scale and thus do not naturally disappear with increased capacity.

\begin{table*}[ht]
\centering
\caption{Performance of different model scales on OODBench. The table shows results for the 2B and 7B variants of the Qwen2-VL series across various data domains on OODBench. It can be observed that there is no monotonic relationship between model scale and OOD robustness; increasing parameter scale does not consistently improve model performance under semantic OOD conditions. Where, OOD-S' and OOD-H' denote the initial benchmark datasets.}
\label{tab: model scale effect}
\resizebox{12cm}{!}{
\begin{tabular}{>{\centering\arraybackslash}m{3.2cm} cccccccc}
\toprule
\hline
\multirow{2}{*}{Model} & \multirow{2}{*}{Data Type} &
\multicolumn{6}{c}{\textbf{OODBench} Performance} \\
\cline{3-8}
 & & Num & Accuracy(\%) & F1(\%) & Precision(\%) & Recall(\%) & MCC(\%) \\
\midrule
Random Chance & - & - & 50.00 & 50.00 & 50.00 & 50.00 & 0.00 \\
\midrule

\multicolumn{8}{c}{\textbf{COCO}} \\
\midrule
\multirow{4}{*}{Qwen2-VL-2B}
    & OOD-S   & 6000  & 78.78 & 74.62 & 92.85 & 62.37 & 60.95 \\
    & OOD-H   & 6000  & 80.63 & 76.53 & 97.13 & 63.13 & 65.40 \\
    & OOD-S'  & 7972  & 78.84 & 74.61 & 93.23 & 62.19 & 61.17 \\
    & OOD-H'  & 10160 & 81.81 & 78.13 & 97.95 & 64.98 & 67.56 \\

\hline
    \rowcolor{gray!20}
    & OOD-S   & 6000  & 81.57 & 79.98 & 87.52 & 73.63 & 63.94 \\
    \rowcolor{gray!20}
    & OOD-H   & 6000  & 77.67 & 75.48 & 83.66 & 68.77 & 56.23 \\
    \rowcolor{gray!20}
    & OOD-S'  & 7972  & 81.18 & 79.49 & 87.35 & 72.93 & 63.24 \\
    \rowcolor{gray!20}
   \multirow{-4}{*}{Qwen2-VL-7B} & OOD-H'  & 10160 & 78.74 & 76.78 & 84.58 & 70.33 & 58.32 \\
\midrule

\multicolumn{8}{c}{\textbf{LVIS}} \\
\midrule
\multirow{3}{*}{Qwen2-VL-2B}
    & OOD-S   & 6000  & 48.23 & 34.91 & 47.01 & 27.77 & -3.87 \\
    & OOD-H   & 2436  & 41.67 & 32.69 & 38.63 & 28.33 & -17.29 \\
    & OOD-S'  & 13284 & 47.70 & 35.03 & 46.24 & 28.19 & -4.99 \\
\hline
    \rowcolor{gray!20}
    & OOD-S   & 6000  & 55.60 & 53.15 & 56.25 & 50.37 & 11.26 \\
    \rowcolor{gray!20}
    & OOD-H   & 2436  & 49.14 & 50.50 & 49.18 & 51.89 & -1.73 \\
    \rowcolor{gray!20}
    \multirow{-3}{*}{Qwen2-VL-7B} & OOD-S'  & 13284 & 54.91 & 52.80 & 55.39 & 50.45 & 9.86 \\
\midrule

\multicolumn{8}{c}{\textbf{nuScenes}} \\
\midrule
\multirow{4}{*}{Qwen2-VL-2B}
    & OOD-S   & 6000  & 68.00 & 57.62 & 85.29 & 43.50 & 41.30 \\
    & OOD-H   & 6000  & 64.00 & 50.21 & 81.39 & 36.30 & 33.63 \\
    & OOD-S'  & 25800 & 67.30 & 56.76 & 83.77 & 42.92 & 39.64 \\
    & OOD-H'  & 9884  & 66.04 & 52.68 & 82.76 & 37.82 & 38.55 \\
\hline
    \rowcolor{gray!20}
    & OOD-S   & 6000  & 73.53 & 69.84 & 81.16 & 61.30 & 48.54 \\
    \rowcolor{gray!20}
    & OOD-H   & 6000  & 57.55 & 50.05 & 70.99 & 42.53 & 15.83 \\
    \rowcolor{gray!20}
    & OOD-S'  & 25800 & 73.12 & 69.32 & 80.75 & 60.72 & 47.74 \\
    \rowcolor{gray!20}
    \multirow{-4}{*}{Qwen2-VL-7B} & OOD-H'  & 9884  & 59.02 & 51.87 & 62.85 & 44.15 & 18.91 \\
\midrule

\multicolumn{8}{c}{\textbf{CityScapes}} \\
\midrule
\multirow{2}{*}{Qwen2-VL-2B}
    & OOD-S  & 3820 & 88.85 & 88.05 & 94.81 & 82.20 & 78.39 \\
    & OOD-H  & 3836 & 81.20 & 77.28 & 97.69 & 63.92 & 66.51 \\
\hline
    \rowcolor{gray!20}
    & OOD-S  & 3820 & 86.10 & 85.58 & 88.89 & 82.51 & 72.39 \\
    \rowcolor{gray!20}
    \multirow{-2}{*}{Qwen2-VL-7B} & OOD-H  & 3836 & 75.36 & 72.42 & 82.24 & 64.70 & 51.92 \\
\hline
\bottomrule
\end{tabular}}
\end{table*}


\section{CLIP and BLIP2 visualization distribution logits on coco val}
\label{appendix: CLIP_BLIP2 logits visualization}

\textbf{Why is OODBench Challenging?} The appearance of OOD data is usually caused by semantic shift or covariate shift, which means that the generation mechanism of the test data has changed from the training data. Although a model performs well on the training set, it may show degraded or unstable performance when faced with a change in the data distribution. Models usually lack sufficient robustness to such changes, especially when dealing with changes that are not supported by explicit labels or historical data. We divide the OODBench data containing covariate shift through the scoring model, which has shifted its data distribution despite sharing the same label space as the training data. Deep learning models often give high-confidence predictions or even make very confident false predictions when dealing with these data. The fundamental reason why OODBench poses a challenge to models is that it does not follow the distributional assumptions of the training data.
In contrast, OOD data from the real world is highly uncertain and diverse. When confronted with these data, models typically lack the necessary knowledge and ability to make accurate predictions, which can lead to misclassification or high-confidence false predictions. We shows a visualization of the distribution of logits predicted by VLMs when confronted with real-world data.

\begin{figure*}[ht]
    \centering
    \includegraphics[width=0.5\linewidth]{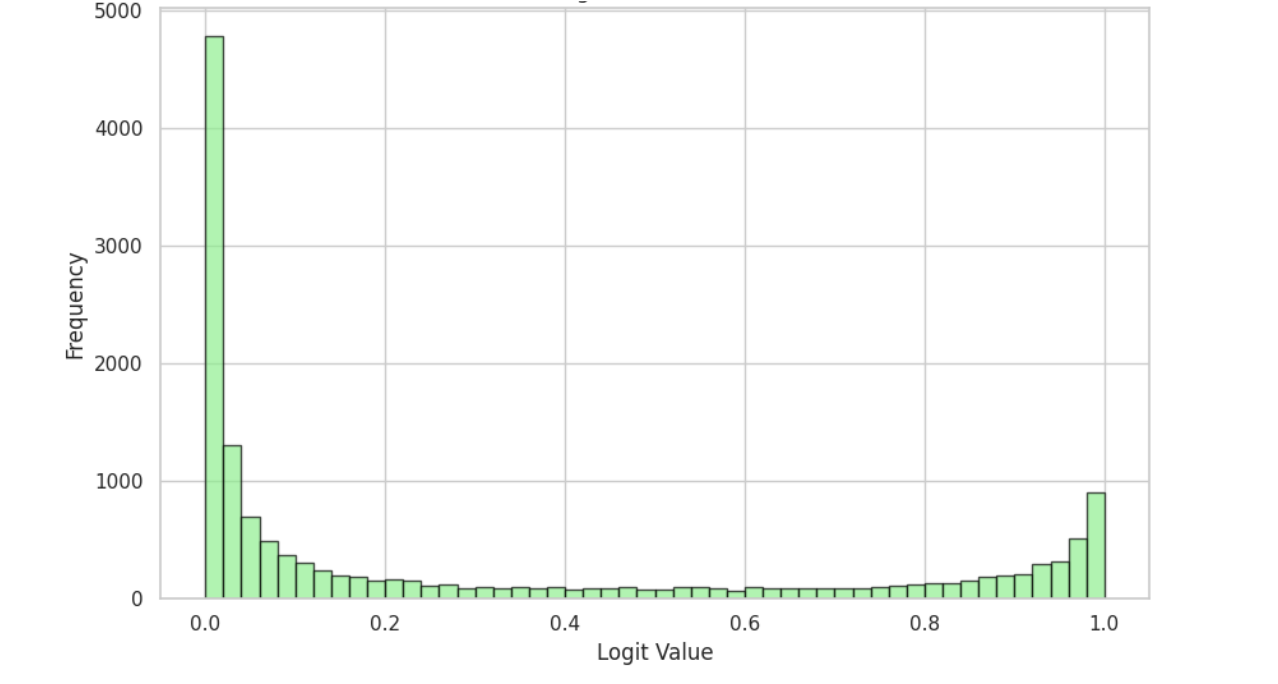}
    \includegraphics[width=0.5\linewidth]{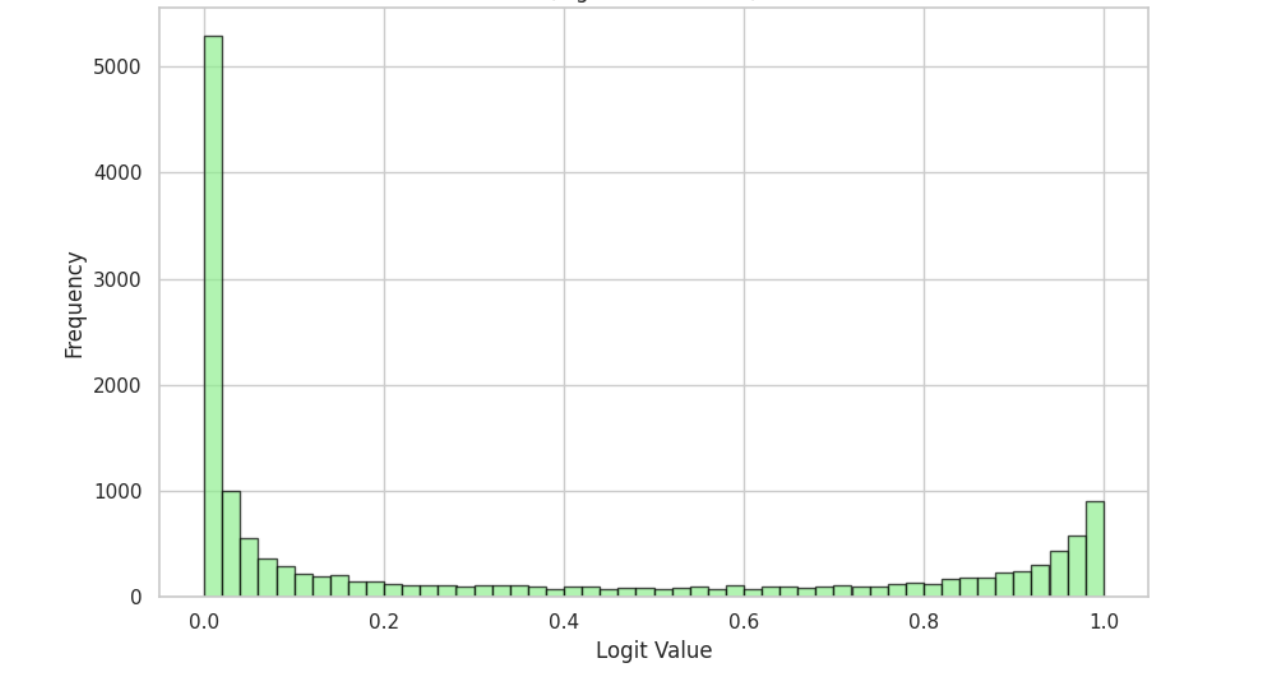}
    \caption{Logistically visualized distributions of CLIP (top image) and BLIP2 (bottom image) on coco val.}
    \label{fig: CLIP&BLIP2 total logits}
\end{figure*}

\begin{figure*}[ht]
    \centering
        \includegraphics[width=0.8\linewidth]{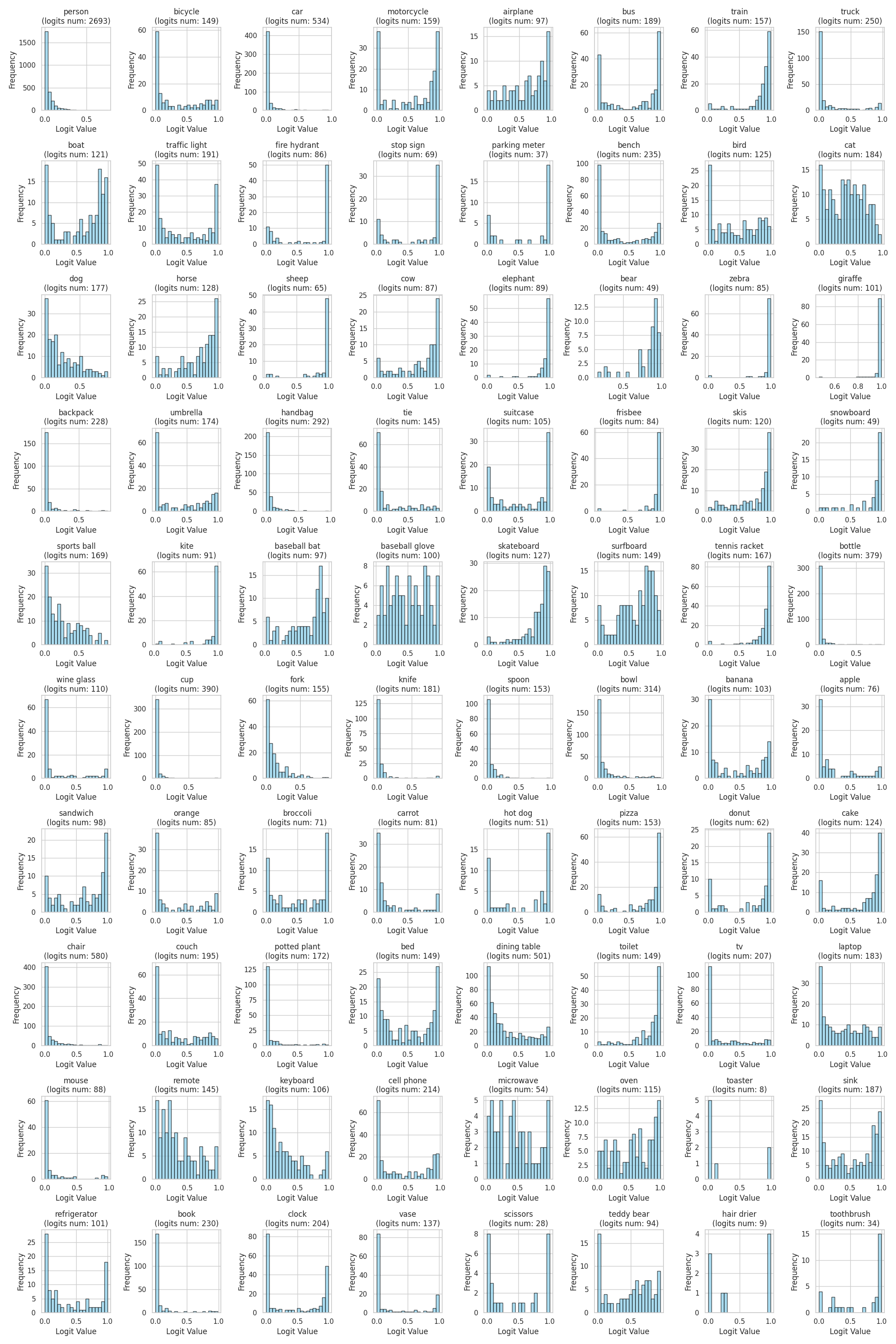}
    \caption{CLIP visualization distribution of 80 category logits on coco val.}
    \label{fig: CLIP category logits}
\end{figure*}

\begin{figure*}[ht]
    \centering
        \includegraphics[width=0.8\linewidth]{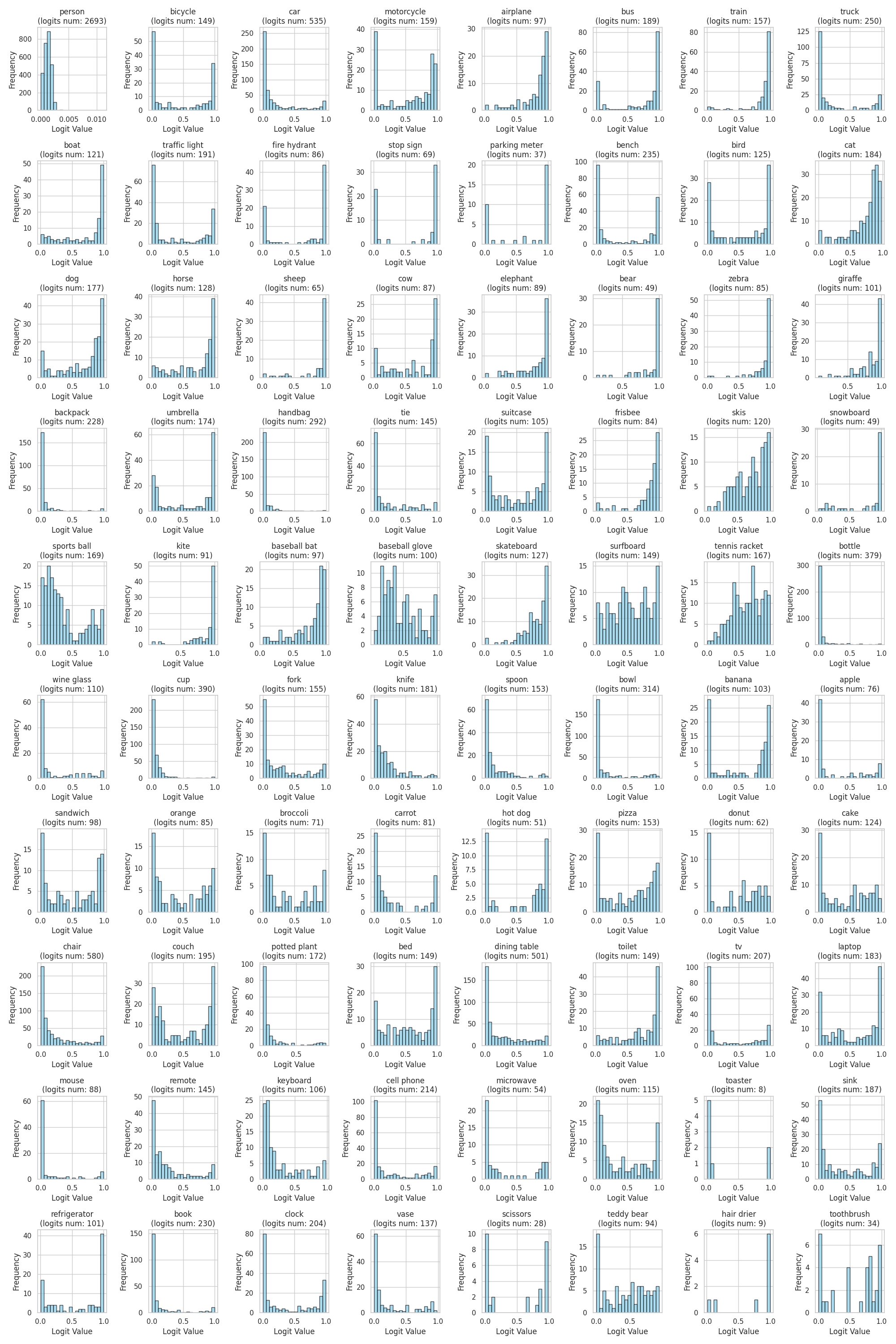}
    \caption{BLIP2 visualization distribution of 80 category logits on coco val.}
    \label{fig: BLIP2 category logits}
\end{figure*}

Figures~\ref{fig: CLIP category logits} and~\ref{fig: BLIP2 category logits} show the results of CLIP and BLIP2 for computing logits on the COCO validation set for 80 labeled categories. Figure~\ref{fig: CLIP&BLIP2 total logits}, on the other hand, summarises the logit results of CLIP and BLIP2 on the COCO validation set for 80 categories. As shown in Figure~\ref{fig: CLIP&BLIP2 total logits}, the logits distribution shows a double peak, appearing on both sides of the distribution, which indicates that the CLIP and BLIP2 models have high confidence in the instance category. However, the single peak on the far left side of Figure~\ref{fig: CLIP&BLIP2 total logits} indicates that both models also have high confidence in determining instance-category mismatches, resulting in high-confidence false predictions.

\section{OODBench Data and Case Demonstration}
\label{appendix: Case Demostration}

\begin{figure*}[ht]
    \centering
        \includegraphics[width=0.65\linewidth]{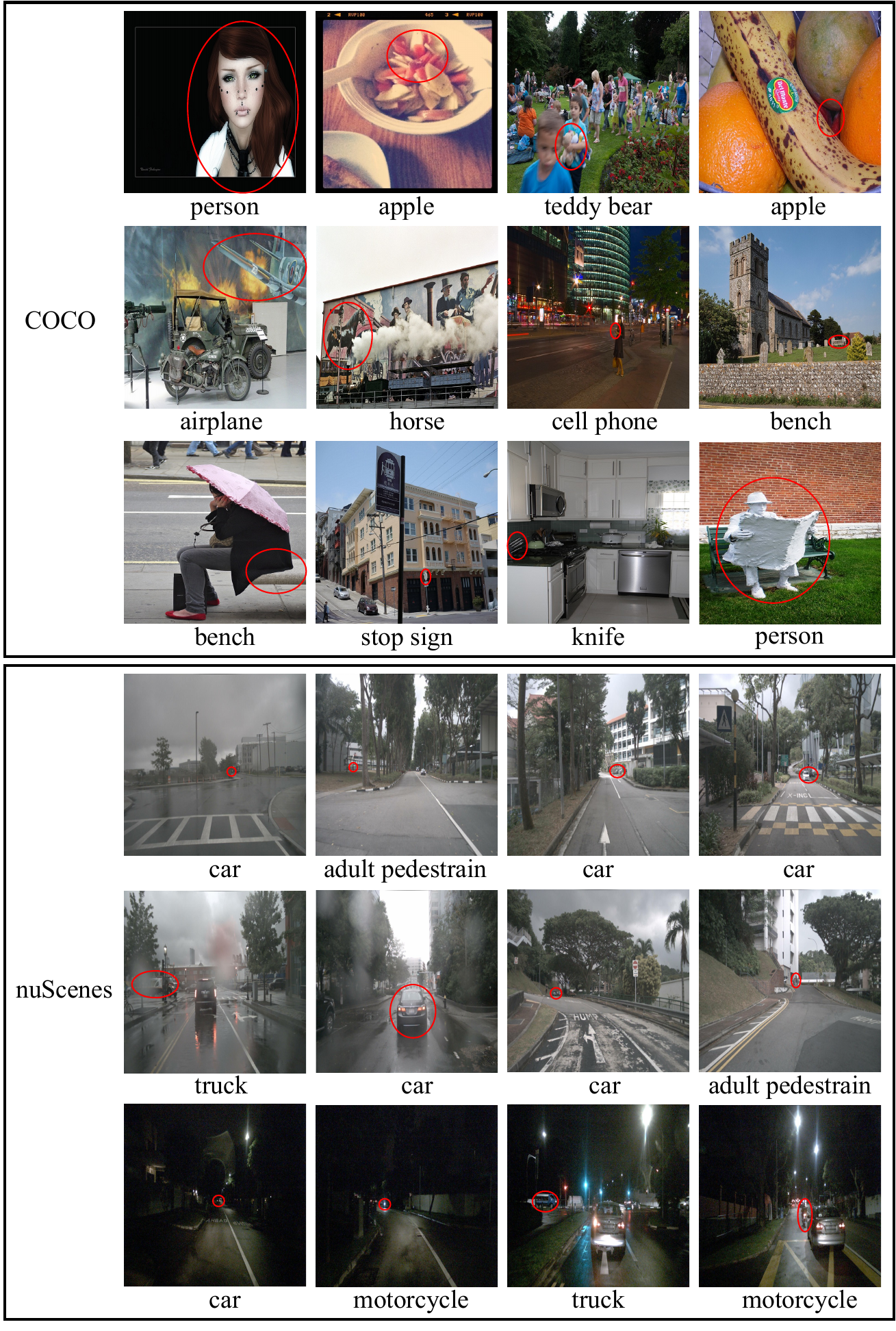}
    \caption{Sample OOD instances from the COCO and nuScenes datasets are divided by OODBench. These examples illustrate diverse out-of-distribution categories across different scenarios and environments.}
    \label{fig: coco_nuscenes_example}
\end{figure*}

\begin{figure*}[ht]
    \centering
        \includegraphics[width=0.65\linewidth]{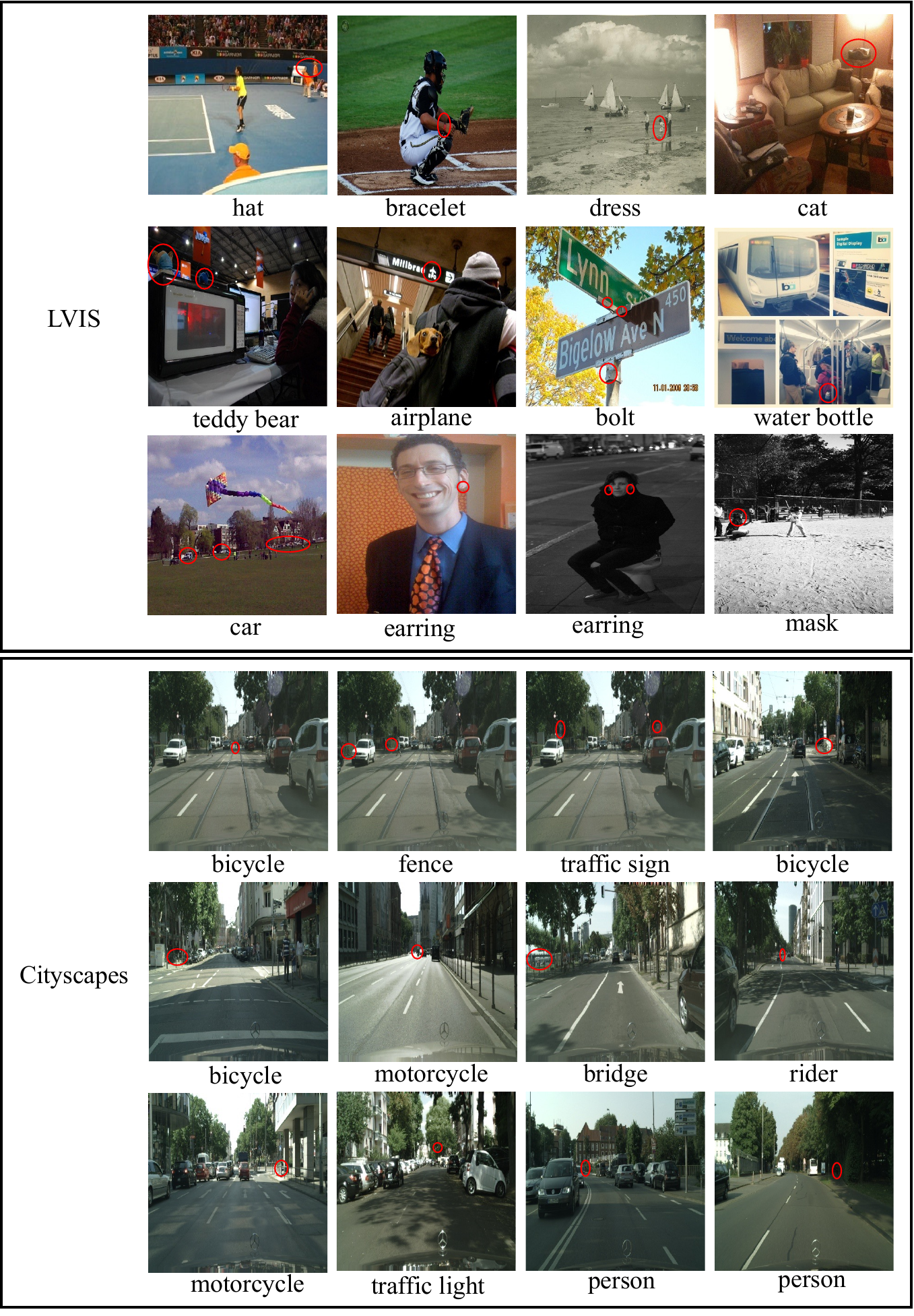}
    \caption{Sample OOD instances divided from the LVIS and Cityscapes datasets by OODBench. The samples highlight distributional shifts in object-centric.}
    \label{fig: lvis_cityscapes_example}
\end{figure*}

In this section, we show some of the data examples constructed by OODBench, covering OOD scenarios divided from datasets such as COCO, LVIS, nuScenes, and Cityscapes. Figures~\ref{fig: coco_nuscenes_example} and~\ref{fig: lvis_cityscapes_example} show the associated visualisations. Through these examples, we aim to help readers visualise the characteristics of different types of out-of-distribution data and the challenges they pose under different tasks and perceptual conditions faced by large multimodal models.

\begin{figure*}[!htbp]
    \centering
        \includegraphics[width=0.8\linewidth]{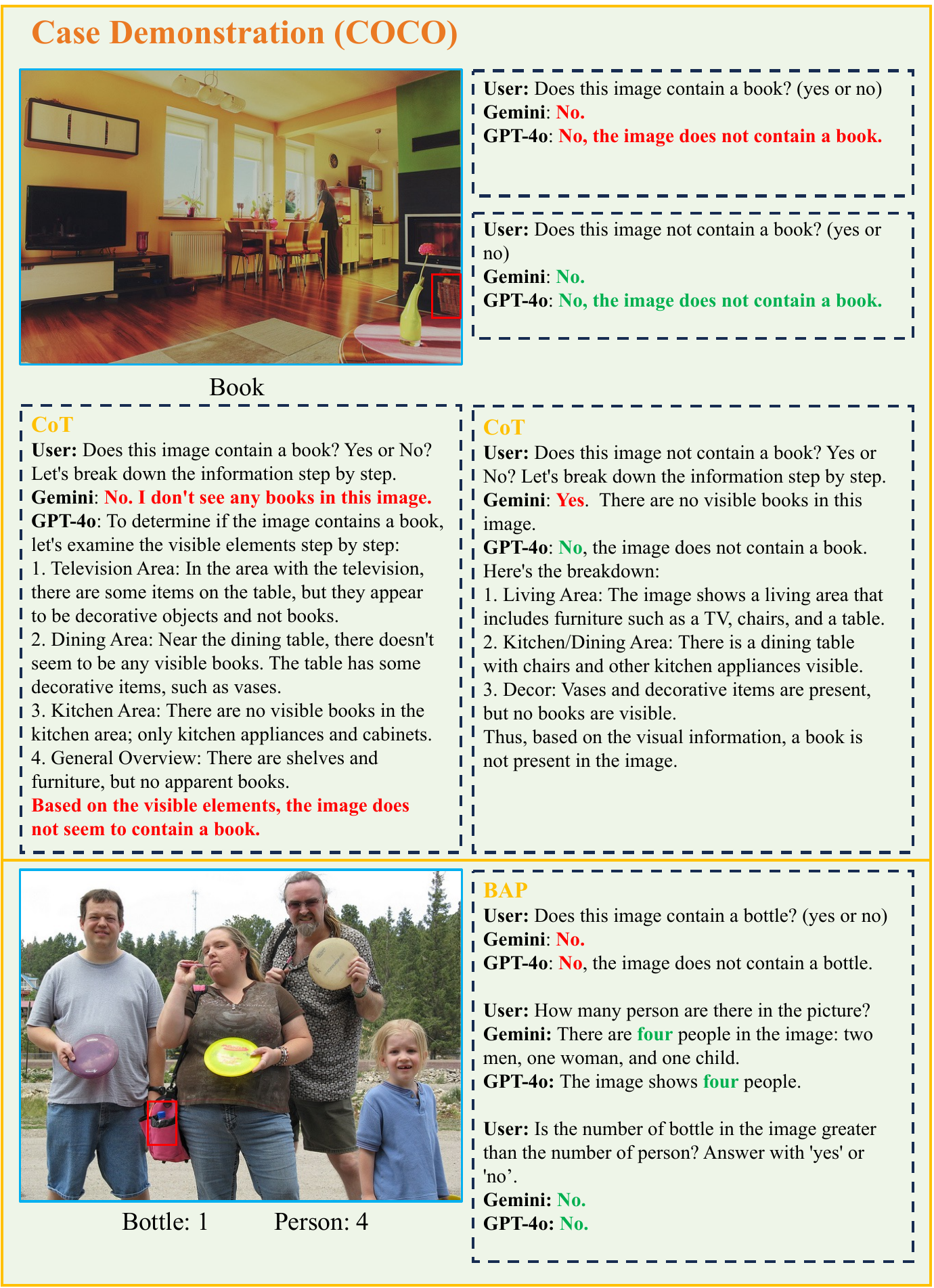}
    \caption{This is a showcase of OODBench examples on the COCO dataset. Examples cover standard question-and-answer, chain-of-thought prompts, and task types such as existential, counting, and logical reasoning. Green font indicates correct responses given by the model, and red font indicates incorrect responses.}
    \label{fig: case demonstration coco}
\end{figure*}

\begin{figure*}[ht]
    \centering
        \includegraphics[width=0.8\linewidth]{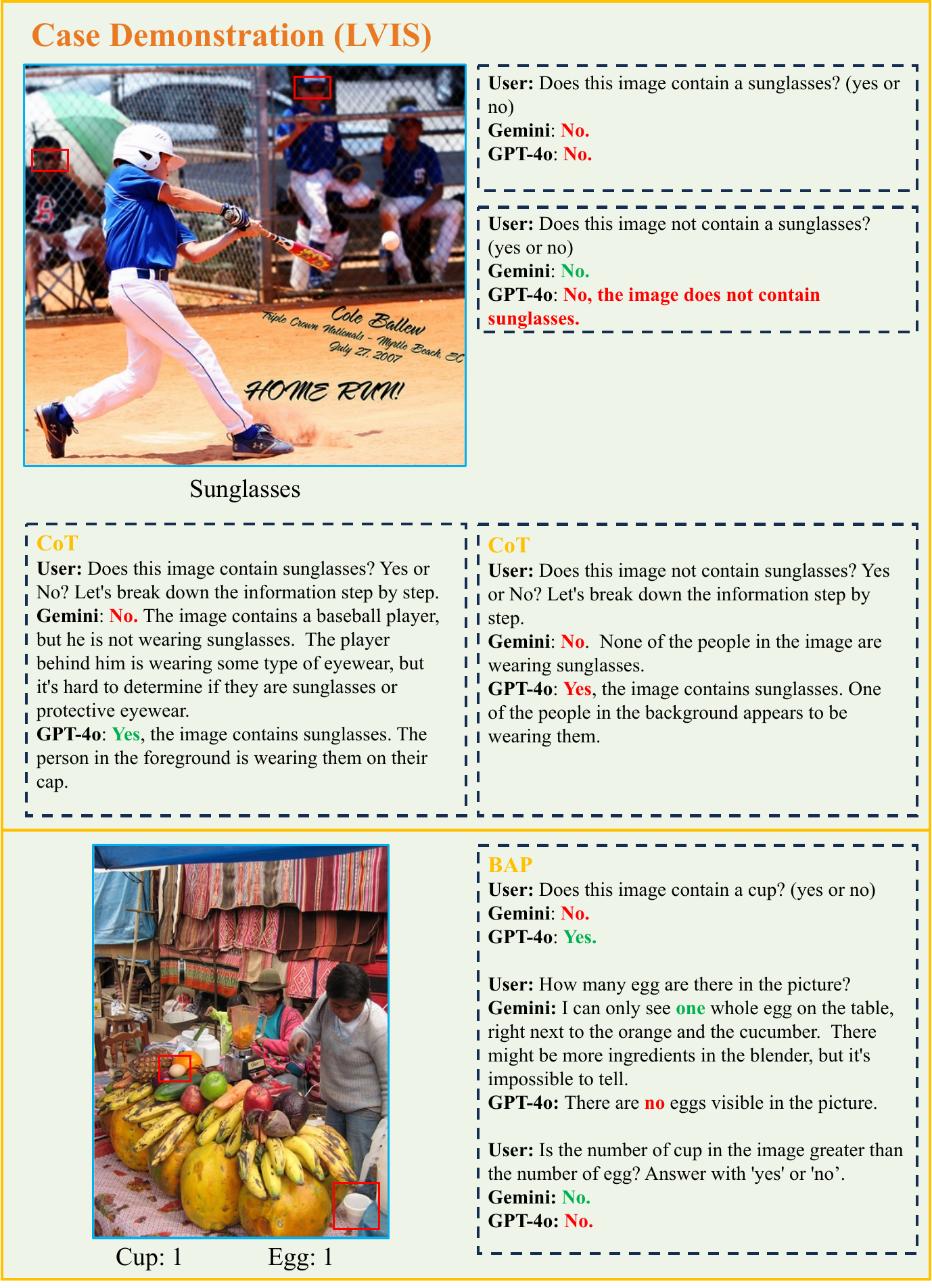}
    \caption{Demonstration of OODBench cases on the LVIS dataset. The questions shown cover a variety of task formats and visualise the performance of the model in an out-of-distribution scenario. Green font indicates correct responses, and red font indicates incorrect responses.}
    \label{fig: case demonstration lvis}
\end{figure*}

\begin{figure*}[ht]
    \centering
        \includegraphics[width=0.8\linewidth]{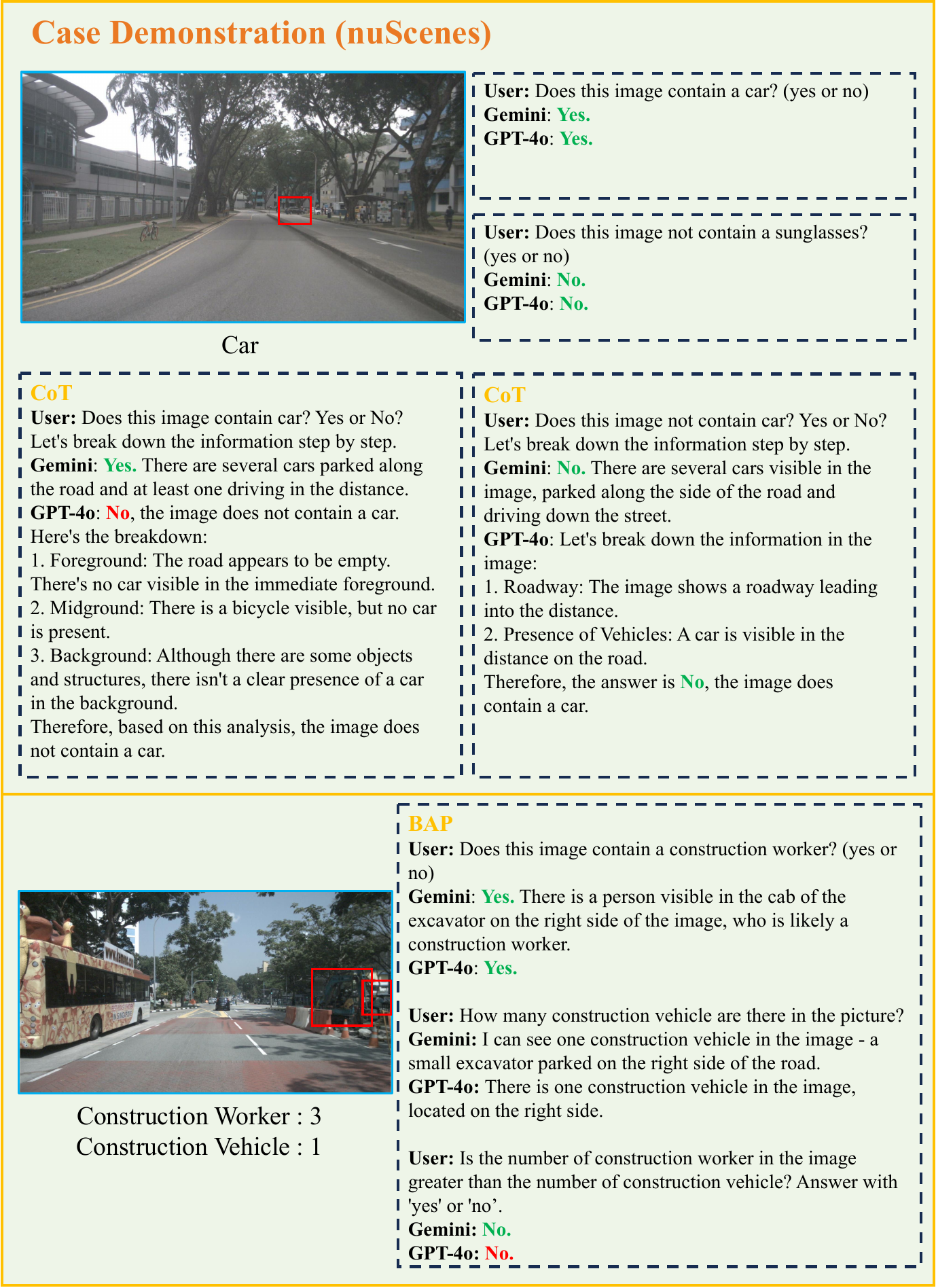}
    \caption{OODBench case demonstrations on the nuScenes dataset. It includes visual Q\&A and complex inference tasks in multimodality; green font indicates the model's correct predictions, and red font indicates incorrect predictions.}
    \label{fig: case demonstration nuscenes}
\end{figure*}

\begin{figure*}[ht]
    \centering
        \includegraphics[width=0.8\linewidth]{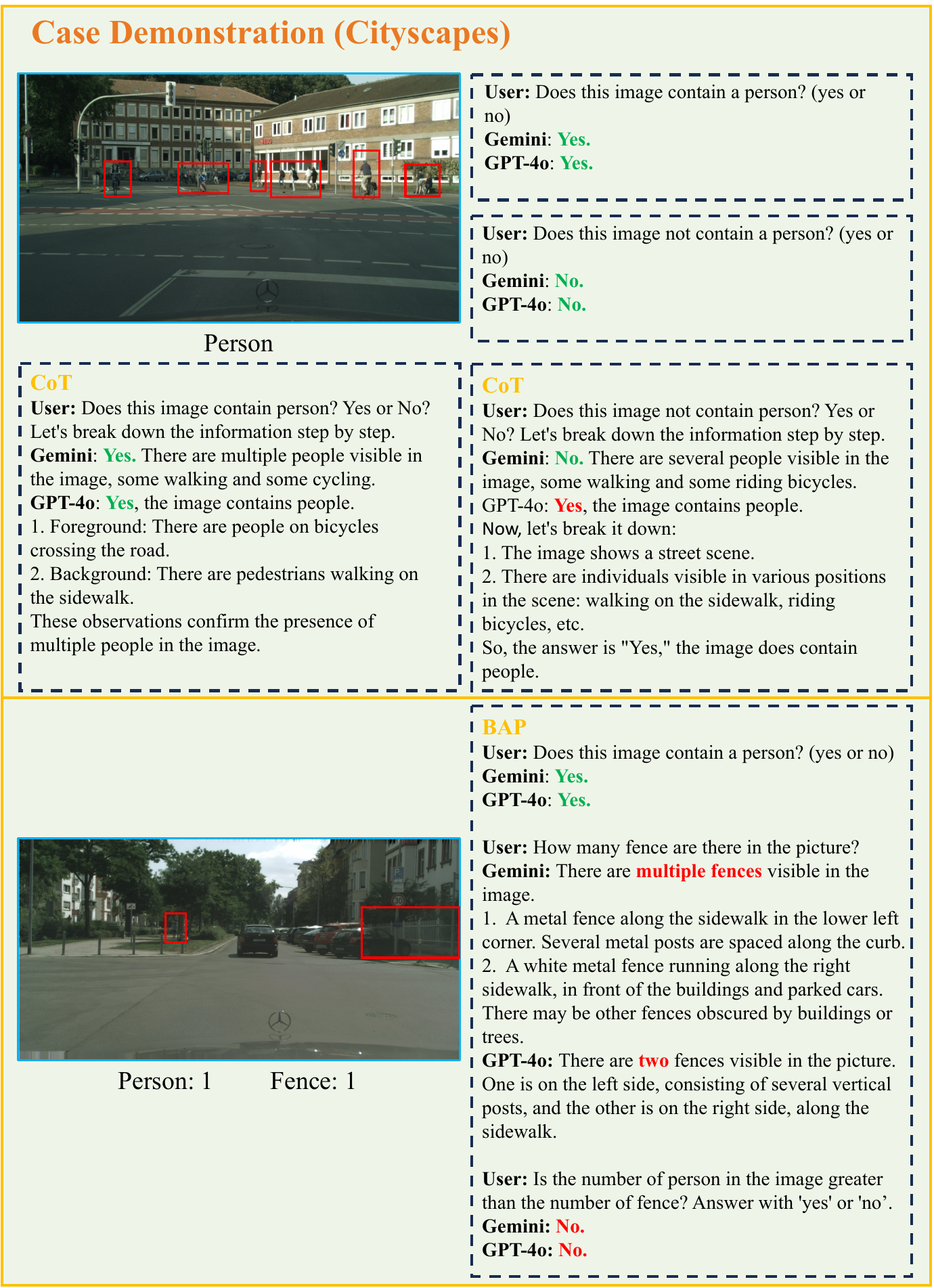}
    \caption{A showcase of OODBench cases on the Cityscapes dataset. A wide range of question types is covered to assess the model's ability to generalize to out-of-distribution environments such as urban streetscapes. Green font represents correct responses, and red font represents incorrect responses.}
    \label{fig: case demonstration cityscapes}
\end{figure*}

To further demonstrate the applicability and challenge of OODBench, we constructed typical examples on four datasets, COCO, LVIS, nuScenes, and Cityscapes, which are shown in Figs.~\ref{fig: case demonstration coco},~\ref{fig: case demonstration lvis},~\ref{fig: case demonstration nuscenes},~\ref{fig: case demonstration cityscapes}, respectively. Each set of examples covers a variety of question types, including standard question answering, Chain-of-Thought prompting, and tasks such as existential, counting, and logical reasoning under BAP metrics. These cases visualise the challenges of perception and reasoning in large multimodal models in different out-of-distribution scenarios.

\begin{figure*}[t]
  \centering
   \includegraphics[width=0.8\linewidth]{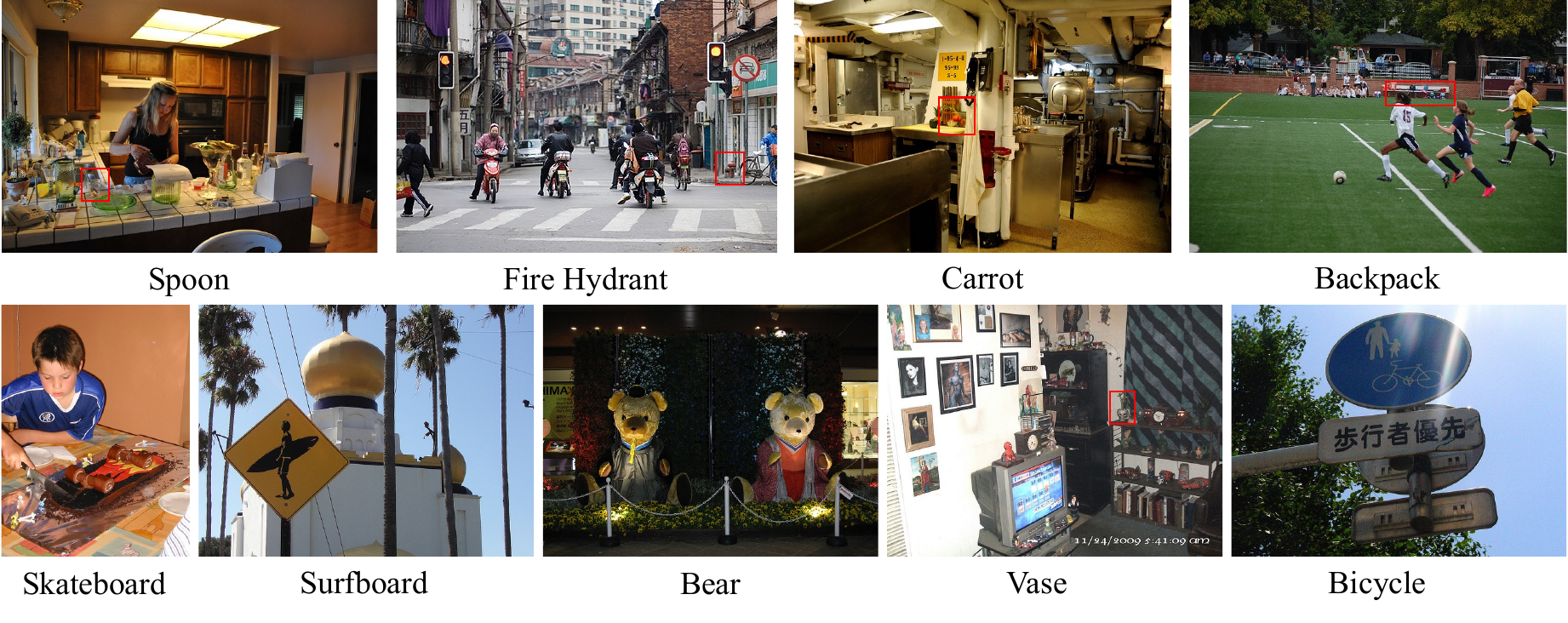}
   \caption{Typical failure cases of GPT-4o on OODBench. Errors primarily fall into two categories: non-main semantic objects and semantic variants.}
   \label{fig: error analysis}
\end{figure*}

\section{Analysis of Error Cases}

As shown in Figure~\ref{fig: error analysis}, we present a qualitative analysis of failure patterns for the representative GPT-4o model on OODBench. Errors primarily cluster into two semantic OOD categories: non-main objects and semantic variants.

\textbf{(1) Failures caused by non-primary semantic objects.} When a clear main semantic object exists in an image (\textit{e.g.}, "people" in the first and fourth examples of the first row in Figure~\ref{fig: error analysis}), the model’s image–caption pretraining objective tends to emphasize alignment with the main semantic object, while learning about non-main semantic objects is insufficient or even absent. Therefore, when we construct scenarios in OODBench where the main semantic object is present, but the question involves a non-main semantic object, the model appears to lack stable image–text associations for these semantics, leading to consistent mispredictions.

\textbf{(2) Failures due to semantic variants.} Another common error category stems from semantic variants. For example, the "skateboard made of cake" in the first sample of the second row in Figure~\ref{fig: error analysis}. During training, the model predominantly encounters “skateboards” with standard appearances drawn from natural image distributions. The joint distribution of images and text forms the model's default semantic template for the concept of "skateboard." When we replace the target object with variants featuring different materials, shapes, or construction methods (such as a "cake-shaped skateboard"), this sample deviates from the training distribution within the model's image-text semantic space. Consequently, the model misclassifies or ignores it.

These qualitative observations suggest that failures under semantic shifts are not random but linked to mismatches between joint image--text distributions and pretrained representations, consistent with our quantitative results.

\section{Limitations}
\label{appendix: limitations}
While OODBench offers a fully automated covariate‐shift benchmark for VLMs, it remains limited in scope. First, it is confined to general natural and autonomous driving domains and does not cover other real‐world domains (\textit{e.g.}, medical imaging, satellite imagery). In these domains, researchers must redefine appropriate OOD types based on their specific semantic structures, task requirements, and distribution characteristics. Simultaneously, the performance of specialized scoring models directly impacts the quality of initial screening, while subsequent expert evaluations (\textit{e.g.}, in medical settings) may incur additional labor costs. Therefore, the OOD definition within this framework should not be directly applied to the aforementioned domains. Second, it focuses solely on static vision data and omits temporal or multimodal variations. Third, it lacks extreme environmental conditions (\textit{e.g.}, severe weather, low light). Extending OODBench to additional domains, dynamic data, and challenging environmental factors will be an important direction for future work.



\end{document}